\documentclass{article}

\usepackage[final]{corl_2026} %

\usepackage[T1]{fontenc}
\usepackage{amsmath}
\usepackage{amssymb}
\usepackage{amsfonts}
\usepackage{todonotes}
\usepackage{multirow}
\usepackage{booktabs}       %
\usepackage[table]{xcolor}  %
\usepackage{siunitx}        %
\usepackage{subcaption}
\usepackage{cleveref}
\usepackage{nicefrac}
\usepackage{mathtools}
\usepackage{placeins}
\usepackage{bm}

\definecolor{highlight_0}{HTML}{ffbf87} %
\definecolor{highlight_1}{HTML}{ffcc9f} %
\definecolor{highlight_2}{HTML}{ffe5cf}  %

\newcommand{\given}{~|~}

\newcommand{\ci}[2]{\,{\tiny$\substack{#2\\#1}$}}
\newcommand{\lowerBetter}{$\downarrow$}
\newcommand{\higherBetter}{$\uparrow$}

\usepackage{graphicx}
\usepackage{booktabs}
\usepackage{array}
\newcommand{\tablesize}{\small}
\usepackage{xcolor}
\newcommand{\detrowlabelACC}[1]{%
    \makebox[0.01\linewidth][c]{%
        \raisebox{1.0em}{\rotatebox{90}{\textbf{#1}}}%
    }%
}
\newcommand{\detrowlabelLLMD}[1]{%
    \makebox[0.01\linewidth][c]{%
        \raisebox{0.5em}{\rotatebox{90}{\textbf{#1}}}%
    }%
}

\title{VLA-FAIL: Efficient Task Failure Detection for Finetuned Vision-Language-Action Models}

\author{
  \textbf{Florian Seligmann}\textsuperscript{\normalfont 1,$\dagger$} \quad
  \textbf{Emiliyan Gospodinov}\textsuperscript{\normalfont 1} \\
  \textbf{Enes Ulas Dincer}\textsuperscript{\normalfont 1} \quad
  \textbf{Gerhard Neumann}\textsuperscript{\normalfont 1,2} \\[0.2cm]
  \textsuperscript{1}Autonomous Learning Robots Lab, Karlsruhe Institute of Technology \\
  \textsuperscript{2}FZI Forschungszentrum Informatik, Karlsruhe \\[0.1cm]
  \textsuperscript{$\dagger$}Correspondence to: \texttt{florian.seligmann@kit.edu}
}

\begin{document}
\maketitle

\begin{abstract}
Vision-language-action models (VLAs) achieve state-of-the-art performance on many robotic manipulation tasks, yet they can still behave unpredictably in out-of-distribution scenarios.  
Runtime failure detection is therefore essential for the safe real-world deployment of VLAs. 
However, existing task failure detectors require computationally expensive action sampling, are based on architectural assumptions that limit their applicability to VLAs, or need access to failure rollouts.
We propose VLA-FAIL, a lightweight and broadly applicable failure detection framework for VLAs that combines two novel failure detectors with minimal overhead, without requiring failure data. 
The first, last-layer Mahalanobis distance (LLMD), detects out-of-distribution states by measuring token-wise deviations in last-layer features relative to the training data.
The second, action chunk consistency (ACC), exploits the temporal overlap induced by receding-horizon control and detects failures when consecutive action chunks become inconsistent.
To capture the trade-off between detection accuracy and detection latency, we introduce \textsc{AUCPDT}, a threshold-independent metric that jointly evaluates precision, recall, and detection time.
Through extensive real-world and simulation experiments, we demonstrate that LLMD and ACC capture complementary failure modes whose combination enables reliable and early failure detection across diverse tasks, frequently outperforming significantly more expensive baseline methods.
Project page: \url{https://vla-fail.github.io/vla-fail-2026}
\end{abstract}


\section{Introduction}

Vision-language-action models (VLAs) have emerged as powerful generalist policies for robotic manipulation, combining pretrained vision-language representations with large-scale robotic datasets across tasks and embodiments~\citep{black2024pi0,intelligence2025pi05,gr00tn1,lee2025molmoact,fang2026molmoact2,kim24openvla}.
Despite their strong performance, VLAs exhibit a wide range of failure modes in out-of-distribution scenarios, even after finetuning on the downstream task \citep{gu2026safe}.
For the safe real-world deployment of VLAs, it is therefore crucial to predict such failures as early as possible during runtime to enable human intervention \citep{ensemble_dagger,lee2025diff} or safe fallbacks \citep{brunke2022safe-learning}.

However, reliable runtime failure detection is challenging. 
A supervised formulation requires representative failure rollouts, but collecting failures is costly and potentially unsafe. %
Novelty-based detection is insufficient, as out-of-distribution states might not necessarily imply task failures, while some failures might arise from internally inconsistent action generation without an obvious visual anomaly. 
A practical failure detector should therefore require no failure data, accurately detect failures, be applicable to state-of-the-art VLA architectures, and incur minimal computational overhead.
This last requirement is critical because real-time VLA inference is already expensive, and additional latency can itself perturb closed-loop execution and induce out-of-distribution states.

We present \textbf{VLA-FAIL}, a lightweight failure detection framework for finetuned VLAs that requires no failure data or external models and incurs minimal computational overhead.
It monitors two complementary indicators of unreliable execution: internal feature shift and action plan inconsistencies. 
The first detector, last-layer Mahalanobis distance (LLMD), measures token-wise deviations from the training data's feature distribution. 
By directly monitoring internal representations computed from a fixed prior noise sample, LLMD removes feature variation introduced by stochastic action sampling and can detect out-of-distribution states even when actions appear smooth.
Our second detector, action chunk consistency (ACC), targets erratic behavior.
ACC measures the velocity-normalized disagreement between overlapping action chunks in receding-horizon control, where persistent inconsistency signals likely failure.
VLA-FAIL triggers whenever either LLMD or ACC exceeds its calibrated threshold, enabling accurate, early detection of task failures in VLAs.

We evaluate VLA-FAIL on two state-of-the-art VLAs, $\pi_{0.5}$ \citep{intelligence2025pi05} and X-VLA \citep{zheng2026xvla}, across six diverse real-world manipulation tasks and the large-scale LIBERO-Plus simulation benchmark \citep{fei25libero-plus}.
Our results show that LLMD and ACC perform well on separate tasks, with only their combination, VLA-FAIL, being robust across tasks, frequently outperforming significantly more expensive baselines.
To evaluate not only whether failures are detected but also when they are detected, we introduce AUCPDT, a threshold-independent metric that jointly captures precision, recall, and detection time.

\begin{figure}[t!]
    \centering
    \hfill%
    \begin{subfigure}[b]{0.67\textwidth}
        \includegraphics[width=\textwidth]{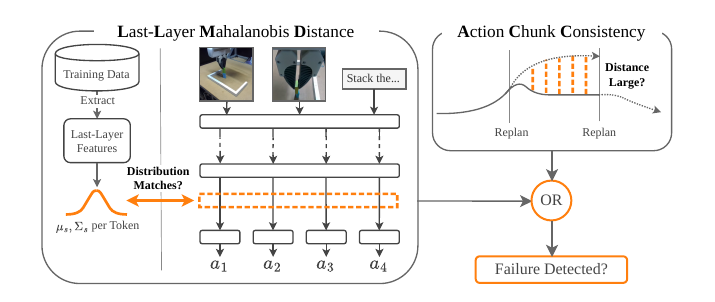}
        \caption{VLA-FAIL}
        \label{fig:overview}
    \end{subfigure}%
    \hfill%
    \begin{subfigure}[b]{0.32\textwidth}
        \centering
        \includegraphics[width=\textwidth, trim={0 0.25cm 0 0}, clip]{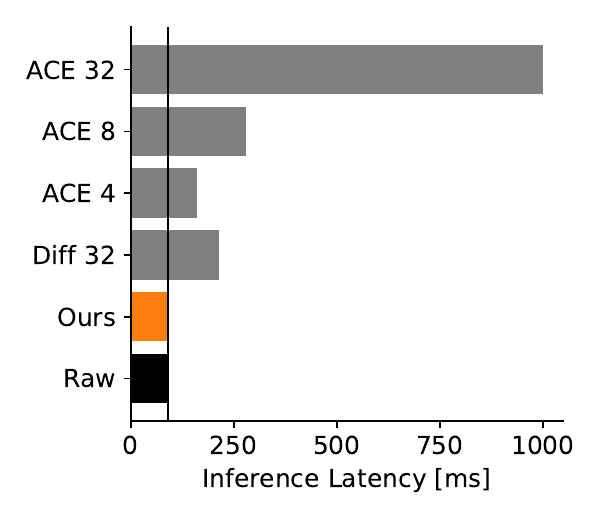}
        \caption{Inference Latency}
        \label{fig:latency}
    \end{subfigure}
    \hfill%
    \caption{\textbf{(a)} Our task failure detection pipeline: We combine LLMD, which detects unlikely last-layer features under the feature distribution of the training data, with ACC, which measures the consistency between overlapping parts of successive action chunks. \textbf{(b)} Action chunk sample time for various failure detectors (X-VLA~\citep{zheng2026xvla}, RTX 5090, constant \qty{17}{\milli\second} of VLM excluded). Baselines are not real-time capable due to their significant computational overhead. ACE~\citep{romer2026fiper} and Diff~\citep{lee2025diff} use the indicated number of action samples, and "Raw" represents no failure detection.}\vspace{-5.12pt} %
    \label{fig:teaser}
\end{figure}

Our contributions are three-fold: \textbf{1)} We propose VLA-FAIL, a lightweight runtime failure detection framework for VLAs that incurs minimal computational overhead and requires no failure data, combining LLMD, a token-wise last-layer feature Mahalanobis distance adapted to flow matching VLAs via fixed prior noise, and ACC, a velocity-normalized action chunk consistency score computed from already generated actions. 
\textbf{2)} We introduce AUCPDT, a threshold-independent metric that evaluates the trade-off between accurate and early failure detection.
\textbf{3)} Across real-world and simulated manipulation tasks, we show that VLA-FAIL enables robust failure detection and frequently outperforms substantially more expensive baselines.

\section{Related Work}
\label{sec:related-work}

\textbf{Task Failure Detection for VLAs.}
SAFE~\citep{gu2026safe} studies multitask failure detection for VLAs, where detector models are trained on last-layer feature representations from both successful and failure rollouts.
RC-NF~\citep{Zhou_2026_CVPR} avoids failure data by learning a conditional normalizing flow model on successful demonstration data, but relies on external object-centric perception.
CycleVLA \citep{ma2026cyclevla} and FailSafe \citep{lin2025failsafe} use VLM reasoning to detect failing rollouts.
In contrast, VLA-FAIL does not require failure data or external VLMs and perception pipelines.
We see these approaches as complementary to our work, but unsuitable for the goal of fast detection of potentially unknown failure modes.

\textbf{Failure Signals for Generative Robot Policies.}
For the broader class of generative imitation learning policies, a body of work detects task failure by monitoring observational shifts, action uncertainty, temporal inconsistency, or semantic progress.
FIPER \citep{romer2026fiper} predicts failures without failure data by combining random network distillation on observation embeddings with action-chunk entropy from multiple sampled actions, while Sentinel~\citep{AgiaSinhaEtAl2024} combines STAC, which compares the action distributions of overlapping consecutive action chunks, with vision-language model (VLM)-based video question answering. 
FAIL-Detect~\citep{xu2025can} evaluates OOD-based, smoothness-based, and consistency-based strategies and finds that a learned flow-matching density estimation model in the observation embedding space performs best. 
FIDeL~\citep{rolland2026failure} detects visual anomalies via optimal transport to nominal demonstrations and filters them further with a VLM.
Independent of the policy, forward dynamics model-based anomaly detectors \citep{Malhotra2015LongST,park2016multimodal_monitoring,liu2018ano_pred} measure the discrepancy between predicted future observations from a learned dynamics model and real future observations.
Such approaches, while superficially similar to our ACC detector, require an additional trained model and are expensive and noisy when applied to image observations.
In concurrent work, \citet{zheng2026rewindilonlinefailuredetection} develop Rewind-IL, which uses the discrepancy between overlapping action chunks as a failure signal.
Another line of work learns VLM-based reward models \citep{tan2025robo,liang2026robometer,lee2025roboreward}, which can infer task failure from image observations. 
In the context of robot-gated DAgger~\citep{ross2011dagger}, ensemble variance \citep{ensemble_dagger} and diffusion loss \citep{lee2025diff} have been proposed to detect task failure and trigger expert interventions. 
In contrast, our VLA-FAIL framework does not require multiple action samples or expensive diffusion loss evaluations per environment timestep, and does not rely on slow VLMs or video models.

\textbf{OOD Detection for Deep Neural Networks.}
The feature-based detector of VLA-FAIL builds on general out-of-distribution (OOD) detection methods for deep neural networks. Lee et al.~\citep{lee2018maha-ood} develop an OOD score using a Mahalanobis distance on the last-layer features of image classification models relative to the training data. Subsequent works improve this principle further \citep{ren2021simplefixmahalanobisdistance,muller2025mahalanobis}, but remain in the context of classification models. 
The last-layer Mahalanobis distance is closely related to the last-layer Laplace approximation \citep{NEURIPS2021_a7c95857}, a principled and computationally efficient method from Bayesian deep learning.
We extend Mahalanobis-based OOD detection to flow matching models with continuous actions and noisy features.

\section{Method}

VLA-FAIL detects task failures by combining two complementary signals.
The last-layer Mahalanobis distance (LLMD, \Cref{sec:method:llmd}) measures deviations in the policy's internal representations relative to the finetuning data, allowing us to capture failures even when actions appear consistent.
We combine LLMD with action chunk consistency (ACC, \Cref{sec:method:acc}), which detects inconsistencies between overlapping actions chunks.
Their thresholded combination defines \textbf{F}ailure Detection via \textbf{A}ction \textbf{I}nconsistency and \textbf{L}ast-Layer Features (VLA-FAIL, \Cref{sec:method:combination}), a lightweight detector with minimal computational overhead that requires no failure data and no auxiliary model.

\textbf{Problem Formulation}.
We consider a vision-language-action model (VLA) that controls a robotics system.
At environment timestep $t$, the VLA receives an observation $o \in \mathcal{O}$ and predicts a sequence of $H$ actions $a_{1:H} \in \mathcal{A}^{H}$, where $\mathcal{A} = \mathbb{R}^D$ is the $D$-dimensional action space.
The observation $o$ consists of the robot's proprioceptive state, multiple camera images taken at $t$, and a language instruction.
Using receding horizon control, the first $R \leq H$ predicted actions are applied to the environment before the VLA is re-queried.
After $T$ timesteps, the episode ends, and the policy receives a ground-truth success or failure signal.
The goal of task failure detection is to predict failures before the episode horizon $T$ is reached.

\subsection{Last-Layer Token-Wise Mahalanobis Distance (LLMD)}
\label{sec:method:llmd}

Feature-based out-of-distribution detection provides insights into model behavior that actions alone cannot capture, with the features directly before the last layer of a neural network being particularly indicative of out-of-distribution (OOD) inputs \citep{lee2018maha-ood,ren2021simplefixmahalanobisdistance,muller2025mahalanobis}.
We assume the VLA decomposes into a feature extractor $f:~\mathcal{O} \to \mathcal{F}$, where $\mathcal{F} = \mathbb{R}^{H \times F}$, that maps observations to the high-dimensional last-layer feature space $\mathcal{F}$ and a final linear layer $g: \mathcal{F} \to \mathcal{A}^H$ that individually projects each feature token to an action (\Cref{fig:overview}).
While we focus on flow-matching experts due to their prevalence in modern VLAs \citep{black2024pi0,intelligence2025pi05,reuss2025flower,zheng2026xvla,gr00tn1,fang2026molmoact2},
LLMD equally applies to diffusion \citep{li2025vitra} and discrete VLAs \citep{kim24openvla,lee2025molmoact}.

\textbf{Fixed Prior Noise.}
To avoid mode-averaging of the expert's action distribution while ensuring continuous actions, VLAs typically use flow matching.
Instead of directly predicting the action $a$ given an observation $o$, the policy iteratively predicts a time-dependent flow velocity $v_t(o, a_t)$.
Integrating this velocity over $t$, starting from a simple prior $a_0 \sim \mathcal{N}(0, I)$, yields the final action $a \coloneqq a_1 = a_0 + \int_0^1 v_t(o, a_t) \; \mathrm{d}t$.
The action expert is therefore conditioned not only on the current observation $o$, but also on the flow matching timestep $t$ and the current noisy action $a_t$.
This introduces observation-independent noise in the last-layer features.
To isolate the influence of the observation, we aim to fix $a_t$ to a single value that is in-distribution for all $o$.
Since $p_t(a \given o)$ depends on $o$ for all $t>0$, fixing $a_{t>0}$ would introduce artificial covariate shift.
However, at $t=0$, the noisy action distribution equals the prior $p_0(a \given o)  = \mathcal{N}(a; \; 0, I)$.
We therefore sample a single, fixed noise vector $a_0^* \sim \mathcal{N}(0, I)$ to compute training data statistics and evaluate rollouts on the reduced feature space defined by
\begin{equation}
    f^*(o) \coloneqq f(o, t=0, a_0^*), \quad a_0^* \sim \mathcal{N}(0, I)~\text{random but fixed}.
\end{equation}
Computing $f^*(o)$ requires only a single model forward pass that can be executed in parallel with action sampling.
Alternatively, if multi-modal behavior is not required, the action itself can be sampled starting from $a_0^*$, eliminating any computational overhead.

\textbf{Token-Wise Mahalanobis Distance.}
To detect OOD states given an observation $o$ during rollout, we measure the similarity between its noise-free last-layer features $f^*(o)$ and the distribution of features in the training data using a Mahalanobis distance.
Empirically, it is beneficial to compute the Mahalanobis distance \emph{token-wise}, i.e., separately for each token position.
We speculate that this accounts for differences in feature distributions across token positions, and ablate this choice in \Cref{sec:discussion}.
Let $f^*(o)_s$, $s \in \{1, \dots, H\}$, denote the $s$-th last-layer feature token.
After VLA finetuning, we perform a single, gradient-free pre-processing step on the dataset $\mathcal{D} = \{(o_i, a_i), i \in \{1, ..., |\mathcal{D}|\}\}$ to compute the token-wise mean $\mu_s$ and centered covariance matrix $\Sigma_s$:
\begin{align}
    \mu_s &= \frac{1}{|\mathcal{D}|} \sum_{i=1}^{|\mathcal{D}|} f^*(o_i)_s , &
    \Sigma_s &= \frac{1}{|\mathcal{D}|} \sum_{i=1}^{|\mathcal{D}|} (f^*(o_i)_s - \mu_s) (f^*(o_i)_s - \mu_s)^\top.
\end{align}
For numerical stability, we use the regularized covariance $\Sigma_s^{\lambda} = \Sigma_s + \lambda I$, $\lambda \ll 1$.
During a rollout, the LLMD score is the maximum of the squared token-wise Mahalanobis distances:
\begin{align}
    s^t_\text{LLMD}(o_t) = \max_{s \in \{1, \dots, H\}} \left[(f^*(o_t)_s - \mu_s)^\top \left(\Sigma_s^\lambda\right)^{-1} (f^*(o_t)_s - \mu_s)\right]
\end{align}
where higher values indicate more likely failure. 
The maximum aggregation ensures that a single unlikely feature triggers a task failure.
We find that $s^t_\text{LLMD}$ requires no temporal smoothing.

\begin{figure}[t]
    \centering
    \begin{subfigure}[t]{0.45\textwidth}
        \centering
        \includegraphics[width=\textwidth, trim={0 0 0 12pt},clip]{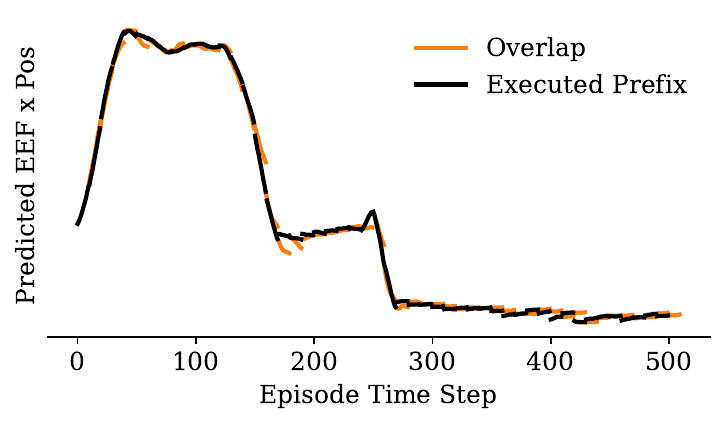}
        \caption{Successful episode}
    \end{subfigure}%
\hfill%
    \begin{subfigure}[t]{0.45\textwidth}
        \centering
        \includegraphics[width=\textwidth, trim={0 0 0 12pt},clip]{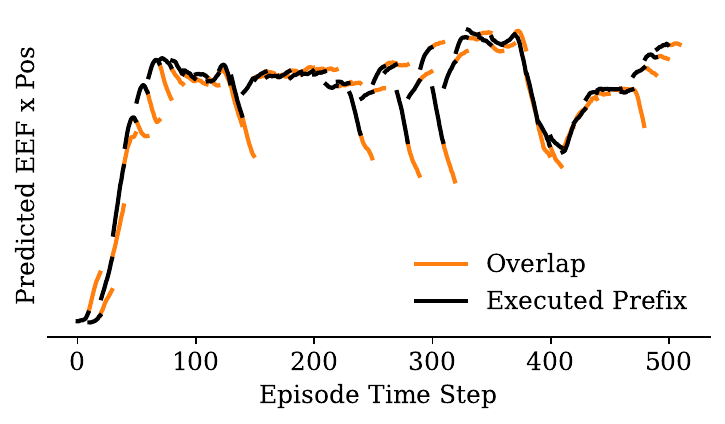}
        \caption{Failed episode}
    \end{subfigure}
    \caption{Predicted end-effector (EEF) $x$ position over episode time for a successful and a failed episode of $\pi_{0.5}$ on \texttt{Blocks}. In successful episodes, the predicted but not executed actions of an action chunk typically overlap with the next action chunk. In contrast, the failed episode shows substantial misalignment between consecutive action chunks. ACC measures this inconsistency.}
    \label{fig:acc-overlap-plots}
\end{figure}

\subsection{Action Chunk Consistency (ACC)}
\label{sec:method:acc}
Inconsistent action sequences are a typical sign of out-of-distribution states.
Under receding horizon control, only the first $R < H$ actions are executed, causing an overlap between the unexecuted suffix $a^{t-1}_{R+1:H}$ of the previous action chunk with the prefix $a^t_{1:H-R}$ of the next inference step. 
Failing policies generally lack self-consistent planning, leading to divergent actions, as shown in \Cref{fig:acc-overlap-plots}.
We therefore propose \emph{action chunk consistency} (ACC) as a failure detector, which measures the velocity-normalized discrepancy between overlapping action sequences.
Requiring only a single action sample per timestep, ACC remains computationally feasible for real-time control with VLAs.

We normalize ACC by the robot's average velocity, since the same absolute action mismatch is more critical during slow, precise motions than during fast motions.
Let
\begin{equation}
    v^t_d = \max \Big(\max_{s \in \{1, ..., H-R\}} a^{t}_{s, d} - \min_{s \in \{1, ..., H-R\}} a^{t}_{s, d}, \; v_\text{min} \Big)
\end{equation}
denote the average velocity in action dimension $d$ at timestep $t$, clamped to a minimum velocity of $v_\text{min}$ to avoid inflation of the ACC score for a near-stationary robot.
The unsmoothed ACC score at timestep $t$ is then the velocity-normalized mean absolute error across $D$ action dimensions:
\begin{equation}
    \hat{s}^t_\text{ACC} = \frac{1}{D} \frac{1}{H - R} \sum_{d=1}^D \frac{1}{v^t_d} \sum_{i = 1}^{H - R} | a^{t-1}_{R + i, d} - a^{t}_{i, d}|%
\end{equation}
In practice, we calculate the ACC score using only the $D=3$ predicted absolute end-effector position, which can be obtained for any control method.
Since replanning can naturally occur during successful execution, and only persistent replanning indicates likely failure, we apply an exponential moving average with smoothing parameter $\alpha \in [0, 1)$ on the ACC score:
\begin{equation}
    s^t_\text{ACC} = \alpha s^{t - 1}_\text{ACC} + (1 - \alpha) \hat{s}^t_\text{ACC}.
\end{equation}
Empirically, we find that strong smoothing ($\alpha = 0.9$) performs well across all tasks and is required for good detection rates.

\subsection{VLA-FAIL}
\label{sec:method:combination}
ACC and LLMD have distinct advantages: ACC can reliably detect many task failures, including those in which the robot exhibits dangerous behavior, whereas LLMD can detect failures early, as it does not rely on erratic robot behavior and does not need temporal smoothing.
We combine ACC and LLMD into a single failure detector, FAIL, that alerts if either method detects failure:
\begin{equation}
    F_{\text{FAIL}}(t) := (s_{\text{ACC}}^{t} \geq \tau_\text{ACC}) \lor (s_{\text{LLMD}}^{t} \geq \tau_{\,\text{LLMD}})
\end{equation}
for score thresholds $\tau_\text{ACC} \in \mathbb{R}$ and $\tau_\text{LLMD} \in \mathbb{R}$ that are determined via a time-constant conformal prediction band on calibration data \citep{romer2026fiper}.
We do not use a time-dependent threshold \citep{xu2025faildetect,romer2026fiper,gu2026safe} as it is not applicable to episodes that vary significantly in length, such as in our real-world \texttt{Drawer} task.

\section{Experiments}
\label{sec:eval}

To validate that our methods can reliably detect task failure early, we evaluate them across six diverse real-world tasks and the large-scale Libero-Plus simulation benchmark.

\textbf{Baselines and Models.}
We compare FAIL against three action- and loss-based methods.
\textbf{ACE} \citep{romer2026fiper} computes the entropy of the policy's action distribution at each environment timestep.
\textbf{STAC} \citep{AgiaSinhaEtAl2024} evaluates the distributional divergence between the overlapping parts of consecutive action distributions.
\textbf{Diff} \citep{lee2025diff} uses the policy's predicted action chunk as a pseudo-label to evaluate the diffusion loss across multiple diffusion timesteps and prior noise samples.
We use $32$ samples for all baselines and evaluate them on replays of policy rollouts, as they are not real-time capable (\Cref{fig:latency}).
We benchmark FAIL and baselines on two state-of-the-art VLAs:
$\bm{\pi_{0.5}}$ \citep{intelligence2025pi05} is a large, 3.6B-parameter VLA based on the PaliGemma VLM \citep{beyer2024paligemma} that fuses the action expert and VLM features at every layer via cross-attention.
\textbf{X-VLA} \citep{zheng2026xvla} is a smaller 0.9B-parameter VLA that uses the Florence-2-Large encoder-decoder VLM \citep{xiao2024florence} and conditions the action expert only on the final encoder features.

\textbf{Real-World Tasks.}
We evaluate the task failure detectors on six diverse, challenging real-world tasks, requiring high precision (\texttt{Blocks}, \texttt{Stack T}), with multi-modal demonstrations (\texttt{Cups}), long, multi-stage episodes (\texttt{Kitchen}, \texttt{Drawer}), and requiring language-conditioning (\texttt{Mixer}).
We evaluate all policies on roughly $80$ rollouts, and, for each rollout, evaluate the failure detectors across 3 seeds to capture the variance in baselines and LLMD's prior noise sample.
See \Cref{appendix:eval_protocol} for further details.

\begin{table}[ht]
    \caption{Threshold-independent detector performance on the \textbf{real-world tasks}. PR and PDT (see~\Cref{sec:eval:metrics}) refer to the areas under the respective curves, and the small numbers indicate $95\%$ confidence intervals. FAIL achieves performance comparable to, and frequently better than, baseline methods, while incurring minimal computational overhead. While ACC and LLMD perform well on some tasks individually, only FAIL performs well on nearly all tasks. Colors mark the \colorbox{highlight_0}{best}, \colorbox{highlight_1}{second-best}, and \colorbox{highlight_2}{third-best} methods.}
    \centering
    \begin{tabular}{l @{\hspace{2pt}} l @{\hspace{4pt}} l *{3}{S[table-format=1.2, table-space-text-post={\ci{0.00}{0.00}}]} | S[table-format=1.2] *{2}{S[table-format=1.3, table-space-text-post={\ci{0.00}{0.00}}]}}
        \toprule
         & & & {ACE} & {Diff} & {STAC} & {ACC} & {LLMD} & {FAIL} \\
        \midrule

\multirow{4}{*}{\texttt{Blocks}} & \multirow{2}{*}{X-VLA} & PR\higherBetter & 0.82{\ci{0.80}{0.83}} & \cellcolor{highlight_2} 0.84{\ci{0.82}{0.86}} & 0.35{\ci{0.35}{0.35}} & \cellcolor{highlight_0} 0.92 & 0.83{\ci{0.82}{0.84}} & \cellcolor{highlight_1} 0.86{\ci{0.85}{0.87}} \\
& & PDT\lowerBetter & 0.32{\ci{0.31}{0.33}} & 0.28{\ci{0.27}{0.30}} & 0.46{\ci{0.46}{0.46}} & \cellcolor{highlight_0} 0.24 & \cellcolor{highlight_2} 0.26{\ci{0.25}{0.26}} & \cellcolor{highlight_1} 0.25{\ci{0.24}{0.25}} \\
 & \multirow{2}{*}{$\pi_{0.5}$} & PR\higherBetter & 0.60{\ci{0.56}{0.63}} & \cellcolor{highlight_1} 0.85{\ci{0.77}{0.93}} & \cellcolor{highlight_2} 0.84{\ci{0.82}{0.85}} & \cellcolor{highlight_0} 0.95 & 0.73{\ci{0.72}{0.74}} & 0.81{\ci{0.80}{0.81}} \\
& & PDT\lowerBetter & 0.45{\ci{0.42}{0.47}} & \cellcolor{highlight_2} 0.31{\ci{0.26}{0.36}} & 0.36{\ci{0.35}{0.37}} & \cellcolor{highlight_0} 0.24 & \cellcolor{highlight_2} 0.31{\ci{0.31}{0.32}} & \cellcolor{highlight_1} 0.27{\ci{0.27}{0.27}} \\

\midrule

\multirow{4}{*}{\texttt{Drawer}} & \multirow{2}{*}{X-VLA} & PR\higherBetter & \cellcolor{highlight_2} 0.89{\ci{0.87}{0.91}} & \cellcolor{highlight_1} 0.90{\ci{0.89}{0.91}} & 0.18{\ci{0.18}{0.18}} & 0.83 & \cellcolor{highlight_1} 0.90{\ci{0.89}{0.91}} & \cellcolor{highlight_0} 0.91{\ci{0.90}{0.92}} \\
& & PDT\lowerBetter & \cellcolor{highlight_2} 0.23{\ci{0.23}{0.24}} & 0.25{\ci{0.25}{0.25}} & 0.71{\ci{0.71}{0.71}} & 0.33 & \cellcolor{highlight_0} 0.19{\ci{0.19}{0.19}} & \cellcolor{highlight_1} 0.22{\ci{0.22}{0.22}} \\
 & \multirow{2}{*}{$\pi_{0.5}$} & PR\higherBetter & 0.31{\ci{0.28}{0.35}} & 0.63{\ci{0.50}{0.76}} & \cellcolor{highlight_0} 0.72{\ci{0.70}{0.73}} & 0.62 & \cellcolor{highlight_1} 0.69{\ci{0.66}{0.72}} & \cellcolor{highlight_2} 0.68{\ci{0.65}{0.71}} \\
& & PDT\lowerBetter & 0.68{\ci{0.66}{0.69}} & \cellcolor{highlight_2} 0.49{\ci{0.42}{0.57}} & 0.50{\ci{0.50}{0.50}} & 0.53 & \cellcolor{highlight_0} 0.45{\ci{0.44}{0.46}} & \cellcolor{highlight_1} 0.47{\ci{0.45}{0.48}} \\

\midrule

\multirow{4}{*}{\texttt{Cups}} & \multirow{2}{*}{X-VLA} & PR\higherBetter & 0.89{\ci{0.88}{0.89}} & 0.91{\ci{0.91}{0.91}} & \cellcolor{highlight_0} 0.98{\ci{0.98}{0.98}} & \cellcolor{highlight_1} 0.97 & 0.78{\ci{0.78}{0.79}} & \cellcolor{highlight_2} 0.93{\ci{0.92}{0.93}} \\
& & PDT\lowerBetter & \cellcolor{highlight_1} 0.28{\ci{0.28}{0.29}} & \cellcolor{highlight_0} 0.27{\ci{0.27}{0.27}} & 0.42{\ci{0.42}{0.42}} & \cellcolor{highlight_1} 0.28 & \cellcolor{highlight_2} 0.33{\ci{0.33}{0.34}} & \cellcolor{highlight_1} 0.28{\ci{0.28}{0.29}} \\
 & \multirow{2}{*}{$\pi_{0.5}$} & PR\higherBetter & 0.41{\ci{0.39}{0.43}} & 0.51{\ci{0.50}{0.53}} & \cellcolor{highlight_1} 0.61{\ci{0.60}{0.61}} & \cellcolor{highlight_2} 0.60 & 0.58{\ci{0.57}{0.59}} & \cellcolor{highlight_0} 0.63{\ci{0.62}{0.64}} \\
& & PDT\lowerBetter & 0.61{\ci{0.60}{0.62}} & 0.57{\ci{0.55}{0.58}} & \cellcolor{highlight_2} 0.52{\ci{0.51}{0.52}} & \cellcolor{highlight_1} 0.51 & \cellcolor{highlight_2} 0.52{\ci{0.52}{0.53}} & \cellcolor{highlight_0} 0.49{\ci{0.49}{0.50}} \\

\midrule

\multirow{4}{*}{\texttt{Kitchen}} & \multirow{2}{*}{X-VLA} & PR\higherBetter & \cellcolor{highlight_1} 0.98{\ci{0.97}{0.99}} & \cellcolor{highlight_1} 0.98{\ci{0.98}{0.98}} & \cellcolor{highlight_2} 0.30{\ci{0.30}{0.30}} & \cellcolor{highlight_0} 1.00 & \cellcolor{highlight_0} 1.00{\ci{1.00}{1.00}} & \cellcolor{highlight_0} 1.00{\ci{1.00}{1.00}} \\
& & PDT\lowerBetter & \cellcolor{highlight_2} 0.24{\ci{0.24}{0.24}} & \cellcolor{highlight_1} 0.21{\ci{0.20}{0.21}} & 0.55{\ci{0.55}{0.55}} & 0.25 & \cellcolor{highlight_0} 0.20{\ci{0.20}{0.20}} & \cellcolor{highlight_1} 0.21{\ci{0.21}{0.21}} \\
 & \multirow{2}{*}{$\pi_{0.5}$} & PR\higherBetter & \cellcolor{highlight_1} 0.95{\ci{0.95}{0.96}} & 0.75{\ci{0.66}{0.85}} & \cellcolor{highlight_2} 0.92{\ci{0.91}{0.92}} & \cellcolor{highlight_0} 0.99 & 0.62{\ci{0.41}{0.82}} & 0.74{\ci{0.64}{0.84}} \\
& & PDT\lowerBetter & \cellcolor{highlight_1} 0.20{\ci{0.19}{0.21}} & 0.30{\ci{0.27}{0.34}} & \cellcolor{highlight_2} 0.26{\ci{0.25}{0.26}} & \cellcolor{highlight_0} 0.17 & 0.33{\ci{0.20}{0.46}} & 0.28{\ci{0.16}{0.39}} \\

\midrule

\multirow{4}{*}{\texttt{Stack T}} & \multirow{2}{*}{X-VLA} & PR\higherBetter & \cellcolor{highlight_2} 0.96{\ci{0.96}{0.97}} & \cellcolor{highlight_1} 0.97{\ci{0.97}{0.97}} & 0.59{\ci{0.59}{0.59}} & \cellcolor{highlight_0} 0.98 & \cellcolor{highlight_1} 0.97{\ci{0.97}{0.97}} & \cellcolor{highlight_0} 0.98{\ci{0.98}{0.98}} \\
& & PDT\lowerBetter & \cellcolor{highlight_1} 0.09{\ci{0.09}{0.09}} & \cellcolor{highlight_0} 0.08{\ci{0.08}{0.08}} & \cellcolor{highlight_2} 0.21{\ci{0.21}{0.21}} & \cellcolor{highlight_1} 0.09 & \cellcolor{highlight_0} 0.08{\ci{0.08}{0.08}} & \cellcolor{highlight_0} 0.08{\ci{0.08}{0.08}} \\
 & \multirow{2}{*}{$\pi_{0.5}$} & PR\higherBetter & 0.82{\ci{0.81}{0.82}} & \cellcolor{highlight_2} 0.92{\ci{0.90}{0.94}} & 0.87{\ci{0.86}{0.88}} & \cellcolor{highlight_1} 0.95 & \cellcolor{highlight_1} 0.95{\ci{0.95}{0.96}} & \cellcolor{highlight_0} 0.96{\ci{0.96}{0.97}} \\
& & PDT\lowerBetter & 0.24{\ci{0.24}{0.24}} & \cellcolor{highlight_2} 0.13{\ci{0.12}{0.14}} & 0.17{\ci{0.17}{0.17}} & \cellcolor{highlight_1} 0.12 & \cellcolor{highlight_0} 0.10{\ci{0.10}{0.10}} & \cellcolor{highlight_0} 0.10{\ci{0.09}{0.10}} \\

\midrule

\multirow{4}{*}{\texttt{Mixer}} & \multirow{2}{*}{X-VLA} & PR\higherBetter & 0.93{\ci{0.93}{0.93}} & \cellcolor{highlight_2} 0.95{\ci{0.95}{0.96}} & 0.46{\ci{0.46}{0.46}} & \cellcolor{highlight_1} 0.97 & \cellcolor{highlight_0} 0.98{\ci{0.98}{0.98}} & \cellcolor{highlight_1} 0.97{\ci{0.97}{0.97}} \\
& & PDT\lowerBetter & \cellcolor{highlight_2} 0.16{\ci{0.15}{0.16}} & \cellcolor{highlight_1} 0.14{\ci{0.14}{0.14}} & 0.34{\ci{0.34}{0.34}} & \cellcolor{highlight_1} 0.14 & \cellcolor{highlight_0} 0.11{\ci{0.11}{0.11}} & \cellcolor{highlight_0} 0.11{\ci{0.11}{0.11}} \\
 & \multirow{2}{*}{$\pi_{0.5}$} & PR\higherBetter & \cellcolor{highlight_0} 0.93{\ci{0.92}{0.93}} & 0.61{\ci{0.54}{0.68}} & 0.70{\ci{0.69}{0.71}} & \cellcolor{highlight_1} 0.85 & 0.77{\ci{0.76}{0.78}} & \cellcolor{highlight_2} 0.79{\ci{0.79}{0.79}} \\
& & PDT\lowerBetter & \cellcolor{highlight_0} 0.24{\ci{0.24}{0.25}} & 0.45{\ci{0.40}{0.49}} & 0.43{\ci{0.42}{0.45}} & \cellcolor{highlight_2} 0.28 & \cellcolor{highlight_2} 0.28{\ci{0.27}{0.28}} & \cellcolor{highlight_1} 0.27{\ci{0.27}{0.27}} \\

        \bottomrule
    \label{tab:eval-real}
    \end{tabular}
\end{table}%

\paragraph{Simulated Tasks.}
We perform a large-scale evaluation on the Libero-Plus benchmark \citep{fei25libero-plus}, which uses the tasks from Libero \citep{liu2023libero} and adds various perturbations to test model generalization.
We use the official Libero checkpoints for X-VLA and $\pi_{0.5}$ and evaluate all methods across 3 seeds.
See \Cref{app:simulation_environments} for task details.

\subsection{Evaluation Metrics}
\label{sec:eval:metrics}

A good task failure detector must reliably detect failures as early as possible within an episode.
We treat failure detection as a binary classification problem and evaluate overall reliability using the area under the precision-recall (PR) curve (\textbf{AUCPR}), which eliminates threshold selection as a confounding variable (for threshold-dependent results see \Cref{sec:thresh-real}).
For FAIL, we use a simple fusion strategy to obtain a single classifier score: we apply individual rank transformations to bring the LLMD and ACC scores to a uniform scale, and then take their minimum, corresponding to the logical OR in \Cref{sec:method:combination}.
This strategy, while not strictly optimal, is task-agnostic and requires no prior knowledge of the relative performance of LLMD and ACC.

\textbf{Penalized Detection Time.}
AUCPR does not distinguish early from late detections.
Naively measuring the average first step of detected failure is also insufficient, because a detector that always triggers at $t=0$ achieves zero delay but unusable precision.
We therefore introduce the area under the penalized detection time curve (\textbf{AUCPDT}), which evaluates the trade-off between precision and detection latency in a threshold-independent way.
Assuming episode time is normalized to [0, 1], the penalized detection time ($\text{PDT}$) for a failed episode $e$ evaluated at score threshold $\tau$ is
\begin{equation}\text{PDT}_e(\tau) = \begin{cases}
t  &\text{if $t \in [0, 1)$ is the first timestep with $s^t(e) \geq \tau$} \\
1  &\text{if failure is not detected at threshold $\tau$},
\end{cases}
\end{equation}
where $s^t(e)$ is the detector's score at time $t$ on episode $e$. 
We assign missed failures a maximal detection time, assuming that failure can be automatically detected at the episode horizon.
The PDT measures the latency reduction relative to this automatic baseline.
For an episode dataset $\mathcal{D}$, let $\mathcal{D}_f \subset \mathcal{D}$ denote the subset of failed episodes.
The expected detection time for a given threshold is then the average across all failures: $\text{PDT}_{\mathcal{D}}(\tau) = \frac{1}{|\mathcal{D}_f|} \sum_{i=1}^{|\mathcal{D}_f|} \text{PDT}_{e_i}(\tau)$.

To compute the AUCPDT, we compute precision and $\text{PDT}_\mathcal{D}$ at each unique score threshold and retain only Pareto-optimal thresholds.
This includes all thresholds that are Pareto-optimal in terms of precision and recall, but adds points at identical recall with different trade-offs between precision and PDT.
The area under this function yields a single, bounded, threshold-independent metric, where lower values indicate more precise and earlier detectors. 
AUCPDT does not require a specific "point of failure" timestep, selection of which requires substantial annotation work and is fundamentally arbitrary when policies are allowed to retry a failed subtask, such as in our experiments.
For further details, see \Cref{app:pdt-integration} and the exemplary PR and PDT plots in \Cref{fig:pr-pdt-xvla-block} and \Cref{fig:pr-pdt-pi-block}.

\begin{table}[h!]
    \caption{Threshold-independent detector performance on \textbf{Libero-Plus}. PR and PDT (see~\Cref{sec:eval:metrics}) refer to the areas under the respective curves. The top three performing methods are highlighted. FAIL achieves performance comparable to baseline methods, while incurring no computational overhead. Colors mark the \colorbox{highlight_0}{best}, \colorbox{highlight_1}{second-best}, and \colorbox{highlight_2}{third-best} methods.}
    \label{tab:eval-libero}

    \centering
    \begin{tabular}{l l l *{3}{S[table-format=1.2, table-space-text-post={\ci{0.00}{0.00}}]} | *{3}{S[table-format=1.2, table-space-text-post={\ci{0.00}{0.00}}]}}
        \toprule
         & & & {ACE} & {Diff} & {STAC} & {ACC} & {LLMD} & {FAIL} \\
        \midrule

\multirow{4}{*}{\texttt{Object}} & \multirow{2}{*}{X-VLA} & PR\higherBetter & 0.89{\ci{0.89}{0.90}} & 0.92{\ci{0.91}{0.92}} & \cellcolor{highlight_0} 1.00{\ci{0.99}{1.00}} & \cellcolor{highlight_1} 0.97{\ci{0.97}{0.98}} & 0.90{\ci{0.89}{0.90}} & \cellcolor{highlight_2} 0.94{\ci{0.94}{0.95}} \\
& & PDT\lowerBetter & \cellcolor{highlight_0} 0.12{\ci{0.12}{0.13}} & \cellcolor{highlight_0} 0.12{\ci{0.11}{0.13}} & 0.31{\ci{0.31}{0.31}} & 0.24{\ci{0.23}{0.24}} & \cellcolor{highlight_2} 0.15{\ci{0.15}{0.15}} & \cellcolor{highlight_1} 0.14{\ci{0.14}{0.14}} \\

 & \multirow{2}{*}{$\pi_{0.5}$} & PR\higherBetter & \cellcolor{highlight_1} 0.60{\ci{0.59}{0.61}} & 0.46{\ci{0.43}{0.49}} & 0.40{\ci{0.39}{0.41}} & \cellcolor{highlight_0} 0.66{\ci{0.64}{0.67}} & 0.43{\ci{0.42}{0.44}} & \cellcolor{highlight_2} 0.54{\ci{0.53}{0.55}} \\
& & PDT\lowerBetter & \cellcolor{highlight_0} 0.51{\ci{0.51}{0.51}} & 0.67{\ci{0.66}{0.69}} & 0.73{\ci{0.73}{0.74}} & \cellcolor{highlight_1} 0.55{\ci{0.54}{0.57}} & 0.66{\ci{0.65}{0.66}} & \cellcolor{highlight_2} 0.60{\ci{0.59}{0.60}} \\

\midrule

\multirow{4}{*}{\texttt{Spatial}} & \multirow{2}{*}{X-VLA} & PR\higherBetter & 0.84{\ci{0.83}{0.84}} & 0.82{\ci{0.81}{0.83}} & \cellcolor{highlight_0} 0.99{\ci{0.99}{0.99}} & \cellcolor{highlight_1} 0.94{\ci{0.94}{0.94}} & 0.85{\ci{0.85}{0.85}} & \cellcolor{highlight_2} 0.92{\ci{0.92}{0.93}} \\
& & PDT\lowerBetter & \cellcolor{highlight_1} 0.19{\ci{0.19}{0.20}} & 0.23{\ci{0.22}{0.24}} & 0.36{\ci{0.36}{0.36}} & 0.31{\ci{0.30}{0.31}} & \cellcolor{highlight_2} 0.20{\ci{0.20}{0.20}} & \cellcolor{highlight_0} 0.18{\ci{0.18}{0.18}} \\

 & \multirow{2}{*}{$\pi_{0.5}$} & PR\higherBetter & \cellcolor{highlight_2} 0.74{\ci{0.72}{0.77}} & 0.64{\ci{0.59}{0.68}} & 0.45{\ci{0.39}{0.51}} & \cellcolor{highlight_0} 0.78{\ci{0.77}{0.79}} & 0.67{\ci{0.65}{0.68}} & \cellcolor{highlight_1} 0.76{\ci{0.73}{0.79}} \\
& & PDT\lowerBetter & \cellcolor{highlight_0} 0.45{\ci{0.43}{0.46}} & 0.59{\ci{0.55}{0.62}} & 0.72{\ci{0.69}{0.74}} & \cellcolor{highlight_1} 0.50{\ci{0.49}{0.51}} & 0.57{\ci{0.56}{0.57}} & \cellcolor{highlight_2} 0.52{\ci{0.52}{0.52}} \\

\midrule

\multirow{4}{*}{\texttt{Goal}} & \multirow{2}{*}{X-VLA} & PR\higherBetter & 0.81{\ci{0.81}{0.81}} & 0.65{\ci{0.64}{0.65}} & \cellcolor{highlight_0} 1.00{\ci{0.99}{1.00}} & \cellcolor{highlight_1} 0.96{\ci{0.96}{0.96}} & 0.87{\ci{0.87}{0.88}} & \cellcolor{highlight_2} 0.94{\ci{0.93}{0.95}} \\
& & PDT\lowerBetter & \cellcolor{highlight_1} 0.22{\ci{0.22}{0.22}} & 0.38{\ci{0.37}{0.38}} & 0.30{\ci{0.29}{0.30}} & \cellcolor{highlight_2} 0.28{\ci{0.27}{0.28}} & \cellcolor{highlight_1} 0.22{\ci{0.22}{0.22}} & \cellcolor{highlight_0} 0.20{\ci{0.19}{0.20}} \\

 & \multirow{2}{*}{$\pi_{0.5}$} & PR\higherBetter & \cellcolor{highlight_0} 0.84{\ci{0.82}{0.86}} & 0.74{\ci{0.73}{0.76}} & 0.50{\ci{0.48}{0.52}} & 0.77{\ci{0.76}{0.79}} & \cellcolor{highlight_2} 0.78{\ci{0.75}{0.80}} & \cellcolor{highlight_1} 0.79{\ci{0.78}{0.81}} \\
& & PDT\lowerBetter & \cellcolor{highlight_0} 0.33{\ci{0.31}{0.35}} & 0.46{\ci{0.45}{0.47}} & 0.60{\ci{0.59}{0.62}} & \cellcolor{highlight_2} 0.41{\ci{0.39}{0.42}} & 0.42{\ci{0.40}{0.43}} & \cellcolor{highlight_1} 0.39{\ci{0.38}{0.40}} \\

\midrule

\multirow{4}{*}{\texttt{10}} & \multirow{2}{*}{X-VLA} & PR\higherBetter & 0.75{\ci{0.74}{0.76}} & 0.77{\ci{0.76}{0.77}} & \cellcolor{highlight_0} 1.00{\ci{1.00}{1.00}} & \cellcolor{highlight_1} 0.96{\ci{0.96}{0.96}} & 0.86{\ci{0.85}{0.88}} & \cellcolor{highlight_2} 0.94{\ci{0.93}{0.95}} \\
& & PDT\lowerBetter & 0.28{\ci{0.28}{0.29}} & 0.28{\ci{0.28}{0.29}} & 0.30{\ci{0.29}{0.31}} & \cellcolor{highlight_2} 0.26{\ci{0.25}{0.26}} & \cellcolor{highlight_1} 0.22{\ci{0.21}{0.24}} & \cellcolor{highlight_0} 0.19{\ci{0.18}{0.20}} \\

 & \multirow{2}{*}{$\pi_{0.5}$} & PR\higherBetter & \cellcolor{highlight_2} 0.75{\ci{0.72}{0.79}} & 0.70{\ci{0.68}{0.72}} & 0.46{\ci{0.45}{0.48}} & \cellcolor{highlight_0} 0.80{\ci{0.79}{0.82}} & 0.70{\ci{0.68}{0.73}} & \cellcolor{highlight_1} 0.79{\ci{0.77}{0.81}} \\
& & PDT\lowerBetter & \cellcolor{highlight_1} 0.42{\ci{0.38}{0.45}} & 0.49{\ci{0.47}{0.51}} & 0.62{\ci{0.60}{0.64}} & \cellcolor{highlight_0} 0.39{\ci{0.38}{0.41}} & \cellcolor{highlight_2} 0.43{\ci{0.42}{0.45}} & \cellcolor{highlight_0} 0.39{\ci{0.37}{0.40}} \\

\midrule
\midrule

\multirow{4}{*}{Average} & \multirow{2}{*}{X-VLA} & PR\higherBetter & 0.82{\ci{0.76}{0.88}} & 0.79{\ci{0.68}{0.90}} & \cellcolor{highlight_0} 1.00{\ci{0.99}{1.00}} & \cellcolor{highlight_1} 0.96{\ci{0.94}{0.97}} & 0.87{\ci{0.85}{0.89}} & \cellcolor{highlight_2} 0.94{\ci{0.93}{0.95}} \\
& & PDT\lowerBetter & \cellcolor{highlight_1} 0.20{\ci{0.14}{0.27}} & \cellcolor{highlight_2} 0.25{\ci{0.15}{0.36}} & 0.31{\ci{0.29}{0.34}} & 0.27{\ci{0.24}{0.30}} & \cellcolor{highlight_1} 0.20{\ci{0.17}{0.23}} & \cellcolor{highlight_0} 0.18{\ci{0.15}{0.20}} \\

 & \multirow{2}{*}{$\pi_{0.5}$} & PR\higherBetter & \cellcolor{highlight_1} 0.73{\ci{0.64}{0.83}} & 0.63{\ci{0.51}{0.76}} & 0.45{\ci{0.41}{0.49}} & \cellcolor{highlight_0} 0.75{\ci{0.69}{0.82}} & 0.64{\ci{0.49}{0.79}} & \cellcolor{highlight_2} 0.72{\ci{0.60}{0.84}} \\
& & PDT\lowerBetter & \cellcolor{highlight_0} 0.43{\ci{0.35}{0.50}} & 0.55{\ci{0.46}{0.65}} & 0.67{\ci{0.60}{0.73}} & \cellcolor{highlight_1} 0.46{\ci{0.39}{0.54}} & 0.52{\ci{0.41}{0.63}} & \cellcolor{highlight_2} 0.47{\ci{0.37}{0.57}} \\

\bottomrule
    \end{tabular}
\end{table}

\section{Results and Discussion}
\label{sec:discussion}
\Cref{tab:eval-real} and \Cref{tab:eval-libero} summarize the threshold-independent results for the real-world tasks and Libero-Plus, respectively.
We refer to \Cref{sec:thresh-dependent} for threshold-dependent results.

\textbf{FAIL Reliably Detects Failures Early.}
Overall, we find that FAIL achieves failure detection performance comparable to, and often better than, significantly more expensive baseline methods across simulated and real-world tasks.
FAIL is the only method to place within the top-three best performing methods on nearly all real-world and simulated tasks.
Crucially, FAIL achieves this with only marginal computational overhead, thereby enabling real-time deployment.

\textbf{LLMD and ACC Capture Distinct Failure Modes.}
While we find that LLMD and ACC perform well on individual tasks, only their combination, FAIL, is reliable across tasks.
ACC alone already achieves a high detection accuracy, while LLMD detects failures early, but less accurately.
FAIL frequently achieves the lowest overall AUCPDT, such as on the real-world \texttt{Cups}, \texttt{Stack T}, and \texttt{Mixer} tasks, particularly for $\pi_{0.5}$, and for X-VLA on the Libero-Plus task suites. 
Qualitatively, ACC performs well on tasks where the robot exhibits rapid, jerky motions in OOD situations, such as the real-world \texttt{Blocks} and \texttt{Kitchen} tasks and all Libero-Plus task suites.
LLMD tends to achieve higher detection rates than ACC on tasks where the policy repeatedly retries an action in an infinite loop or resorts to "default" actions that do not correspond to the current environment state.
Qualitative examples are shown in \Cref{appendix:qualitative_results}.

\textbf{Token-Wise LLMD Is Necessary.}
Using a single global feature mean and covariance matrix across all token positions captures significantly less useful information for OOD detection than per-token statistics, for example, increasing AUCPDT from $0.19$ to $0.24$ for X-VLA on the \texttt{Drawer} task (see \Cref{tab:llmd-token-ablation} for full results).
We hypothesize that this is because the model's internal behavior differs depending on how far it needs to predict into the future, where actions farther from the current timestep are, in general, harder to predict and therefore require distinct features.

\textbf{Mahalanobis Distance Achieves a Good Performance-Complexity Tradeoff}
In \Cref{tab:eval-llmd-alternatives}, we compare the token-wise Mahalanobis distance to the distance metrics from \citet{romer2026fiper} (RND) and \citet{xu2025faildetect} (logpZO) using our deterministic prior-noise last-layer features.
Neither baseline outperforms LLMD on average in terms of AUCPDT, while requiring additional training during preprocessing and increasing implementation complexity.

\textbf{Velocity Normalization Improves ACC.}
On the simulated Libero-Plus benchmark, we find that velocity normalization is crucial for achieving high failure detection rates, for example, reducing AUCPDT from $0.38$ to $0.28$ for X-VLA on the \texttt{Spatial} suite (see \Cref{tab:acc-vel-norm-ablation-libero} for full results).
On real-world tasks, the difference is smaller, but it remains consistent across tasks, as shown in \Cref{tab:acc-vel-norm-ablation-real}.

\textbf{AUCPDT Highlights Differences in Detection Times.}
As expected, we find that AUCPDT correlates with AUCPR, as both penalize high false-negative and false-positive rates.
However, AUCPDT reveals that failure detectors with similar detection rates detect failures at different timesteps.
Typically, we find that LLMD detects failure earlier than baselines, including ACC, when comparing at similar AUCPR, such as with X-VLA on \texttt{Kitchen} or $\pi_{0.5}$ on \texttt{Stack T}.

\textbf{ACC and STAC.}
While ACC can be viewed as a velocity-normalized single-sample estimator of STAC, we find that ACC outperforms STAC on nearly all real-world tasks.
On \texttt{Libero-Plus}, STAC achieves better detection performance in terms of AUCPR, while ACC detects failures earlier, as the respective AUCPDT shows.
We hypothesize that ACC reduces false detections during mode selection, as ACC evaluates planning inconsistencies only with executed action chunks, not against counterfactual trajectories, which is beneficial when policies decide between action modes.

\textbf{VLA-FAIL is Fast}
As \Cref{tab:overhead} demonstrates, the runtime time and memory overhead of VLA-FAIL during action sampling is small.
These numbers were obtained with action sampling and feature extraction running in parallel on an NVIDIA RTX 5090 GPU.
The one-off preprocessing step of LLMD is dominated by feature extraction at \SI{60}{\minute} for $\pi_{0.5}$ and \SI{67}{\minute} for X-VLA on the \texttt{Stack-T} task.
Subsequent statistics computation takes only \SI{3}{\minute}.

\begin{table}[t]
    \caption{End-to-end runtime latency and memory overhead of VLA-FAIL on an NVIDIA RTX 5090. The overhead of VLA-FAIL is small.}
    \label{tab:overhead}
    \centering
    \begin{tabular}{c|r @{\hspace{6px}} r | r @{\hspace{6px}} r}
     & no FAIL & FAIL & no FAIL & FAIL \\
     \midrule
     X-VLA & \SI{136}{\milli\second}& \SI{139}{\milli\second} {\scriptsize(+2.2\%)} & \SI{3,717}{\mega\byte} & \SI{3,721}{\mega\byte} {\scriptsize(+0.1\%)} \\
     $\pi_{0.5}$ & \SI{77}{\milli\second} & \SI{78}{\milli\second} {\scriptsize(+1.3\%)} & \SI{7,136}{\mega\byte} & \SI{7,165}{\mega\byte} {\scriptsize(+0.4\%)}
\end{tabular}
\end{table}

\section{Limitations}
LLMD requires access to the finetuning data for a preprocessing pass, which may be expensive for large datasets or impossible when the data is unavailable.
ACC requires receding-horizon control with sufficient action overlap (\Cref{fig:acc-overlap-ablation-drawer}, \Cref{fig:acc-overlap-ablation-blocks}) and does not directly apply to fully open-loop chunk execution.
Due to its design, VLA-FAIL can miss failures that are consistent in features and actions, such as confidently stopping or ignoring a language instruction.
Here, a combination with VLM-based approaches that can detect task failure at the semantic level~\citep{AgiaSinhaEtAl2024,tan2025robo,liang2026robometer,lee2025roboreward} could be beneficial.
VLA-FAIL is empirically motivated and can, like all baseline methods, not provide any formal safety guarantees.
Real-world deployment in safety-critical environments, especially around humans, can therefore not solely rely on its failure predictions for safety.

\section{Conclusion}
We presented VLA-FAIL, a lightweight runtime failure detector for finetuned VLAs that requires no failure data.
FAIL combines LLMD, which measures token-wise deviations in last-layer features relative to the training data, with ACC, which measures velocity-normalized disagreement between overlapping action chunks.
Across real-world and simulated manipulation tasks, FAIL detects complementary failure modes with only marginal computational overhead, making VLA-FAIL a practical step toward human intervention pipelines and the safe real-world deployment of VLAs.
Future work could integrate VLA-FAIL into a closed-loop recovery system, where a human operator takes over when failure is detected, potentially as part of a larger DAgger \citep{ross2011dagger} loop.

\clearpage
\acknowledgments{
The authors would like to thank Emtiyaz Khan and Thomas Möllenhoff for their helpful discussions on uncertainty quantification, Bayesian methods, and out-of-distribution detection, as well as the valuable time spent at the RIKEN Center for Advanced Intelligence Project.
The authors would also like to thank Fabian Lübbe and Patrick Dormanns for their adaptations of $\pi_{0.5}$ and X-VLA to the real-robot infrastructure at ALR.

This work was supported by the European Research Council (ERC) under the European Union’s Horizon Europe programme through the project SMARTI\textsuperscript{3} (Grant Agreement No. 101171393).
This work has been supported by the German Federal Ministry of Research, Technology, and Space (BMFTR) under the Robotics Institute Germany (RIG).
The authors acknowledge support by the state of Baden-Württemberg through bwHPC, as well as the HoreKa supercomputer funded by the Ministry of Science, Research and the Arts Baden-Württemberg and by the German Federal Ministry of Education and Research.
The authors gratefully acknowledge the Gauss Centre for Supercomputing e.V. (www.gauss-centre.eu) for funding this project by providing computing time on the GCS Supercomputer JUWELS~\citep{JUWELS} at Jülich Supercomputing Centre (JSC).
}

\bibliography{references}  %

\clearpage
\appendix
\crefalias{section}{appendix}
\crefalias{subsection}{appendix}

\section{Penalized Detection Time}
\label{app:pdt-integration}
\FloatBarrier
Here, we provide further details on the AUCPDT metric introduced in \Cref{sec:eval:metrics}.

\paragraph{Pareto Front Computation}
Thresholds that are not Pareto-optimal are not relevant for score computation, as they do not represent values that an operator would choose.
As is common with the AUCPR metric, we remove these points before computing the area to smooth the PDT curve.
Note that there can be thresholds at the same recall but with different detection times and precisions, where lowering a score keeps the failure detection unchanged at the episode level but reduces the detection time within episodes, potentially introducing additional false positives.
As these thresholds represent relevant tradeoffs between precision and PDT, we compute the Pareto front in terms of precision and PDT, not precision and recall.

\paragraph{Area Computation}
Let $\tau_k$. $k=1, ..., K$ denote the unique Pareto-optimal thresholds sorted in ascending order.
We calculate the area under the penalized detection time curve over the precision interval from the failure rate to one using right-hand Riemann sums:
\begin{equation}
\text{AUCPDT} =\sum_{k=2}^{K} (P_k - P_{k-1}) PDT_k
\end{equation}
where $P_k$ and $PDT_k$ denote the precision and penalized detection time at threshold $\tau_k$, setting $P_K = 1$, as no detection is made at that threshold.
We only integrate from the failure rate, since lower-precision values are uninformative: They yield worse performance than a naive detector that always predicts failure at $t=0$.
Crucially, the integration domain is fixed independently of each detector's performance, leaving the PDT at each precision value as the only variable influencing the AUCPDT. 
We use step-wise integration to avoid optimistically underestimating detection times between measured points, as achieving a higher precision strictly requires accepting the PDT of the next available threshold.

\begin{figure}
    \centering
    \begin{subfigure}[t]{0.5\textwidth}
        \centering
        \includegraphics[width=\textwidth]{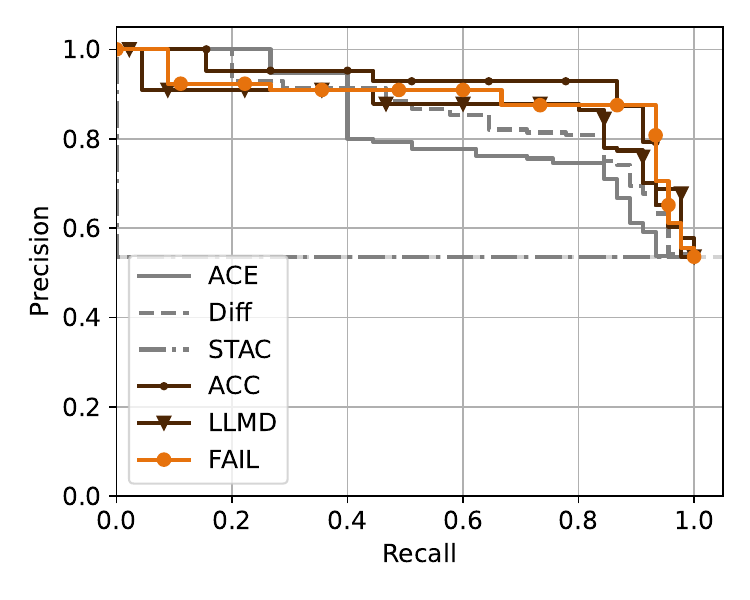}
        \caption{Precision-Recall}
    \end{subfigure}%
\hfill%
    \begin{subfigure}[t]{0.5\textwidth}
        \centering
        \includegraphics[width=\textwidth]{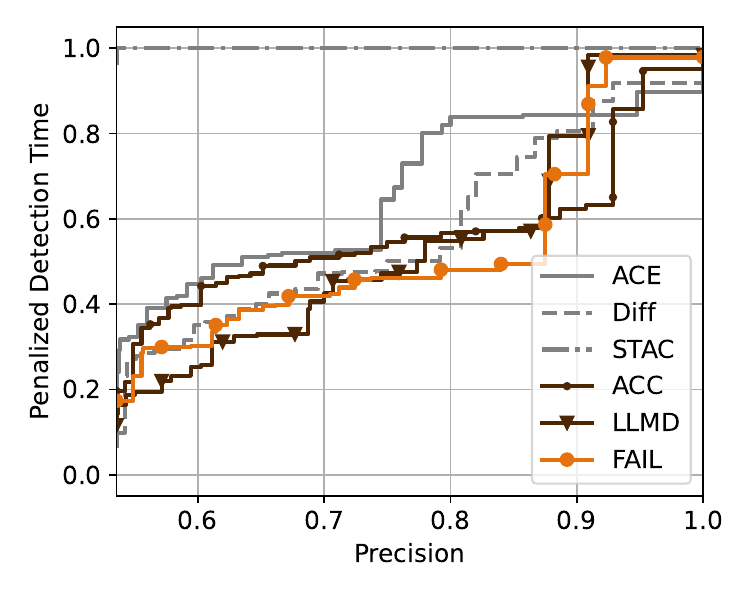}
        \caption{Penalized Detection Time vs. Precision}
    \end{subfigure}
    \caption{X-VLA on \texttt{Blocks}. Here, STAC demonstrates no useful failure signal, leading to a recall that stays constant at the fail rate, and a PDT that is constant at 1.}
    \label{fig:pr-pdt-xvla-block}
\end{figure}

\begin{figure}
    \centering
    \begin{subfigure}[t]{0.5\textwidth}
        \centering
        \includegraphics[width=\textwidth]{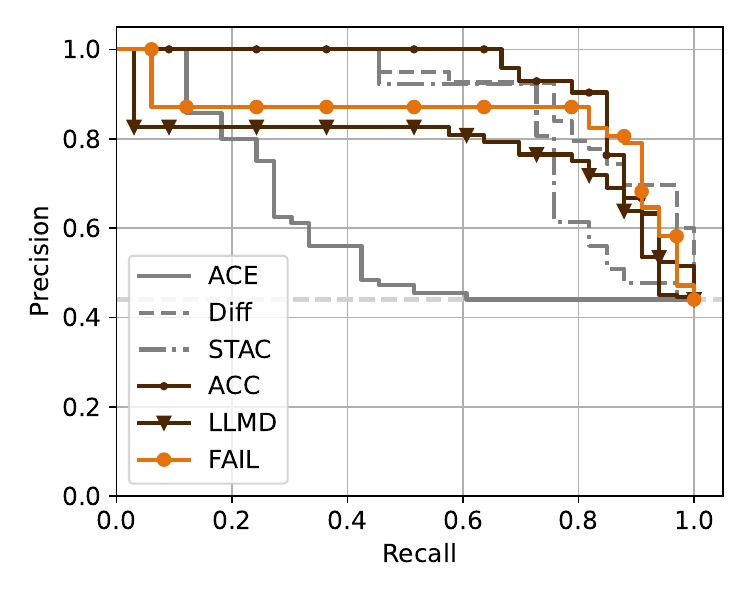}
        \caption{Precision-Recall}
    \end{subfigure}%
\hfill%
    \begin{subfigure}[t]{0.5\textwidth}
        \centering
        \includegraphics[width=\textwidth]{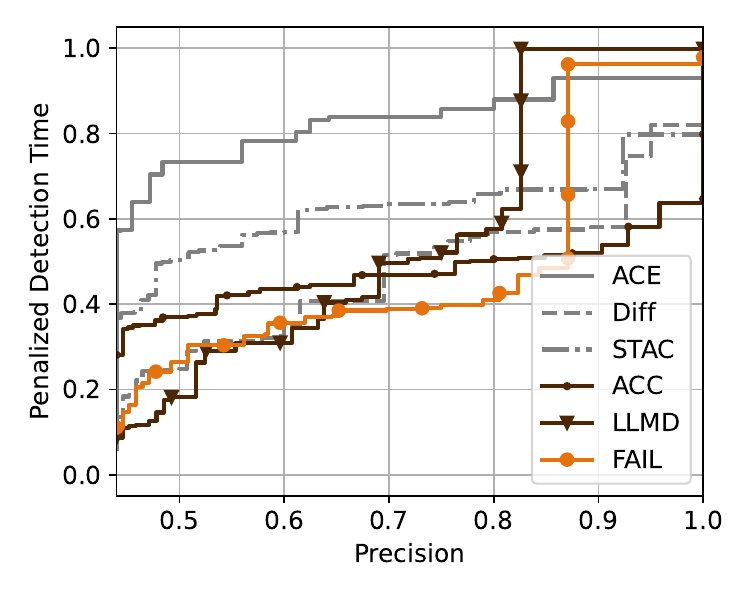}
        \caption{Penalized Detection Time vs. Precision}
    \end{subfigure}
    \caption{$\pi_{0.5}$ on \texttt{Blocks}. ACC performs substantially worse than other methods.}
    \label{fig:pr-pdt-pi-block}
\end{figure}

\paragraph{Exemplary Plots}
\Cref{fig:pr-pdt-xvla-block} and \Cref{fig:pr-pdt-pi-block} show examples of PR and PDT plots.
In both cases, ACC outperforms all other methods in terms of PR. i.e., pure detection rate.
However, it detects failures more slowly than LLMD and FAIL, particularly at low precision values, resulting in a higher AUCPDT.

\paragraph{Relationship to Prior Work}
Compared to the balanced accuracy-detection time plots from \citet{gu2026safe}, AUCPDT makes the trade-off between precision and detection time explicit by grouping recall and detection time, which are inversely correlated, on the same axis.
\citet{romer2026fiper} introduce the timestep-weighted accuracy as another metric that jointly measures detection performance and detection time.
However, this is a threshold-dependent metric.

\FloatBarrier
\section{Ablation Studies}

\paragraph{Token-Wise LLMD}
As \Cref{tab:llmd-token-ablation} demonstrates for the real-world tasks, computing feature statistics per token position improves failure detection rates significantly.
While this result is somewhat counterintuitive, as the same final linear layer projects all features to action, independent of token position, last-layer features accumulate values across all layers of the network, which the final linear layer is trained to discard to produce a clean action.
As such, there can be token-dependent differences in the features that do not translate to different actions.

\paragraph{Velocity Normalization in ACC}
In \Cref{tab:acc-vel-norm-ablation-real}, we ablate the velocity normalization in ACC on real-world tasks.
While the improvement is small, it is consistent across tasks.
On the simulated Libero-Plus benchmark, we see a clearer picture (\Cref{tab:acc-vel-norm-ablation-libero}), with ACC without velocity normalization performing significantly worse than velocity-normalized ACC.

\begin{figure}
    \centering
    \begin{subfigure}[t]{0.5\textwidth}
        \centering
        \includegraphics[width=\textwidth]{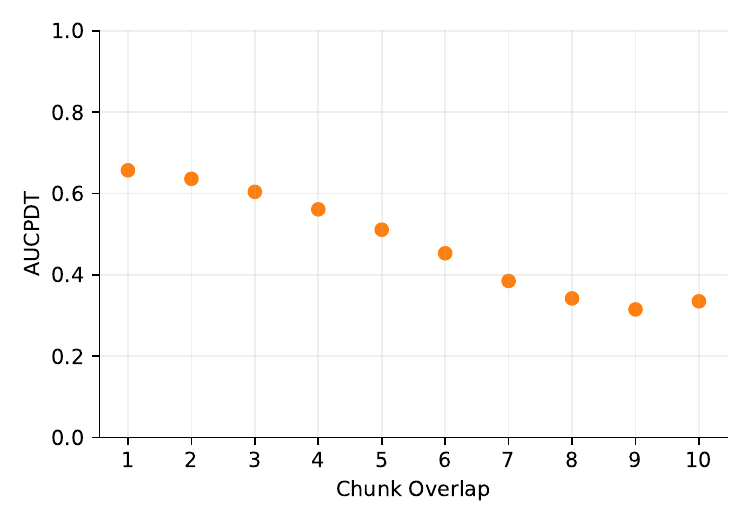}
        \caption{X-VLA}
    \end{subfigure}%
\hfill%
    \begin{subfigure}[t]{0.5\textwidth}
        \centering
        \includegraphics[width=\textwidth]{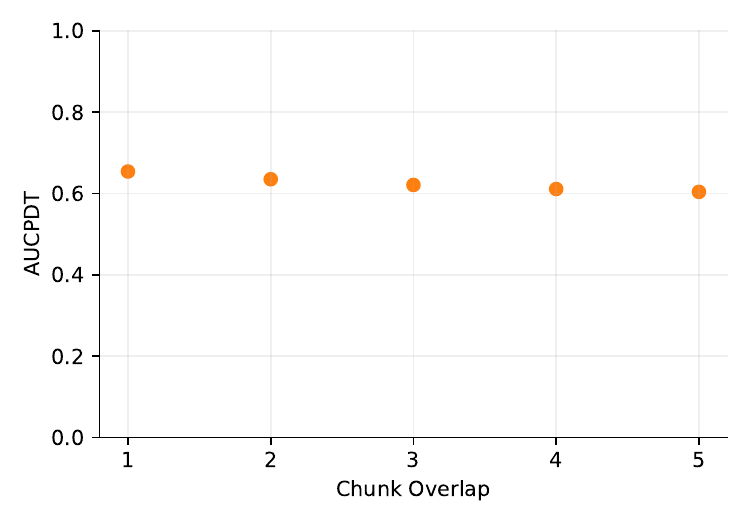}
        \caption{$\pi_{0.5}$}
    \end{subfigure}
    \caption{Dependence of ACC on the number of overlapping actions on the \texttt{Drawer} real-world task. A lower AUCPDT is better. As expected, a smaller chunk overlap reduces ACC's performance. However, even with a single overlapping action, ACC retains some failure detection capability.}
    \label{fig:acc-overlap-ablation-drawer}
\end{figure}

\begin{figure}
    \centering
    \begin{subfigure}[t]{0.5\textwidth}
        \centering
        \includegraphics[width=\textwidth]{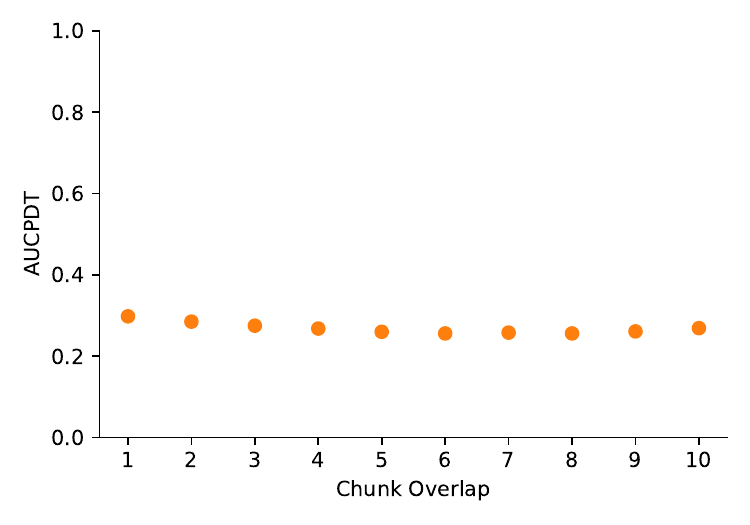}
        \caption{X-VLA}
    \end{subfigure}%
\hfill%
    \begin{subfigure}[t]{0.5\textwidth}
        \centering
        \includegraphics[width=\textwidth]{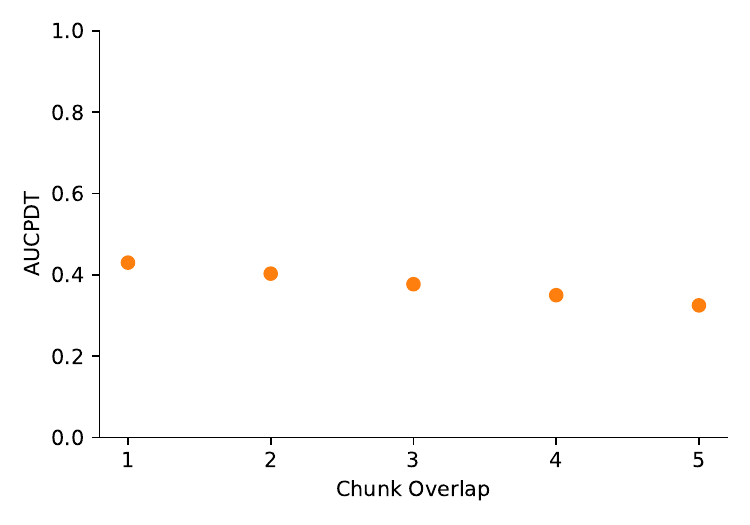}
        \caption{$\pi_{0.5}$}
    \end{subfigure}
    \caption{Dependence of ACC on the number of overlapping actions on the \texttt{Blocks} real-world task. A lower AUCPDT is better. As expected, a smaller chunk overlap reduces ACC's performance. However, even with a single overlapping action, ACC retains some failure detection capability.}
    \label{fig:acc-overlap-ablation-blocks}
\end{figure}

\paragraph{ACC Hyperparameters}
ACC requires overlapping actions between successive action chunks.
As expected, we find that ACC's performance degrades with a smaller overlap between chunks (\Cref{fig:acc-overlap-ablation-drawer}, \Cref{fig:acc-overlap-ablation-blocks}).
However, ACC retains some failure detection capability even for a single overlapping action.
Additionally, we find that ACC is more robust to shorter overlapping action sequences on $\pi_{0.5}$ than on X-VLA.

\begin{figure}
    \centering
    \begin{subfigure}[t]{0.5\textwidth}
        \centering
        \includegraphics[width=\textwidth]{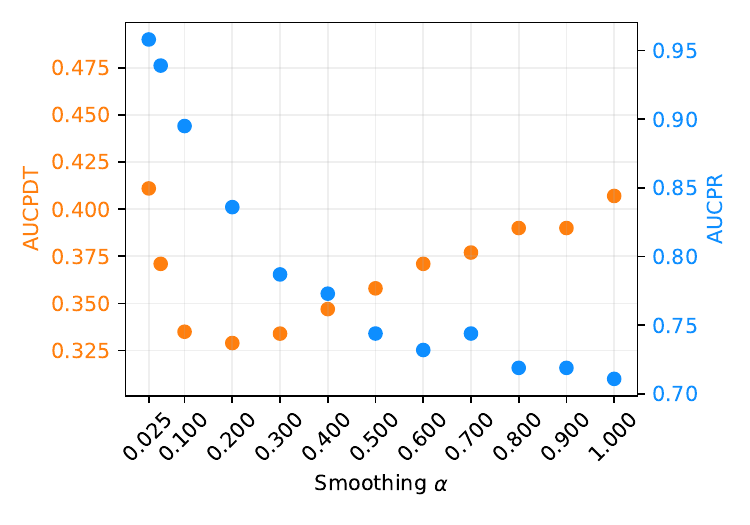}
        \caption{X-VLA}
    \end{subfigure}%
\hfill%
    \begin{subfigure}[t]{0.5\textwidth}
        \centering
        \includegraphics[width=\textwidth]{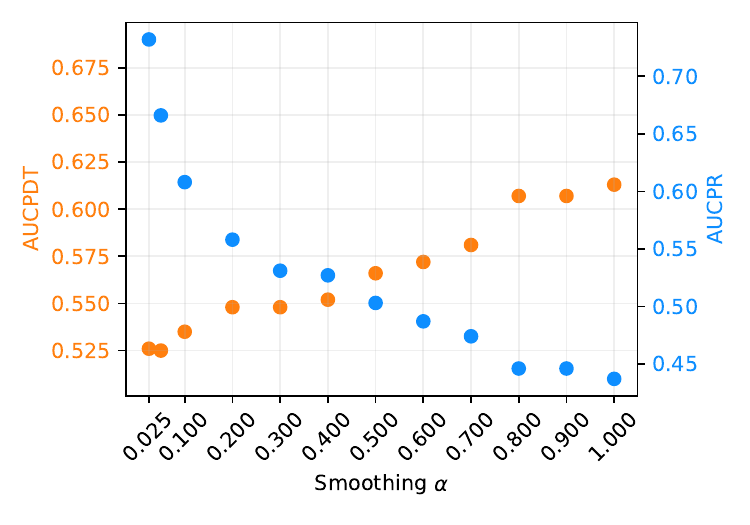}
        \caption{$\pi_{0.5}$}
    \end{subfigure}
    \caption{Dependence of ACC on the smoothing factor $\alpha$ on the \texttt{Drawer} real-world task. A lower AUCPDT is better, while for AUCPR, higher values are better. A lower value of $\alpha$, i.e., more smoothing, increases failure detection accuracy but tends to delay detection, leading to a favorable tradeoff at an intermediate value. We use $\alpha=0.1$ for all tasks.}
    \label{fig:acc-smoothing-ablation-drawer}
\end{figure}

The performance of ACC also changes with the smoothing factor $\alpha$, with lower values of $\alpha$ generally improving failure detection accuracy (AUCPR) but delaying failure detection (AUCPDT), as demonstrated in \Cref{fig:acc-smoothing-ablation-drawer}.
We use a fixed value of $\alpha=0.1$ across all experiments.

\begin{figure}
    \centering
    \begin{subfigure}[t]{0.5\textwidth}
        \centering
        \includegraphics[width=\textwidth]{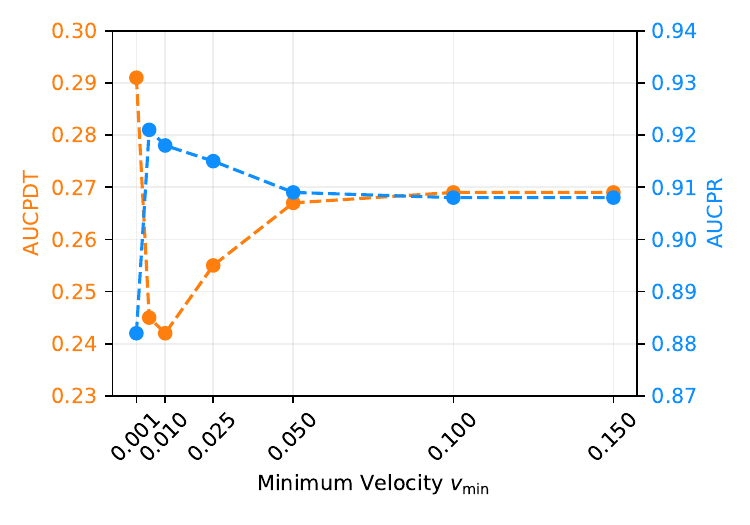}
        \caption{X-VLA}
    \end{subfigure}%
\hfill%
    \begin{subfigure}[t]{0.5\textwidth}
        \centering
        \includegraphics[width=\textwidth]{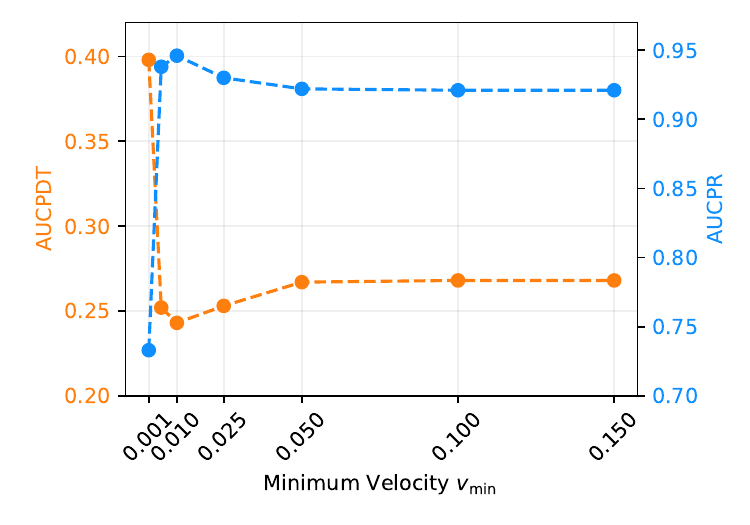}
        \caption{$\pi_{0.5}$}
    \end{subfigure}
    \caption{Dependence of ACC on the minimum velocity $v_\text{min}$ on the \texttt{Blocks} real-world task. A lower AUCPDT is better, while for AUCPR, higher values are better. Lower minimum velocities tend to work better, as they enable more accurate velocity scaling. However, there is a sharp drop at values smaller than $0.01$, where the velocity scaling starts amplifying noise. We use $v_\text{min}=0.01$ for all real-world tasks.}
    \label{fig:acc-min-vel-ablation-blocks}
\end{figure}

ACC's detection performance generally improves with smaller minimum velocities $v_\text{min}$ (\Cref{fig:acc-min-vel-ablation-blocks}).
However, on the real-world tasks, values below $0.01$ lead to a sharp drop in performance, as the velocity scaling starts amplifying noise for a stationary robot.

\begin{table}
    \caption{Global vs. token-wise LLMD on the real-world tasks. Global feature statistics perform significantly worse than per-token statistics.}
    \label{tab:llmd-token-ablation}

    \centering
    \begin{tabular}{c c c S[table-format=1.2, table-space-text-post={\ci{0.00}{0.00}}] S[table-format=1.2, table-space-text-post={\ci{0.00}{0.00}}]}
        \toprule
         & & & {Global} & {Token-Wise} \\
         \midrule
         \multirow{4}{*}{\texttt{Kitchen}} & \multirow{2}{*}{X-VLA} & PR\higherBetter & 0.67{\ci{0.67}{0.67}} & \cellcolor{highlight_0} 1.00{\ci{1.00}{1.00}} \\
         & & PDT\lowerBetter & 0.42{\ci{0.42}{0.42}} & \cellcolor{highlight_0} 0.20{\ci{0.20}{0.20}} \\
         & \multirow{2}{*}{$\pi_{0.5}$} & PR\higherBetter & 0.57{\ci{0.48}{0.65}} & \cellcolor{highlight_0} 0.62{\ci{0.41}{0.82}} \\
         & & PDT\lowerBetter & 0.39{\ci{0.32}{0.50}} & \cellcolor{highlight_0} 0.33{\ci{0.20}{0.46}} \\
         \midrule
         \multirow{4}{*}{\texttt{Drawer}} & \multirow{2}{*}{X-VLA} & PR\higherBetter & 0.86{\ci{0.85}{0.86}} & \cellcolor{highlight_0} 0.90{\ci{0.89}{0.91}} \\
         & & PDT\lowerBetter & 0.24{\ci{0.23}{0.25}} & \cellcolor{highlight_0} 0.19{\ci{0.19}{0.19}} \\
         & \multirow{2}{*}{$\pi_{0.5}$} & PR\higherBetter & 0.43{\ci{0.43}{0.43}} & \cellcolor{highlight_0} 0.69{\ci{0.66}{0.72}} \\
         & & PDT\lowerBetter & 0.63{\ci{0.62}{0.63}} & \cellcolor{highlight_0} 0.45{\ci{0.44}{0.46}}\\
         \midrule
         \multirow{4}{*}{\texttt{Blocks}} & \multirow{2}{*}{X-VLA} & PR\higherBetter & 0.67{\ci{0.66}{0.68}} & \cellcolor{highlight_0} 0.83{\ci{0.82}{0.84}} \\
         & & PDT\lowerBetter & 0.36{\ci{0.35}{0.37}} & \cellcolor{highlight_0} 0.26{\ci{0.25}{0.26}} \\
         & \multirow{2}{*}{$\pi_{0.5}$} & PR\higherBetter & 0.70{\ci{0.68}{0.72}} & \cellcolor{highlight_0} 0.73{\ci{0.72}{0.74}} \\
         & & PDT\lowerBetter & 0.37{\ci{0.36}{0.38}} & \cellcolor{highlight_0} 0.31{\ci{0.31}{0.32}}\\
         \midrule
         \multirow{4}{*}{\texttt{Cups}} & \multirow{2}{*}{X-VLA} & PR\higherBetter & 0.67{\ci{0.66}{0.67}} & \cellcolor{highlight_0} 0.78{\ci{0.78}{0.79}} \\
         & & PDT\lowerBetter & 0.42{\ci{0.41}{0.42}} & \cellcolor{highlight_0} 0.33{\ci{0.33}{0.34}} \\
         & \multirow{2}{*}{$\pi_{0.5}$} & PR\higherBetter & 0.67{\ci{0.66}{0.68}} & \cellcolor{highlight_0} 0.78{\ci{0.78}{0.79}} \\
         & & PDT\lowerBetter & 0.42{\ci{0.41}{0.42}} & \cellcolor{highlight_0} 0.28{\ci{0.28}{0.29}} \\
         \midrule
         \multirow{4}{*}{\texttt{Stack T}} & \multirow{2}{*}{X-VLA} & PR\higherBetter & 0.84{\ci{0.83}{0.84}} & \cellcolor{highlight_0} 0.97{\ci{0.97}{0.97}} \\
         & & PDT\lowerBetter & 0.18{\ci{0.18}{0.19}} & \cellcolor{highlight_0} 0.08{\ci{0.08}{0.08}} \\
         & \multirow{2}{*}{$\pi_{0.5}$} & PR\higherBetter & 0.92{\ci{0.92}{0.92}} & \cellcolor{highlight_0} 0.95{\ci{0.95}{0.96}} \\
         & & PDT\lowerBetter & 0.14{\ci{0.14}{0.14}} & \cellcolor{highlight_0} 0.10{\ci{0.10}{0.10}} \\
         \midrule
         \multirow{4}{*}{\texttt{Mixer}} & \multirow{2}{*}{X-VLA} & PR\higherBetter & 0.74{\ci{0.72}{0.75}} & \cellcolor{highlight_0} 0.98{\ci{0.98}{0.98}} \\
         & & PDT\lowerBetter & 0.28{\ci{0.28}{0.29}} & \cellcolor{highlight_0} 0.11{\ci{0.11}{0.11}} \\
         & \multirow{2}{*}{$\pi_{0.5}$} & PR\higherBetter & 0.54{\ci{0.52}{0.56}} & \cellcolor{highlight_0} 0.77{\ci{0.76}{0.78}} \\
         & & PDT\lowerBetter & 0.48{\ci{0.46}{0.50}}  & \cellcolor{highlight_0} 0.28{\ci{0.27}{0.28}} \\
         \midrule
    \end{tabular}
\end{table}

\begin{table}
    \caption{ACC with and without velocity normalization on real-world tasks. Velocity normalization improves detection rates and detection time on most tasks.}
    \label{tab:acc-vel-norm-ablation-real}

    \centering
    \begin{tabular}{c c c S[table-format=1.2] S[table-format=1.2]}
        \toprule
         & & & {No Vel. Norm.} & {Vel. Norm.} \\
         \midrule
         \multirow{4}{*}{\texttt{Blocks}} & \multirow{2}{*}{X-VLA} & PR\higherBetter & 0.91 & \cellcolor{highlight_0} 0.92 \\
         & & PDT\lowerBetter & 0.27 & \cellcolor{highlight_0} 0.24 \\
         & \multirow{2}{*}{$\pi_{0.5}$} & PR\higherBetter & 0.92 & \cellcolor{highlight_0} 0.95 \\
         & & PDT\lowerBetter & 0.27 & \cellcolor{highlight_0} 0.24 \\
         \midrule
         \multirow{4}{*}{\texttt{Drawer}} & \multirow{2}{*}{X-VLA} & PR\higherBetter & \cellcolor{highlight_0} 0.90 & 0.83 \\
         & & PDT\lowerBetter & \cellcolor{highlight_0} 0.33 & \cellcolor{highlight_0} 0.33  \\
         & \multirow{2}{*}{$\pi_{0.5}$} & PR\higherBetter & 0.61 & \cellcolor{highlight_0} 0.62 \\
         & & PDT\lowerBetter & 0.54 & \cellcolor{highlight_0} 0.53 \\
         \midrule
         \multirow{4}{*}{\texttt{Cups}} & \multirow{2}{*}{X-VLA} & PR\higherBetter & 0.96 & \cellcolor{highlight_0} 0.97 \\
         & & PDT\lowerBetter & 0.29 & \cellcolor{highlight_0} 0.28 \\
         & \multirow{2}{*}{$\pi_{0.5}$} & PR\higherBetter & 0.54 & \cellcolor{highlight_0} 0.60 \\
         & & PDT\lowerBetter & 0.55 & \cellcolor{highlight_0} 0.51 \\
         \midrule
         \multirow{4}{*}{\texttt{Kitchen}} & \multirow{2}{*}{X-VLA} & PR\higherBetter &\cellcolor{highlight_0}  1.00 & \cellcolor{highlight_0} 1.00 \\
         & & PDT\lowerBetter & 0.24 & \cellcolor{highlight_0} 0.20 \\
         & \multirow{2}{*}{$\pi_{0.5}$} & PR\higherBetter & \cellcolor{highlight_0} 0.99 & \cellcolor{highlight_0} 0.99 \\
         & & PDT\lowerBetter & \cellcolor{highlight_0} 0.17 & \cellcolor{highlight_0} 0.17 \\
         \midrule
         \multirow{4}{*}{\texttt{Stack T}} & \multirow{2}{*}{X-VLA} & PR\higherBetter & 0.97 & \cellcolor{highlight_0} 0.98 \\
         & & PDT\lowerBetter & \cellcolor{highlight_0} 0.09 & \cellcolor{highlight_0} 0.09 \\
         & \multirow{2}{*}{$\pi_{0.5}$} & PR\higherBetter & 0.91 & \cellcolor{highlight_0} 0.95 \\
         & & PDT\lowerBetter & 0.14 & \cellcolor{highlight_0} 0.12 \\
         \midrule
         \multirow{4}{*}{\texttt{Mixer}} & \multirow{2}{*}{X-VLA} & PR\higherBetter & 0.96 & \cellcolor{highlight_0} 0.98 \\
         & & PDT\lowerBetter & \cellcolor{highlight_0} 0.15 & \cellcolor{highlight_0} 0.11 \\
         & \multirow{2}{*}{$\pi_{0.5}$} & PR\higherBetter & 0.82 & \cellcolor{highlight_0} 0.85 \\
         & & PDT\lowerBetter & 0.29 & \cellcolor{highlight_0} 0.28 \\
         \bottomrule
    \end{tabular}
\end{table}

\begin{table}
    \caption{ACC with and without velocity normalization on \texttt{Libero-Plus}. Velocity normalization consistently improves detection rates and detection time across most task suites.}
    \label{tab:acc-vel-norm-ablation-libero}

    \centering
    \begin{tabular}{c c c S[table-format=1.2] S[table-format=1.2]}
        \toprule
         & & & {No Vel. Norm.} & {Vel. Norm.} \\
         \midrule
         \multirow{4}{*}{\texttt{Object}} & \multirow{2}{*}{X-VLA} & PR\higherBetter & 0.96 & \cellcolor{highlight_0} 0.97 \\
         & & PDT\lowerBetter & 0.23 & \cellcolor{highlight_0} 0.22 \\
         & \multirow{2}{*}{$\pi_{0.5}$} & PR\higherBetter & \cellcolor{highlight_0} 0.68 & 0.65 \\
         & & PDT\lowerBetter & \cellcolor{highlight_0} 0.52 & 0.58 \\
         \midrule
         \multirow{4}{*}{\texttt{Spatial}} & \multirow{2}{*}{X-VLA} & PR\higherBetter & 0.86 & \cellcolor{highlight_0} 0.95 \\
         & & PDT\lowerBetter & 0.38 & \cellcolor{highlight_0} 0.28 \\
         & \multirow{2}{*}{$\pi_{0.5}$} & PR\higherBetter & 0.77 & \cellcolor{highlight_0} 0.80 \\
         & & PDT\lowerBetter & 0.48 & \cellcolor{highlight_0} 0.49 \\
         \midrule
         \multirow{4}{*}{\texttt{Goal}} & \multirow{2}{*}{X-VLA} & PR\higherBetter & 0.89 & \cellcolor{highlight_0} 0.95 \\
         & & PDT\lowerBetter & 0.32 & \cellcolor{highlight_0} 0.26 \\
         & \multirow{2}{*}{$\pi_{0.5}$} & PR\higherBetter & 0.86 & \cellcolor{highlight_0} 0.91 \\
         & & PDT\lowerBetter & 0.35 & \cellcolor{highlight_0} 0.33 \\
         \midrule
         \multirow{4}{*}{\texttt{10}} & \multirow{2}{*}{X-VLA} & PR\higherBetter & 0.75 & \cellcolor{highlight_0} 0.96 \\
         & & PDT\lowerBetter & 0.39 & \cellcolor{highlight_0} 0.23 \\
         & \multirow{2}{*}{$\pi_{0.5}$} & PR\higherBetter & 0.77 & \cellcolor{highlight_0} 0.79 \\
         & & PDT\lowerBetter & 0.42 & \cellcolor{highlight_0} 0.40 \\
         \midrule
    \end{tabular}
\end{table}

\FloatBarrier
\section{Additional Baselines}

\subsection{Alternatives to Mahalanobis}
\label{app:baselines:maha}
Here, we compare alternatives to the Mahalanobis distance for computing the typicality of last-layer features.
We apply all alternatives to the last-layer features obtained via our deterministic prior noise feature extraction scheme (\cref{sec:method:llmd}).
Random network distillation \citep{burda2018exploration} was used by \citet{romer2026fiper} on observation embeddings to detect task failure.
LogpZO \citep{xu2025faildetect} trains a flow-matching model on observation embeddings.
At runtime, it maps embeddings back to the corresponding prior noise and uses the norm of the noise vector to evaluate how likely the feature is under the prior.
Both methods apply equally well to our last-layer features.
We call them LL-RND and LL-logpZO.

We use the hyperparameters published by \citet{romer2026fiper} and \citet{xu2025faildetect}, adapting only the neural network's input size to our feature dimension.
For LL-logpZO, we train the convolutional U-Net on the sequence of action tokens, thus learning the joint distribution over tokens.
LL-RND is applied per token, using one predictor and one target network across all sequence positions.
We use the mean MSE over all positions as the failure score.

\Cref{tab:eval-llmd-alternatives} shows the performance of LL-RND and LL-logpZO on the real-world tasks.
For computational reasons, we evaluate only a single seed for these baselines, except on \texttt{Kitchen}, where we evaluate three seeds due to high variance in last-layer features visible in LLMD.
While both baselines outperform LLMD on individual tasks, LLMD performs equally well on average in terms of AUCPDT.
In particular, LL-logpZO shows inconsistent AUCPDT and AUCPR across tasks, while LL-RND performs similarly to LLMD on most tasks.
However, both baselines require an additional training step, increasing preprocessing time on an NVIDIA RTX 5090 GPU from \SI{70}{\minute} for LLMD to \SI{153}{\minute} for LL-RND and \SI{183}{\minute} for LL-logpZO.
Overall, we recommend the token-wise Mahalanobis distance as a good default distance metric, as it requires no additional training and is substantially simpler to implement.

Interestingly, the baselines benefit substantially from our novel feature choice: On $\pi_{0.5}$, random-noise features from all timesteps worsen LL-RND's AUCPR from $0.73$ to $0.49$ and AUCPDT from $0.35$ to $0.51$.
This highlights that our feature construction transfers to other methods that depend on stable last-layer features.

\begin{table}[ht]
    \caption{Threshold-independent detector performance of additional baselines on the \textbf{real-world tasks}. PR and PDT (see~\Cref{sec:eval:metrics}) refer to the areas under the respective curves, and the small numbers indicate $95\%$ confidence intervals. The \colorbox{highlight_0}{best} method is highlighted. While LL-RND and LL-lopZO outperform LLMD on some tasks, LLMD performs on average on par with both baselines in terms of PDT, while being significantly simpler to implement and requiring no additional training.}
    \label{tab:eval-llmd-alternatives}
    \centering
    \begin{tabular}{l l @{\hspace{4pt}} l *{2}{S[table-format=1.2]} | *{1}{S[table-format=1.3, table-space-text-post={\ci{0.00}{0.00}}]}}
        \toprule
         & & & {LL-RND} & {LL-logpZO} & {LLMD} \\
        \midrule

\multirow{4}{*}{\texttt{Blocks}} & \multirow{2}{*}{X-VLA} & PR\higherBetter & 0.85 & \cellcolor{highlight_0} 0.87 & 0.83{\ci{0.82}{0.84}} \\
& & PDT\lowerBetter & 0.27 & \cellcolor{highlight_0} 0.25 & 0.26{\ci{0.25}{0.26}} \\
& \multirow{2}{*}{$\pi_{0.5}$} & PR\higherBetter & 0.71 & \cellcolor{highlight_0} 0.88 & 0.73{\ci{0.72}{0.74}} \\
& & PDT\lowerBetter & 0.35 & \cellcolor{highlight_0} 0.26 & 0.31{\ci{0.31}{0.32}} \\

\midrule

\multirow{4}{*}{\texttt{Drawer}} & \multirow{2}{*}{X-VLA} & PR\higherBetter & 0.84 & 0.88 & \cellcolor{highlight_0} 0.90{\ci{0.89}{0.91}} \\
& & PDT\lowerBetter & 0.21 &0.21 & \cellcolor{highlight_0} 0.19{\ci{0.19}{0.19}} \\
& \multirow{2}{*}{$\pi_{0.5}$} & PR\higherBetter & \cellcolor{highlight_0} 0.78 & 0.77 & 0.69{\ci{0.66}{0.72}} \\
& & PDT\lowerBetter & 0.41 & \cellcolor{highlight_0} 0.40 & 0.45{\ci{0.44}{0.46}} \\

\midrule

\multirow{4}{*}{\texttt{Kitchen}} & \multirow{2}{*}{X-VLA} & PR\higherBetter & \cellcolor{highlight_0} 1.00 & 0.96 & \cellcolor{highlight_0} 1.00{\ci{1.00}{1.00}} \\
& & PDT\lowerBetter & \cellcolor{highlight_0} 0.20 & 0.24 & \cellcolor{highlight_0} 0.20{\ci{0.20}{0.20}} \\
& \multirow{2}{*}{$\pi_{0.5}$} & PR\higherBetter & 0.65{\ci{0.41}{0.88}} & \cellcolor{highlight_0} 0.74{\ci{0.32}{1.00}} & 00.62{\ci{0.41}{0.82}} \\
& & PDT\lowerBetter & 0.32{\ci{0.18}{0.46}} & \cellcolor{highlight_0} 0.25{\ci{0.05}{0.45}} & 0.33{\ci{0.20}{0.46}} \\

\midrule

\multirow{4}{*}{\texttt{Stack T}} & \multirow{2}{*}{X-VLA} & PR\higherBetter & \cellcolor{highlight_0} 0.97 & 0.78 & \cellcolor{highlight_0} 0.97{\ci{0.97}{0.97}} \\
& & PDT\lowerBetter & 0.09 & 0.20 & \cellcolor{highlight_0} 0.08{\ci{0.08}{0.08}} \\
& \multirow{2}{*}{$\pi_{0.5}$} & PR\higherBetter & \cellcolor{highlight_0} 0.97 & \cellcolor{highlight_0} 0.97 & 0.95{\ci{0.95}{0.96}} \\
& & PDT\lowerBetter & \cellcolor{highlight_0} 0.09 & 0.11 & 0.10{\ci{0.10}{0.10}} \\

\midrule

\multirow{4}{*}{\texttt{Mixer}} & \multirow{2}{*}{X-VLA} & PR\higherBetter & 0.97 &  0.97 & \cellcolor{highlight_0} 0.98{\ci{0.98}{0.98}} \\
& & PDT\lowerBetter & \cellcolor{highlight_0} 0.10 & 0.13 & 0.11{\ci{0.11}{0.11}} \\
& \multirow{2}{*}{$\pi_{0.5}$} & PR\higherBetter & \cellcolor{highlight_0} 0.77 & 0.61 & \cellcolor{highlight_0} 0.77{\ci{0.77}{0.77}} \\
& & PDT\lowerBetter & 0.30 & 0.45 & \cellcolor{highlight_0} 0.28{\ci{0.27}{0.28}} \\

\midrule

\multirow{4}{*}{\texttt{Cups}} & \multirow{2}{*}{X-VLA} & PR\higherBetter & 0.86 & \cellcolor{highlight_0} 0.95 & 0.78{\ci{0.78}{0.79}} \\
& & PDT\lowerBetter & 0.32 & \cellcolor{highlight_0} 0.24 & 0.33{\ci{0.33}{0.34}} \\
& \multirow{2}{*}{$\pi_{0.5}$} & PR\higherBetter & 0.48 & \cellcolor{highlight_0} 0.61 & 0.58{\ci{0.57}{0.59}} \\
& & PDT\lowerBetter & 0.61 & \cellcolor{highlight_0} 0.51 & 0.52{\ci{0.52}{0.52}} \\

\midrule
\midrule

\multirow{4}{*}{\texttt{Average}} & \multirow{2}{*}{X-VLA} & PR\higherBetter & \cellcolor{highlight_0} 0.92 & 0.90 & 0.91 \\
& & PDT\lowerBetter & \cellcolor{highlight_0} 0.20 & 0.21 & \cellcolor{highlight_0} 0.20 \\
& \multirow{2}{*}{$\pi_{0.5}$} & PR\higherBetter & 0.73 & \cellcolor{highlight_0} 0.76 & 0.72 \\
& & PDT\lowerBetter & 0.35 & \cellcolor{highlight_0} 0.33 & \cellcolor{highlight_0} 0.33 \\

        \bottomrule
    \end{tabular}
\end{table}

\subsection{FAIL vs RAPT}
\label{app:baselines:rapt}
The Recurrent Anomaly Probabilistic Trajectory Model (RAPT) \citep{munn2026rapt} detects task failure by predicting future proprioceptive observations from a history of proprioceptive states and actions using a recurrent model.
While originally designed for humanoid locomotion tasks, in which policies only observe proprioception, one can also apply RAPT to robot proprioception in manipulation tasks.
We train RAPT on current proprioception (EEF pose, gripper state) and previous absolute EEF actions as inputs, and evaluate RAPT once per VLA inference.
As \Cref{tab:rapt} demonstrates, FAIL outperforms RAPT on nearly all tasks and metrics, without requiring an additional training step.

\begin{table}[ht]    
    \caption{Threshold-independent detector performance of RAPT on the \textbf{real-world tasks}. PR and PDT (see~\Cref{sec:eval:metrics}) refer to the areas under the respective curves, and the small numbers indicate $95\%$ confidence intervals. The \colorbox{highlight_0}{best} method is highlighted. FAIL performs on par with or outperforms RAPT on nearly all tasks and metrics.}
    \label{tab:rapt}
    \centering

    \begin{tabular}{l l @{\hspace{4pt}} l S[table-format=1.2] | *{1}{S[table-format=1.3, table-space-text-post={\ci{0.00}{0.00}}]}}
         & & & {RAPT} & {FAIL} \\
        \midrule

\multirow{4}{*}{\texttt{Drawer}} & \multirow{2}{*}{X-VLA} & PR\higherBetter & \cellcolor{highlight_0} 0.93 & 0.91{\ci{0.90}{0.92}} \\
& & PDT\lowerBetter & 0.28 & \cellcolor{highlight_0} 0.22{\ci{0.22}{0.22}} \\
& \multirow{2}{*}{$\pi_{0.5}$} & PR\higherBetter & \cellcolor{highlight_0} 0.72 & 0.68{\ci{0.65}{0.71}} \\
& & PDT\lowerBetter & \cellcolor{highlight_0} 0.39 & 0.47{\ci{0.45}{0.48}} \\

\midrule

\multirow{4}{*}{\texttt{Kitchen}} & \multirow{2}{*}{X-VLA} & PR\higherBetter & 0.52 & \cellcolor{highlight_0} 1.00{\ci{1.00}{1.00}} \\
& & PDT\lowerBetter & 0.53 & \cellcolor{highlight_0} 0.21{\ci{0.21}{0.21}} \\
& \multirow{2}{*}{$\pi_{0.5}$} & PR\higherBetter & \cellcolor{highlight_0} 0.78 & 0.74{\ci{0.64}{0.84}} \\
& & PDT\lowerBetter & 0.30 & \cellcolor{highlight_0} 0.28{\ci{0.16}{0.39}} \\

\midrule

\multirow{4}{*}{\texttt{Stack T}} & \multirow{2}{*}{X-VLA} & PR\higherBetter & \cellcolor{highlight_0} 0.98 & \cellcolor{highlight_0} 0.98{\ci{0.98}{0.98}} \\
& & PDT\lowerBetter & \cellcolor{highlight_0} 0.08 & \cellcolor{highlight_0} 0.08{\ci{0.08}{0.08}} \\
& \multirow{2}{*}{$\pi_{0.5}$} & PR\higherBetter & 0.92 & \cellcolor{highlight_0} 0.96{\ci{0.96}{0.97}} \\
& & PDT\lowerBetter & 0.11 & \cellcolor{highlight_0} 0.10{\ci{0.09}{0.10}} \\

\midrule

\multirow{4}{*}{\texttt{Mixer}} & \multirow{2}{*}{X-VLA} & PR\higherBetter & \cellcolor{highlight_0} 0.97 & \cellcolor{highlight_0} 0.97{\ci{0.97}{0.97}} \\
& & PDT\lowerBetter & 0.16 & \cellcolor{highlight_0} 0.11{\ci{0.11}{0.11}} \\
& \multirow{2}{*}{$\pi_{0.5}$} & PR\higherBetter & 0.63 & \cellcolor{highlight_0} 0.79{\ci{0.79}{0.79}} \\
& & PDT\lowerBetter & 0.46 & \cellcolor{highlight_0} 0.27{\ci{0.27}{0.27}} \\

\midrule
\midrule

\multirow{4}{*}{\texttt{Average}} & \multirow{2}{*}{X-VLA} & PR\higherBetter & 0.85 & \cellcolor{highlight_0} 0.94 \\
& & PDT\lowerBetter & 0.28 & \cellcolor{highlight_0} 0.19 \\
& \multirow{2}{*}{$\pi_{0.5}$} & PR\higherBetter & 0.71 & \cellcolor{highlight_0} 0.77 \\
& & PDT\lowerBetter & 0.38 & \cellcolor{highlight_0} 0.31 \\

    \end{tabular}
\end{table}

\FloatBarrier
\section{Threshold-Dependent Results}
\label{sec:thresh-dependent}
This section contains threshold-dependent results for the real-world tasks and Libero-Plus.
These results are less reliable for comparing methods than the threshold-independent results in the main paper, as threshold selection is a major confounding factor and there is no best threshold~\citep{xu2025faildetect}.
We select all thresholds using a constant conformal prediction (CP) band at significance level $0.05$, following \citet{romer2026fiper}.
We do not use time-dependent conformal prediction bands, as they are not applicable to episodes of varying length, such as in the \texttt{Drawer} task.

\FloatBarrier
\subsection{Real-World Tasks}
\label{sec:thresh-real}
We summarize threshold-dependent results for the real-world tasks in \Cref{tab:eval-real-thresh0.95-pt1} and \Cref{tab:eval-real-thresh0.95-pt2}.
Note that we only use $20$ rollouts for threshold calibration, which were randomly selected from the successful evaluation rollouts and excluded from subsequent threshold-dependent evaluation.
Due to the small number of rollouts and the low significance level, a single outlier in the calibration set can significantly raise the score threshold, leading to low recall for some methods on some tasks.

\begin{table}[ht]
    \caption{Threshold-dependent detector performance on \textbf{real-world} tasks, using a constant CP band at significance level $0.05$. The top three performing methods are highlighted. Note that we are using only 20 calibration rollouts, making some methods underperform in terms of recall due to outliers in the calibration set.}
    \label{tab:eval-real-thresh0.95-pt1}

    \centering
    \begin{tabular}{l l l *{3}{S[table-format=1.2, table-space-text-post={\ci{0.00}{0.00}}]} | S[table-format=1.2] *{2}{S[table-format=1.2, table-space-text-post={\ci{0.00}{0.00}}]}}
        \toprule
         & & & {ACE} & {Diff} & {STAC} & {ACC} & {LLMD} & {FAIL} \\
        \midrule

\multirow{12}{*}{\texttt{Blocks}} & \multirow{6}{*}{X-VLA} & Rec & \cellcolor{highlight_2} 0.39{\ci{0.35}{0.43}} & 0.22{\ci{0.16}{0.29}} & 0.00{\ci{0.00}{0.00}} & \cellcolor{highlight_1} 0.47 & 0.07{\ci{0.02}{0.13}} & \cellcolor{highlight_0} 0.52{\ci{0.50}{0.53}} \\
& & Prec & \cellcolor{highlight_0} 1.00{\ci{1.00}{1.00}} & \cellcolor{highlight_0} 1.00{\ci{1.00}{1.00}} & \cellcolor{highlight_2} 0.00{\ci{0.00}{0.00}} & \cellcolor{highlight_0} 1.00 & \cellcolor{highlight_0} 1.00{\ci{1.00}{1.00}} & \cellcolor{highlight_0} 1.00{\ci{1.00}{1.00}} \\
& & Acc & \cellcolor{highlight_2} 0.57{\ci{0.55}{0.60}} & 0.45{\ci{0.41}{0.50}} & 0.30{\ci{0.30}{0.30}} & \cellcolor{highlight_1} 0.62 & 0.35{\ci{0.31}{0.39}} & \cellcolor{highlight_0} 0.66{\ci{0.65}{0.67}} \\
& & WACC & \cellcolor{highlight_2} 0.70{\ci{0.68}{0.72}} & 0.61{\ci{0.58}{0.64}} & 0.50{\ci{0.50}{0.50}} & \cellcolor{highlight_1} 0.73 & 0.54{\ci{0.51}{0.57}} & \cellcolor{highlight_0} 0.76{\ci{0.75}{0.77}} \\
& & F1 & \cellcolor{highlight_2} 0.56{\ci{0.52}{0.60}} & 0.36{\ci{0.27}{0.45}} & 0.00{\ci{0.00}{0.00}} & \cellcolor{highlight_1} 0.64 & 0.14{\ci{0.04}{0.23}} & \cellcolor{highlight_0} 0.68{\ci{0.67}{0.70}} \\
& & TNR & \cellcolor{highlight_0} 1.00{\ci{1.00}{1.00}} & \cellcolor{highlight_0} 1.00{\ci{1.00}{1.00}} & \cellcolor{highlight_0} 1.00{\ci{1.00}{1.00}} & \cellcolor{highlight_0} 1.00 & \cellcolor{highlight_0} 1.00{\ci{1.00}{1.00}} & \cellcolor{highlight_0} 1.00{\ci{1.00}{1.00}} \\
 & \multirow{4}{*}{$\pi_{0.5}$} & Rec & 0.29{\ci{0.25}{0.33}} & \cellcolor{highlight_2} 0.51{\ci{0.21}{0.80}} & \cellcolor{highlight_1} 0.62{\ci{0.50}{0.74}} & \cellcolor{highlight_0} 0.82 & 0.03{\ci{0.03}{0.03}} & \cellcolor{highlight_0} 0.82{\ci{0.82}{0.82}} \\
& & Prec & 0.78{\ci{0.75}{0.82}} & \cellcolor{highlight_2} 0.93{\ci{0.86}{1.00}} & \cellcolor{highlight_1} 0.94{\ci{0.92}{0.96}} & 0.90 & \cellcolor{highlight_0} 1.00{\ci{1.00}{1.00}} & 0.90{\ci{0.90}{0.90}} \\
& & Acc & 0.53{\ci{0.51}{0.55}} & \cellcolor{highlight_2} 0.67{\ci{0.53}{0.82}} & \cellcolor{highlight_1} 0.75{\ci{0.68}{0.81}} & \cellcolor{highlight_0} 0.84 & 0.42{\ci{0.42}{0.42}} & \cellcolor{highlight_0} 0.84{\ci{0.84}{0.84}} \\
& & WACC & 0.59{\ci{0.57}{0.60}} & \cellcolor{highlight_2} 0.71{\ci{0.61}{0.82}} & \cellcolor{highlight_1} 0.78{\ci{0.73}{0.82}} & \cellcolor{highlight_0} 0.84 & 0.52{\ci{0.52}{0.52}} & \cellcolor{highlight_0} 0.84{\ci{0.84}{0.84}} \\
& & F1 & 0.43{\ci{0.38}{0.47}} & \cellcolor{highlight_2} 0.62{\ci{0.38}{0.87}} & \cellcolor{highlight_1} 0.74{\ci{0.66}{0.82}} & \cellcolor{highlight_0} 0.86 & 0.06{\ci{0.06}{0.06}} & \cellcolor{highlight_0} 0.86{\ci{0.86}{0.86}} \\
& & TNR & 0.88{\ci{0.85}{0.91}} & \cellcolor{highlight_2} 0.92{\ci{0.85}{1.00}} & \cellcolor{highlight_1} 0.94{\ci{0.91}{0.97}} & 0.86 & \cellcolor{highlight_0} 1.00{\ci{1.00}{1.00}} & 0.86{\ci{0.86}{0.86}} \\

        \midrule

\multirow{12}{*}{\texttt{Drawer}} & \multirow{6}{*}{X-VLA} & Rec & \cellcolor{highlight_1} 0.93{\ci{0.91}{0.96}} & \cellcolor{highlight_2} 0.77{\ci{0.70}{0.84}} & 0.00{\ci{0.00}{0.00}} & 0.40 & \cellcolor{highlight_0} 0.96{\ci{0.96}{0.96}} & \cellcolor{highlight_0} 0.96{\ci{0.96}{0.96}} \\
& & Prec & \cellcolor{highlight_1} 0.95{\ci{0.89}{1.01}} & \cellcolor{highlight_0} 1.00{\ci{1.00}{1.00}} & 0.00{\ci{0.00}{0.00}} & \cellcolor{highlight_0} 1.00 & \cellcolor{highlight_2} 0.89{\ci{0.89}{0.89}} & \cellcolor{highlight_2} 0.89{\ci{0.89}{0.89}} \\
& & Acc & \cellcolor{highlight_0} 0.96{\ci{0.94}{0.97}} & \cellcolor{highlight_2} 0.92{\ci{0.89}{0.94}} & 0.53{\ci{0.52}{0.54}} & 0.78 & \cellcolor{highlight_1} 0.94{\ci{0.94}{0.94}} & \cellcolor{highlight_1} 0.94{\ci{0.94}{0.94}} \\
& & WACC & \cellcolor{highlight_0} 0.95{\ci{0.94}{0.96}} & \cellcolor{highlight_2} 0.89{\ci{0.85}{0.92}} & 0.42{\ci{0.41}{0.43}} & 0.70 & \cellcolor{highlight_1} 0.94{\ci{0.94}{0.94}} & \cellcolor{highlight_1} 0.94{\ci{0.94}{0.94}} \\
& & F1 & \cellcolor{highlight_0} 0.94{\ci{0.92}{0.96}} & \cellcolor{highlight_2} 0.87{\ci{0.83}{0.91}} & 0.00{\ci{0.00}{0.00}} & 0.57 & \cellcolor{highlight_1} 0.92{\ci{0.92}{0.92}} & \cellcolor{highlight_1} 0.92{\ci{0.92}{0.92}} \\
& & TNR & \cellcolor{highlight_1} 0.97{\ci{0.93}{1.01}} & \cellcolor{highlight_0} 1.00{\ci{1.00}{1.00}} & 0.84{\ci{0.83}{0.86}} & \cellcolor{highlight_0} 1.00 & \cellcolor{highlight_2} 0.93{\ci{0.93}{0.93}} & \cellcolor{highlight_2} 0.93{\ci{0.93}{0.93}} \\
 & \multirow{4}{*}{$\pi_{0.5}$} & Rec & 0.13{\ci{0.07}{0.19}} & \cellcolor{highlight_2} 0.57{\ci{0.35}{0.80}} & \cellcolor{highlight_0} 0.69{\ci{0.63}{0.75}} & 0.50 & \cellcolor{highlight_1} 0.61{\ci{0.50}{0.71}} & \cellcolor{highlight_0} 0.69{\ci{0.60}{0.78}} \\
& & Prec & 0.54{\ci{0.39}{0.69}} & \cellcolor{highlight_2} 0.64{\ci{0.56}{0.72}} & \cellcolor{highlight_1} 0.69{\ci{0.63}{0.74}} & 0.52 & \cellcolor{highlight_0} 0.73{\ci{0.68}{0.77}} & 0.52{\ci{0.49}{0.55}} \\
& & Acc & 0.65{\ci{0.63}{0.68}} & \cellcolor{highlight_2} 0.74{\ci{0.69}{0.79}} & \cellcolor{highlight_1} 0.78{\ci{0.76}{0.80}} & 0.67 & \cellcolor{highlight_0} 0.79{\ci{0.75}{0.83}} & 0.67{\ci{0.65}{0.70}} \\
& & WACC & 0.53{\ci{0.51}{0.55}} & \cellcolor{highlight_2} 0.70{\ci{0.62}{0.79}} & \cellcolor{highlight_0} 0.76{\ci{0.74}{0.78}} & 0.63 & \cellcolor{highlight_1} 0.74{\ci{0.69}{0.80}} & 0.68{\ci{0.63}{0.72}} \\
& & F1 & 0.20{\ci{0.12}{0.28}} & \cellcolor{highlight_2} 0.59{\ci{0.46}{0.72}} & \cellcolor{highlight_0} 0.69{\ci{0.67}{0.71}} & 0.51 & \cellcolor{highlight_1} 0.66{\ci{0.58}{0.74}} & \cellcolor{highlight_2} 0.59{\ci{0.54}{0.65}} \\
& & TNR & \cellcolor{highlight_0} 0.93{\ci{0.88}{0.98}} & \cellcolor{highlight_2} 0.83{\ci{0.75}{0.91}} & \cellcolor{highlight_2} 0.83{\ci{0.77}{0.89}} & 0.75 & \cellcolor{highlight_1} 0.88{\ci{0.87}{0.89}} & 0.67{\ci{0.65}{0.68}} \\

\midrule

\multirow{12}{*}{\texttt{Cups}} & \multirow{6}{*}{X-VLA} & Rec & 0.68{\ci{0.63}{0.73}} & 0.69{\ci{0.66}{0.71}} & \cellcolor{highlight_0} 1.00{\ci{1.00}{1.00}} & \cellcolor{highlight_2} 0.94 & 0.44{\ci{0.42}{0.45}} & \cellcolor{highlight_1} 0.96{\ci{0.96}{0.96}} \\
& & Prec & 0.96{\ci{0.95}{0.98}} & \cellcolor{highlight_2} 0.97{\ci{0.94}{1.00}} & \cellcolor{highlight_1} 0.98{\ci{0.98}{0.98}} & \cellcolor{highlight_1} 0.98 & 0.88{\ci{0.87}{0.88}} & 0.94{\ci{0.94}{0.94}} \\
& & Acc & 0.80{\ci{0.78}{0.82}} & 0.81{\ci{0.80}{0.82}} & \cellcolor{highlight_0} 0.99{\ci{0.99}{0.99}} & \cellcolor{highlight_1} 0.95 & 0.64{\ci{0.63}{0.65}} & \cellcolor{highlight_2} 0.94{\ci{0.94}{0.94}} \\
& & WACC & 0.82{\ci{0.80}{0.84}} & 0.83{\ci{0.82}{0.84}} & \cellcolor{highlight_0} 0.99{\ci{0.99}{0.99}} & \cellcolor{highlight_1} 0.96 & 0.68{\ci{0.67}{0.68}} & \cellcolor{highlight_2} 0.94{\ci{0.94}{0.94}} \\
& & F1 & 0.80{\ci{0.77}{0.83}} & 0.80{\ci{0.79}{0.82}} & \cellcolor{highlight_0} 0.99{\ci{0.99}{0.99}} & \cellcolor{highlight_1} 0.96 & 0.58{\ci{0.57}{0.59}} & \cellcolor{highlight_2} 0.95{\ci{0.95}{0.95}} \\
& & TNR & \cellcolor{highlight_2} 0.96{\ci{0.94}{0.98}} & \cellcolor{highlight_1} 0.97{\ci{0.94}{1.00}} & \cellcolor{highlight_1} 0.97{\ci{0.97}{0.97}} & \cellcolor{highlight_1} 0.97 & 0.92{\ci{0.92}{0.92}} & 0.92{\ci{0.92}{0.92}} \\
 & \multirow{4}{*}{$\pi_{0.5}$} & Rec & 0.68{\ci{0.63}{0.73}} & 0.69{\ci{0.66}{0.71}} & \cellcolor{highlight_0} 1.00{\ci{1.00}{1.00}} & \cellcolor{highlight_2} 0.94 & 0.44{\ci{0.42}{0.45}} & \cellcolor{highlight_1} 0.96{\ci{0.96}{0.96}} \\
& & Prec & 0.96{\ci{0.95}{0.98}} & \cellcolor{highlight_2} 0.97{\ci{0.94}{1.00}} & \cellcolor{highlight_1} 0.98{\ci{0.98}{0.98}} & \cellcolor{highlight_1} 0.98 & 0.88{\ci{0.87}{0.88}} & 0.94{\ci{0.94}{0.94}} \\
& & Acc & 0.80{\ci{0.78}{0.82}} & 0.81{\ci{0.80}{0.82}} & \cellcolor{highlight_0} 0.99{\ci{0.99}{0.99}} & \cellcolor{highlight_1} 0.95 & 0.64{\ci{0.63}{0.65}} & \cellcolor{highlight_2} 0.94{\ci{0.94}{0.94}} \\
& & WACC & 0.82{\ci{0.80}{0.84}} & 0.83{\ci{0.82}{0.84}} & \cellcolor{highlight_0} 0.99{\ci{0.99}{0.99}} & \cellcolor{highlight_1} 0.96 & 0.68{\ci{0.67}{0.68}} & \cellcolor{highlight_2} 0.94{\ci{0.94}{0.94}} \\
& & F1 & 0.80{\ci{0.77}{0.83}} & 0.80{\ci{0.79}{0.82}} & \cellcolor{highlight_0} 0.99{\ci{0.99}{0.99}} & \cellcolor{highlight_1} 0.96 & 0.58{\ci{0.57}{0.59}} & \cellcolor{highlight_2} 0.95{\ci{0.95}{0.95}} \\
& & TNR & \cellcolor{highlight_2} 0.96{\ci{0.94}{0.98}} & \cellcolor{highlight_1} 0.97{\ci{0.94}{1.00}} & \cellcolor{highlight_1} 0.97{\ci{0.97}{0.97}} & \cellcolor{highlight_1} 0.97 & 0.92{\ci{0.92}{0.92}} & 0.92{\ci{0.92}{0.92}} \\

\bottomrule
    \end{tabular}
\end{table}

\begin{table}[ht]
    \caption{Threshold-dependent detector performance on \textbf{real-world} tasks, using a constant CP band at significance level $0.05$. The top three performing methods are highlighted. Note that we are using only 20 calibration rollouts, making some methods underperform in terms of recall due to outliers in the calibration set.}
    \label{tab:eval-real-thresh0.95-pt2}

    \centering
    \begin{tabular}{l l l *{3}{S[table-format=1.2, table-space-text-post={\ci{0.00}{0.00}}]} | S[table-format=1.2] *{2}{S[table-format=1.2, table-space-text-post={\ci{0.00}{0.00}}]}}
        \toprule
         & & & {ACE} & {Diff} & {STAC} & {ACC} & {LLMD} & {FAIL} \\
        \midrule

\multirow{12}{*}{\texttt{Kitchen}} & \multirow{6}{*}{X-VLA} & Rec & \cellcolor{highlight_1} 0.96{\ci{0.96}{0.96}} & \cellcolor{highlight_2} 0.95{\ci{0.93}{0.98}} & 0.00{\ci{0.00}{0.00}} & \cellcolor{highlight_0} 1.00 & \cellcolor{highlight_0} 1.00{\ci{1.00}{1.00}} & \cellcolor{highlight_0} 1.00{\ci{1.00}{1.00}} \\
& & Prec & 0.94{\ci{0.92}{0.96}} & \cellcolor{highlight_2} 0.95{\ci{0.93}{0.97}} & 0.00{\ci{0.00}{0.00}} & 0.93 & \cellcolor{highlight_1} 0.97{\ci{0.97}{0.97}} & 0.93{\ci{0.93}{0.93}} \\
& & Acc & \cellcolor{highlight_2} 0.94{\ci{0.92}{0.95}} & \cellcolor{highlight_2} 0.94{\ci{0.92}{0.95}} & 0.31{\ci{0.31}{0.31}} & \cellcolor{highlight_1} 0.95 & \cellcolor{highlight_0} 0.98{\ci{0.98}{0.98}} & \cellcolor{highlight_1} 0.95{\ci{0.95}{0.95}} \\
& & WACC & \cellcolor{highlight_2} 0.92{\ci{0.90}{0.95}} & \cellcolor{highlight_1} 0.93{\ci{0.91}{0.95}} & 0.46{\ci{0.46}{0.46}} & \cellcolor{highlight_1} 0.93 & \cellcolor{highlight_0} 0.96{\ci{0.96}{0.96}} & \cellcolor{highlight_1} 0.93{\ci{0.93}{0.93}} \\
& & F1 & \cellcolor{highlight_2} 0.95{\ci{0.94}{0.96}} & \cellcolor{highlight_2} 0.95{\ci{0.94}{0.96}} & 0.00{\ci{0.00}{0.00}} & \cellcolor{highlight_1} 0.97 & \cellcolor{highlight_0} 0.98{\ci{0.98}{0.98}} & \cellcolor{highlight_1} 0.97{\ci{0.97}{0.97}} \\
& & TNR & 0.88{\ci{0.83}{0.93}} & \cellcolor{highlight_2} 0.90{\ci{0.86}{0.95}} & \cellcolor{highlight_1} 0.93{\ci{0.93}{0.93}} & 0.86 & \cellcolor{highlight_1} 0.93{\ci{0.93}{0.93}} & 0.86{\ci{0.86}{0.86}} \\
 & \multirow{4}{*}{$\pi_{0.5}$} & Rec & \cellcolor{highlight_1} 0.82{\ci{0.81}{0.84}} & 0.14{\ci{0.00}{0.39}} & \cellcolor{highlight_2} 0.74{\ci{0.69}{0.79}} & \cellcolor{highlight_0} 0.89 & 0.01{\ci{0.00}{0.03}} & \cellcolor{highlight_0} 0.89{\ci{0.89}{0.89}} \\
& & Prec & \cellcolor{highlight_1} 0.95{\ci{0.93}{0.97}} & 0.67{\ci{0.01}{1.32}} & \cellcolor{highlight_0} 0.97{\ci{0.96}{0.97}} & \cellcolor{highlight_0} 0.97 & 0.17{\ci{0.00}{0.49}} & \cellcolor{highlight_1} 0.95{\ci{0.94}{0.97}} \\
& & Acc & \cellcolor{highlight_2} 0.84{\ci{0.83}{0.85}} & 0.37{\ci{0.19}{0.55}} & 0.79{\ci{0.75}{0.83}} & \cellcolor{highlight_0} 0.90 & 0.26{\ci{0.25}{0.28}} & \cellcolor{highlight_1} 0.89{\ci{0.88}{0.90}} \\
& & WACC & \cellcolor{highlight_2} 0.85{\ci{0.83}{0.87}} & 0.57{\ci{0.45}{0.70}} & 0.83{\ci{0.81}{0.86}} & \cellcolor{highlight_0} 0.91 & 0.48{\ci{0.46}{0.50}} & \cellcolor{highlight_1} 0.89{\ci{0.86}{0.91}} \\
& & F1 & \cellcolor{highlight_2} 0.88{\ci{0.87}{0.89}} & 0.21{\ci{0.00}{0.56}} & 0.84{\ci{0.80}{0.87}} & \cellcolor{highlight_0} 0.93 & 0.02{\ci{0.00}{0.05}} & \cellcolor{highlight_1} 0.92{\ci{0.91}{0.93}} \\
& & TNR & 0.88{\ci{0.83}{0.93}} & \cellcolor{highlight_0} 1.00{\ci{1.00}{1.00}} & \cellcolor{highlight_2} 0.93{\ci{0.93}{0.93}} & \cellcolor{highlight_2} 0.93 & \cellcolor{highlight_1} 0.95{\ci{0.91}{1.00}} & 0.88{\ci{0.83}{0.93}} \\

\midrule

\multirow{12}{*}{\texttt{Stack T}} & \multirow{6}{*}{X-VLA} & Rec & \cellcolor{highlight_2} 0.85{\ci{0.85}{0.85}} & \cellcolor{highlight_1} 0.87{\ci{0.87}{0.87}} & 0.00{\ci{0.00}{0.00}} & \cellcolor{highlight_0} 0.89 & \cellcolor{highlight_1} 0.87{\ci{0.87}{0.87}} & \cellcolor{highlight_0} 0.89{\ci{0.89}{0.89}} \\
& & Prec & \cellcolor{highlight_0} 1.00{\ci{1.00}{1.00}} & \cellcolor{highlight_0} 1.00{\ci{1.00}{1.00}} & \cellcolor{highlight_2} 0.00{\ci{0.00}{0.00}} & \cellcolor{highlight_0} 1.00 & \cellcolor{highlight_0} 1.00{\ci{1.00}{1.00}} & \cellcolor{highlight_0} 1.00{\ci{1.00}{1.00}} \\
& & Acc & \cellcolor{highlight_2} 0.86{\ci{0.86}{0.86}} & \cellcolor{highlight_1} 0.87{\ci{0.87}{0.87}} & 0.02{\ci{0.02}{0.02}} & \cellcolor{highlight_0} 0.89 & \cellcolor{highlight_1} 0.87{\ci{0.87}{0.87}} & \cellcolor{highlight_0} 0.89{\ci{0.89}{0.89}} \\
& & WACC & \cellcolor{highlight_1} 0.93{\ci{0.93}{0.93}} & \cellcolor{highlight_1} 0.93{\ci{0.93}{0.93}} & \cellcolor{highlight_2} 0.50{\ci{0.50}{0.50}} & \cellcolor{highlight_0} 0.95 & \cellcolor{highlight_1} 0.93{\ci{0.93}{0.93}} & \cellcolor{highlight_0} 0.95{\ci{0.95}{0.95}} \\
& & F1 & \cellcolor{highlight_2} 0.92{\ci{0.92}{0.92}} & \cellcolor{highlight_1} 0.93{\ci{0.93}{0.93}} & 0.00{\ci{0.00}{0.00}} & \cellcolor{highlight_0} 0.94 & \cellcolor{highlight_1} 0.93{\ci{0.93}{0.93}} & \cellcolor{highlight_0} 0.94{\ci{0.94}{0.94}} \\
& & TNR & \cellcolor{highlight_0} 1.00{\ci{1.00}{1.00}} & \cellcolor{highlight_0} 1.00{\ci{1.00}{1.00}} & \cellcolor{highlight_0} 1.00{\ci{1.00}{1.00}} & \cellcolor{highlight_0} 1.00 & \cellcolor{highlight_0} 1.00{\ci{1.00}{1.00}} & \cellcolor{highlight_0} 1.00{\ci{1.00}{1.00}} \\
 & \multirow{4}{*}{$\pi_{0.5}$} & Rec & 0.30{\ci{0.24}{0.37}} & 0.35{\ci{0.06}{0.64}} & 0.14{\ci{0.11}{0.16}} & 0.35 & \cellcolor{highlight_2} 0.53{\ci{0.51}{0.56}} & \cellcolor{highlight_0} 0.66{\ci{0.64}{0.68}} \\
& & Prec & 0.96{\ci{0.92}{1.00}} & \cellcolor{highlight_1} 0.99{\ci{0.97}{1.01}} & \cellcolor{highlight_2} 0.97{\ci{0.92}{1.03}} & \cellcolor{highlight_0} 1.00 & \cellcolor{highlight_0} 1.00{\ci{1.00}{1.00}} & \cellcolor{highlight_0} 1.00{\ci{1.00}{1.00}} \\
& & Acc & 0.40{\ci{0.35}{0.44}} & 0.44{\ci{0.20}{0.68}} & 0.26{\ci{0.25}{0.27}} & 0.44 & \cellcolor{highlight_2} 0.60{\ci{0.58}{0.62}} & \cellcolor{highlight_0} 0.71{\ci{0.70}{0.73}} \\
& & WACC & 0.61{\ci{0.60}{0.62}} & 0.66{\ci{0.52}{0.80}} & 0.55{\ci{0.54}{0.57}} & 0.67 & \cellcolor{highlight_1} 0.77{\ci{0.75}{0.78}} & \cellcolor{highlight_0} 0.83{\ci{0.82}{0.84}} \\
& & F1 & 0.46{\ci{0.39}{0.53}} & 0.47{\ci{0.11}{0.83}} & 0.24{\ci{0.20}{0.27}} & 0.52 & \cellcolor{highlight_2} 0.69{\ci{0.67}{0.72}} & \cellcolor{highlight_0} 0.80{\ci{0.78}{0.81}} \\
& & TNR & \cellcolor{highlight_2} 0.92{\ci{0.82}{1.01}} & \cellcolor{highlight_1} 0.97{\ci{0.92}{1.03}} & \cellcolor{highlight_1} 0.97{\ci{0.92}{1.03}} & \cellcolor{highlight_0} 1.00 & \cellcolor{highlight_0} 1.00{\ci{1.00}{1.00}} & \cellcolor{highlight_0} 1.00{\ci{1.00}{1.00}} \\

\midrule

\multirow{12}{*}{\texttt{Mixer}} & \multirow{6}{*}{X-VLA} & Rec & 0.83{\ci{0.81}{0.84}} & 0.85{\ci{0.84}{0.87}} & 0.00{\ci{0.00}{0.00}} & \cellcolor{highlight_1} 0.90 & \cellcolor{highlight_2} 0.89{\ci{0.88}{0.91}} & \cellcolor{highlight_0} 0.92{\ci{0.92}{0.92}} \\
& & Prec & \cellcolor{highlight_1} 0.97{\ci{0.95}{0.98}} & \cellcolor{highlight_0} 0.98{\ci{0.98}{0.98}} & 0.00{\ci{0.00}{0.00}} & \cellcolor{highlight_0} 0.98 & \cellcolor{highlight_0} 0.98{\ci{0.98}{0.98}} & \cellcolor{highlight_0} 0.98{\ci{0.98}{0.98}} \\
& & Acc & 0.82{\ci{0.80}{0.84}} & \cellcolor{highlight_2} 0.85{\ci{0.84}{0.86}} & 0.11{\ci{0.11}{0.11}} & \cellcolor{highlight_1} 0.89 & \cellcolor{highlight_1} 0.89{\ci{0.88}{0.90}} & \cellcolor{highlight_0} 0.91{\ci{0.91}{0.91}} \\
& & WACC & 0.80{\ci{0.74}{0.86}} & 0.84{\ci{0.84}{0.85}} & 0.50{\ci{0.50}{0.50}} & \cellcolor{highlight_1} 0.87 & \cellcolor{highlight_2} 0.86{\ci{0.86}{0.87}} & \cellcolor{highlight_0} 0.88{\ci{0.88}{0.88}} \\
& & F1 & 0.89{\ci{0.88}{0.90}} & 0.91{\ci{0.90}{0.92}} & 0.00{\ci{0.00}{0.00}} & \cellcolor{highlight_1} 0.94 & \cellcolor{highlight_2} 0.93{\ci{0.93}{0.94}} & \cellcolor{highlight_0} 0.95{\ci{0.95}{0.95}} \\
& & TNR & \cellcolor{highlight_2} 0.78{\ci{0.67}{0.89}} & \cellcolor{highlight_1} 0.83{\ci{0.83}{0.83}} & \cellcolor{highlight_0} 1.00{\ci{1.00}{1.00}} & \cellcolor{highlight_1} 0.83 & \cellcolor{highlight_1} 0.83{\ci{0.83}{0.83}} & \cellcolor{highlight_1} 0.83{\ci{0.83}{0.83}} \\
 & \multirow{4}{*}{$\pi_{0.5}$} & Rec & \cellcolor{highlight_0} 0.77{\ci{0.75}{0.79}} & \cellcolor{highlight_2} 0.30{\ci{0.12}{0.47}} & 0.07{\ci{0.07}{0.07}} & 0.17 & 0.24{\ci{0.12}{0.36}} & \cellcolor{highlight_1} 0.33{\ci{0.25}{0.41}} \\
& & Prec & \cellcolor{highlight_1} 0.99{\ci{0.96}{1.01}} & \cellcolor{highlight_2} 0.87{\ci{0.67}{1.06}} & \cellcolor{highlight_0} 1.00{\ci{1.00}{1.00}} & \cellcolor{highlight_0} 1.00 & \cellcolor{highlight_0} 1.00{\ci{1.00}{1.00}} & \cellcolor{highlight_0} 1.00{\ci{1.00}{1.00}} \\
& & Acc & \cellcolor{highlight_0} 0.88{\ci{0.86}{0.90}} & \cellcolor{highlight_2} 0.64{\ci{0.55}{0.74}} & 0.55{\ci{0.55}{0.55}} & 0.60 & 0.63{\ci{0.58}{0.69}} & \cellcolor{highlight_1} 0.68{\ci{0.64}{0.72}} \\
& & WACC & \cellcolor{highlight_0} 0.88{\ci{0.86}{0.90}} & \cellcolor{highlight_2} 0.63{\ci{0.53}{0.73}} & 0.53{\ci{0.53}{0.53}} & 0.59 & 0.62{\ci{0.56}{0.68}} & \cellcolor{highlight_1} 0.67{\ci{0.63}{0.71}} \\
& & F1 & \cellcolor{highlight_0} 0.86{\ci{0.84}{0.89}} & \cellcolor{highlight_2} 0.44{\ci{0.22}{0.65}} & 0.13{\ci{0.13}{0.13}} & 0.29 & 0.38{\ci{0.23}{0.53}} & \cellcolor{highlight_1} 0.50{\ci{0.41}{0.59}} \\
& & TNR & \cellcolor{highlight_1} 0.99{\ci{0.97}{1.01}} & \cellcolor{highlight_2} 0.97{\ci{0.93}{1.00}} & \cellcolor{highlight_0} 1.00{\ci{1.00}{1.00}} & \cellcolor{highlight_0} 1.00 & \cellcolor{highlight_0} 1.00{\ci{1.00}{1.00}} & \cellcolor{highlight_0} 1.00{\ci{1.00}{1.00}} \\

\bottomrule
    \end{tabular}

\end{table}

\FloatBarrier
\subsection{Libero-Plus}
\label{sec:thresh-libero}
We summarize threshold-dependent results for the large-scale Libero-Plus simulation benchmark~\citep{fei25libero-plus} in \Cref{tab:eval-libero-thresh0.95}.
Due to the large number of tasks and the resulting high computational requirements, we report only a single seed per method.
However, given the large number of rollouts, we expect little variation across seeds.

\begin{table}[ht]
    \caption{Threshold-dependent detector performance on \textbf{Libero-Plus}, using a constant CP band at significance level $0.05$. The top three performing methods are highlighted.}
    \label{tab:eval-libero-thresh0.95}

    \centering
    \begin{tabular}{l l l *{6}{S[table-format=1.3]}}
        \toprule
         & & & {ACE} & {Diff} & {STAC} & {ACC} & {LLMD} & {FAIL} \\
        \midrule

\multirow{8}{*}{\texttt{Object}} & \multirow{4}{*}{X-VLA} & WACC & 0.831 & 0.763 & 0.813 & \cellcolor{highlight_1} 0.918 & \cellcolor{highlight_0} 0.994 & \cellcolor{highlight_2} 0.910 \\
& & F1 & 0.796 & 0.693 & 0.770 & \cellcolor{highlight_1} 0.901 & \cellcolor{highlight_0} 0.990 & \cellcolor{highlight_2} 0.885 \\
& & Prec & \cellcolor{highlight_1} 0.965 & 0.898 & \cellcolor{highlight_2} 0.952 & 0.922 & \cellcolor{highlight_0} 0.979 & 0.878 \\
& & Rec & 0.677 & 0.564 & 0.646 & \cellcolor{highlight_2} 0.880 & \cellcolor{highlight_0} 1.000 & \cellcolor{highlight_1} 0.893 \\
 & \multirow{4}{*}{$\pi_{0.5}$} & WACC & 0.595 & 0.751 & \cellcolor{highlight_1} 0.823 & \cellcolor{highlight_2} 0.771 & 0.624 & \cellcolor{highlight_0} 0.832 \\
& & F1 & 0.309 & 0.524 & \cellcolor{highlight_0} 0.650 & \cellcolor{highlight_1} 0.625 & 0.353 & \cellcolor{highlight_2} 0.594 \\
& & Prec & \cellcolor{highlight_2} 0.583 & 0.472 & \cellcolor{highlight_1} 0.599 & \cellcolor{highlight_0} 0.679 & 0.418 & 0.481 \\
& & Rec & 0.211 & \cellcolor{highlight_2} 0.589 & \cellcolor{highlight_1} 0.709 & 0.579 & 0.305 & \cellcolor{highlight_0} 0.775 \\

\midrule

\multirow{8}{*}{\texttt{Spatial}} & \multirow{4}{*}{X-VLA} & WACC & 0.795 & 0.774 & 0.807 & \cellcolor{highlight_1} 0.945 & \cellcolor{highlight_0} 0.964 & \cellcolor{highlight_2} 0.934 \\
& & F1 & 0.728 & 0.704 & 0.751 & \cellcolor{highlight_1} 0.903 & \cellcolor{highlight_0} 0.931 & \cellcolor{highlight_2} 0.884 \\
& & Prec & 0.769 & \cellcolor{highlight_0} 0.875 & \cellcolor{highlight_2} 0.858 & 0.833 & \cellcolor{highlight_1} 0.872 & 0.799 \\
& & Rec & 0.691 & 0.588 & 0.668 & \cellcolor{highlight_2} 0.985 & \cellcolor{highlight_0} 1.000 & \cellcolor{highlight_1} 0.989 \\
 & \multirow{4}{*}{$\pi_{0.5}$} & WACC & 0.738 & 0.780 & \cellcolor{highlight_2} 0.910 & \cellcolor{highlight_0} 0.936 & 0.695 & \cellcolor{highlight_1} 0.926 \\
& & F1 & 0.577 & 0.608 & \cellcolor{highlight_1} 0.705 & \cellcolor{highlight_0} 0.760 & 0.470 & \cellcolor{highlight_2} 0.692 \\
& & Prec & \cellcolor{highlight_0} 0.674 & \cellcolor{highlight_2} 0.610 & 0.582 & \cellcolor{highlight_1} 0.642 & 0.502 & 0.548 \\
& & Rec & 0.504 & 0.605 & \cellcolor{highlight_2} 0.895 & \cellcolor{highlight_1} 0.933 & 0.441 & \cellcolor{highlight_0} 0.941 \\

\midrule

\multirow{8}{*}{\texttt{Goal}} & \multirow{4}{*}{X-VLA} & WACC & 0.656 & 0.756 & 0.800 & \cellcolor{highlight_0} 0.973 & \cellcolor{highlight_1} 0.969 & \cellcolor{highlight_2} 0.965 \\
& & F1 & 0.507 & 0.677 & 0.741 & \cellcolor{highlight_0} 0.955 & \cellcolor{highlight_1} 0.943 & \cellcolor{highlight_2} 0.940 \\
& & Prec & 0.716 & \cellcolor{highlight_1} 0.896 & 0.838 & \cellcolor{highlight_0} 0.923 & 0.893 & \cellcolor{highlight_2} 0.894 \\
& & Rec & 0.392 & 0.543 & 0.665 & \cellcolor{highlight_2} 0.988 & \cellcolor{highlight_0} 1.000 & \cellcolor{highlight_1} 0.990 \\
 & \multirow{4}{*}{$\pi_{0.5}$} & WACC & 0.701 & 0.724 & \cellcolor{highlight_2} 0.811 & \cellcolor{highlight_1} 0.902 & 0.579 & \cellcolor{highlight_0} 0.912 \\
& & F1 & 0.561 & 0.605 & \cellcolor{highlight_2} 0.736 & \cellcolor{highlight_0} 0.841 & 0.293 & \cellcolor{highlight_1} 0.835 \\
& & Prec & 0.820 & \cellcolor{highlight_0} 0.835 & \cellcolor{highlight_1} 0.835 & \cellcolor{highlight_2} 0.833 & 0.575 & 0.787 \\
& & Rec & 0.426 & 0.474 & \cellcolor{highlight_2} 0.658 & \cellcolor{highlight_1} 0.849 & 0.197 & \cellcolor{highlight_0} 0.889 \\

\midrule

\multirow{8}{*}{\texttt{10}} & \multirow{4}{*}{X-VLA} & WACC & 0.672 & \cellcolor{highlight_2} 0.799 & 0.656 & 0.699 & \cellcolor{highlight_0} 0.990 & \cellcolor{highlight_1} 0.821 \\
& & F1 & 0.536 & \cellcolor{highlight_2} 0.755 & 0.502 & 0.583 & \cellcolor{highlight_0} 0.986 & \cellcolor{highlight_1} 0.786 \\
& & Prec & 0.845 & \cellcolor{highlight_2} 0.863 & 0.846 & \cellcolor{highlight_1} 0.897 & \cellcolor{highlight_0} 0.973 & 0.841 \\
& & Rec & 0.392 & \cellcolor{highlight_2} 0.671 & 0.357 & 0.431 & \cellcolor{highlight_0} 1.000 & \cellcolor{highlight_1} 0.737 \\
 & \multirow{4}{*}{$\pi_{0.5}$} & WACC & 0.748 & 0.739 & \cellcolor{highlight_2} 0.781 & \cellcolor{highlight_1} 0.864 & 0.609 & \cellcolor{highlight_0} 0.873 \\
& & F1 & 0.632 & 0.616 & \cellcolor{highlight_2} 0.686 & \cellcolor{highlight_0} 0.767 & 0.374 & \cellcolor{highlight_1} 0.749 \\
& & Prec & \cellcolor{highlight_1} 0.731 & 0.703 & \cellcolor{highlight_0} 0.777 & \cellcolor{highlight_2} 0.717 & 0.594 & 0.648 \\
& & Rec & 0.557 & 0.548 & \cellcolor{highlight_2} 0.614 & \cellcolor{highlight_1} 0.824 & 0.273 & \cellcolor{highlight_0} 0.889 \\

\bottomrule
    \end{tabular}

\end{table}

\FloatBarrier
\section{Qualitative Results}
\label{appendix:qualitative_results}

We provide qualitative examples to illustrate ACC and LLMD detection behavior during deployment. Section \ref{subsec:qualitative_successful_detection} provides examples for accurate detections across different tasks and models while Section \ref{subsec:qualitative_failed_detection} illustrates failure modes we observe during deployment. Additional qualitative analysis with full video rollouts can be found in the project page \url{https://vla-fail.github.io/vla-fail-2026/}.

\subsection{Real-World Successful Failure Detection}
\label{subsec:qualitative_successful_detection}

\begin{figure}[!htbp]
    \centering
    \setlength{\tabcolsep}{2pt}
    \renewcommand{\arraystretch}{0.9}

    \begin{subfigure}[t]{\linewidth}
        \centering
        \caption{$\pi_{0.5}$ Blocks}
        \begin{tabular}{c ccccc}
            & {\small\boldmath$\mathbf{t_1}$} & {\small\boldmath$\mathbf{t_2}$} & {\small\boldmath$\mathbf{t_3}$} & {\small\boldmath$\mathbf{t_4}$} & {\small\boldmath$\mathbf{t_5}$} \\

            \detrowlabelACC{ACC}
            &
            \includegraphics[width=0.18\linewidth]{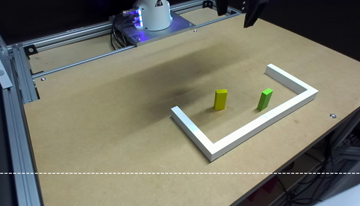} &
            \includegraphics[width=0.18\linewidth]{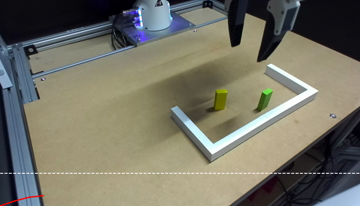} &
            \includegraphics[width=0.18\linewidth]{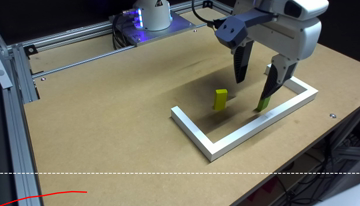} &
            \fcolorbox{red}{white}{\includegraphics[width=0.165\linewidth]{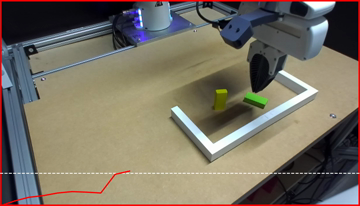}} &
            \fcolorbox{red}{white}{\includegraphics[width=0.165\linewidth]{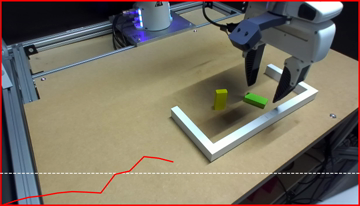}} \\[6pt]

            \detrowlabelLLMD{LLMD}
            &
            \includegraphics[width=0.18\linewidth]{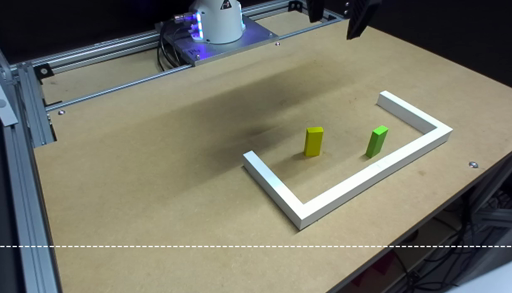} &
            \includegraphics[width=0.18\linewidth]{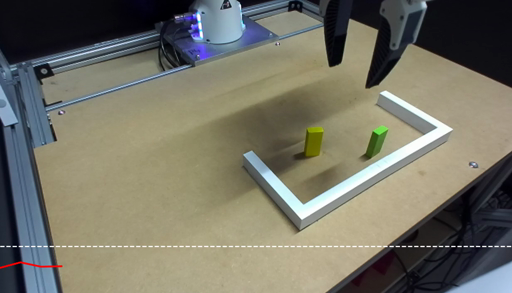} &
            \includegraphics[width=0.18\linewidth]{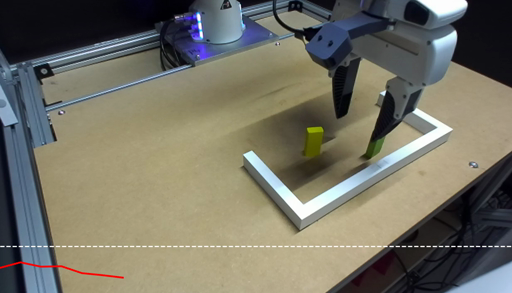} &
            \fcolorbox{red}{white}{\includegraphics[width=0.165\linewidth]{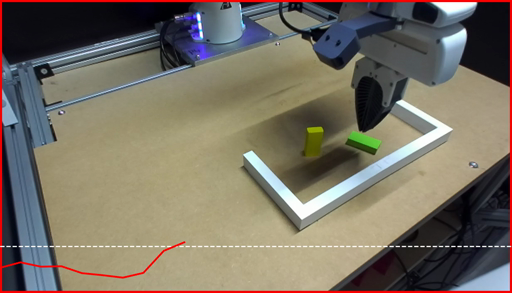}} &
            \fcolorbox{red}{white}{\includegraphics[width=0.165\linewidth]{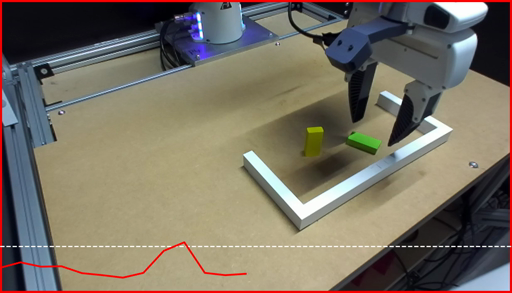}}
        \end{tabular}
    \end{subfigure}

    \begin{subfigure}[t]{\linewidth}
        \centering
        \caption{X-VLA Stack T}
        \begin{tabular}{c ccccc}
            & {\small\boldmath$\mathbf{t_1}$} & {\small\boldmath$\mathbf{t_2}$} & {\small\boldmath$\mathbf{t_3}$} & {\small\boldmath$\mathbf{t_4}$} & {\small\boldmath$\mathbf{t_5}$} \\

            \makebox[0.01\linewidth][c]{%
                \raisebox{1.5em}{\rotatebox{90}{\textbf{ACC}}}%
            }%
            &
            \includegraphics[width=0.18\linewidth]{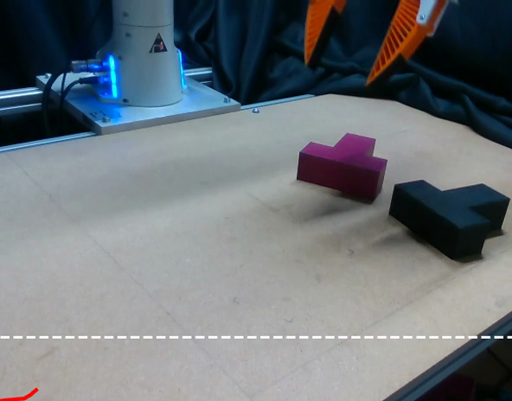} &
            \includegraphics[width=0.18\linewidth]{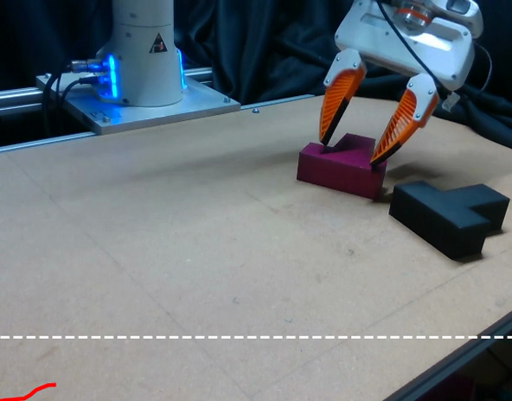} &
            \includegraphics[width=0.18\linewidth]{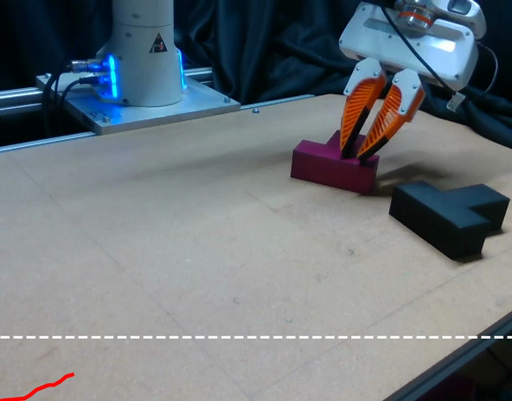} &
            \includegraphics[width=0.18\linewidth]{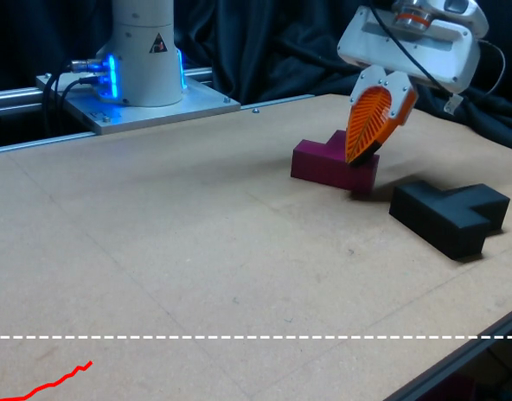} &
            \fcolorbox{red}{white}{\includegraphics[width=0.165\linewidth]{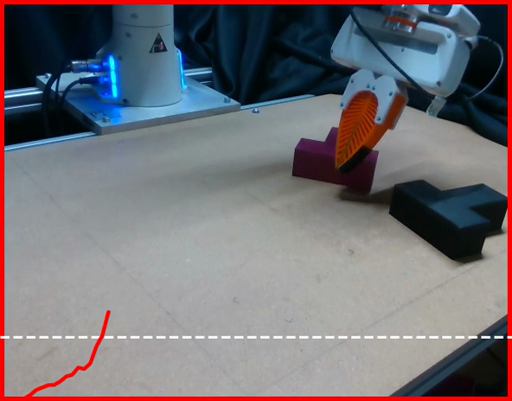}} \\[6pt]

            \makebox[0.01\linewidth][c]{%
                \raisebox{1.0em}{\rotatebox{90}{\textbf{LLMD}}}%
            }%
            &
            \includegraphics[width=0.18\linewidth]{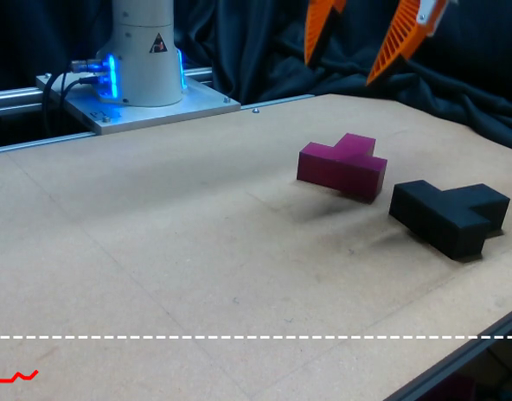} &
            \includegraphics[width=0.18\linewidth]{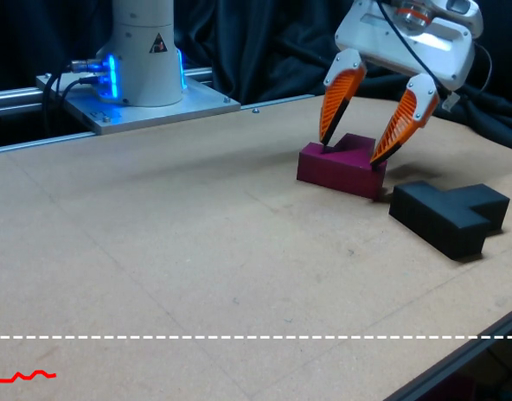} &
            \includegraphics[width=0.18\linewidth]{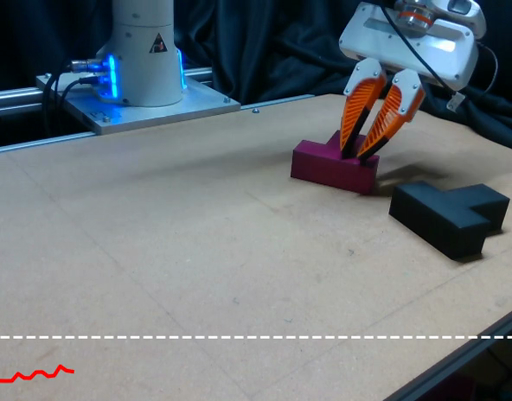} &
            \includegraphics[width=0.18\linewidth]{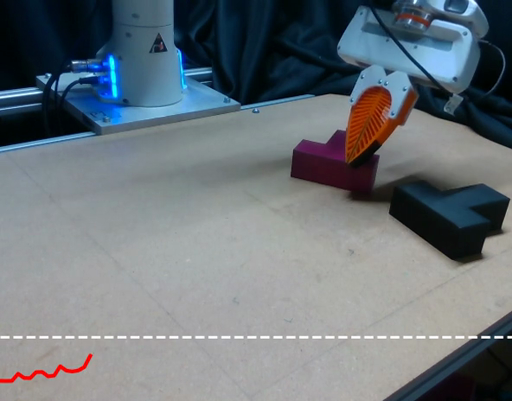} &
            \fcolorbox{red}{white}{\includegraphics[width=0.165\linewidth]{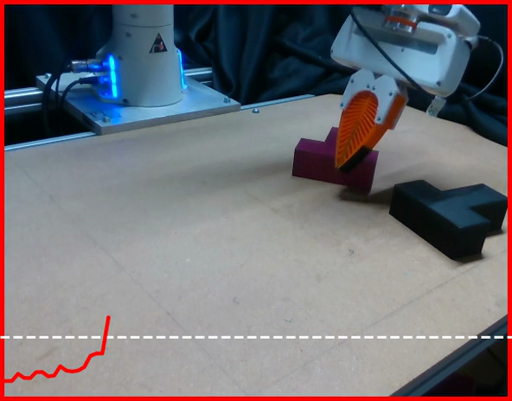}}
        \end{tabular}
    \end{subfigure}

    \begin{subfigure}[t]{\linewidth}
        \centering
        \caption{$\pi_{0.5}$ Cups}
        \begin{tabular}{c ccccc}
            & {\small\boldmath$\mathbf{t_1}$} & {\small\boldmath$\mathbf{t_2}$} & {\small\boldmath$\mathbf{t_3}$} & {\small\boldmath$\mathbf{t_4}$} & {\small\boldmath$\mathbf{t_5}$} \\

            \detrowlabelACC{ACC}
            &
            \includegraphics[width=0.18\linewidth]{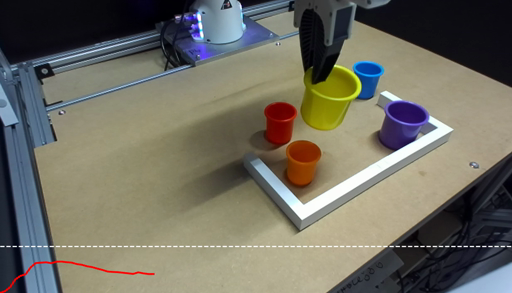} &
            \includegraphics[width=0.18\linewidth]{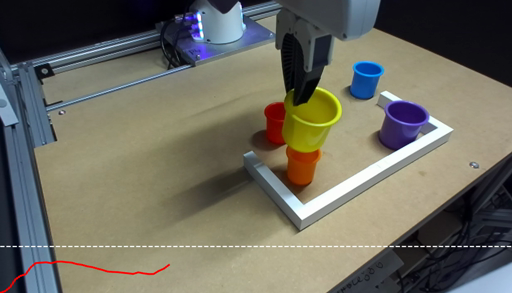} &
            \includegraphics[width=0.18\linewidth]{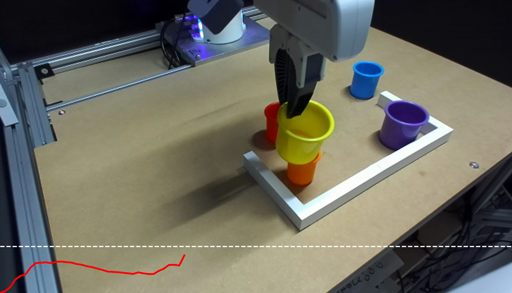} &
            \includegraphics[width=0.18\linewidth]{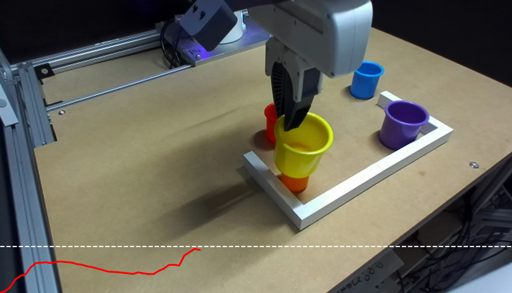} &
            \fcolorbox{red}{white}{\includegraphics[width=0.165\linewidth]{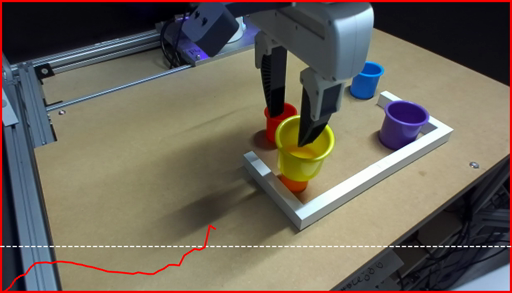}} \\[6pt]

            \detrowlabelLLMD{LLMD}
            &
            \includegraphics[width=0.18\linewidth]{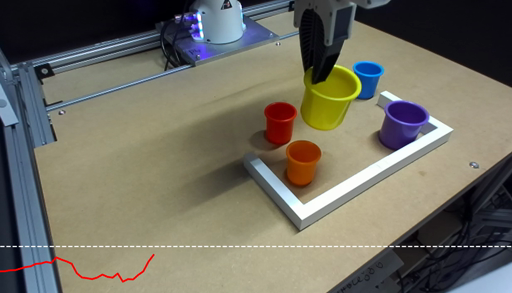} &
            \fcolorbox{red}{white}{\includegraphics[width=0.165\linewidth]{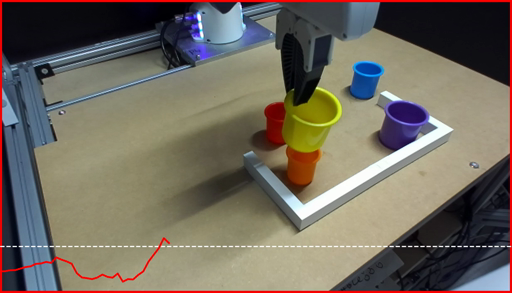}} &
            \fcolorbox{red}{white}{\includegraphics[width=0.165\linewidth]{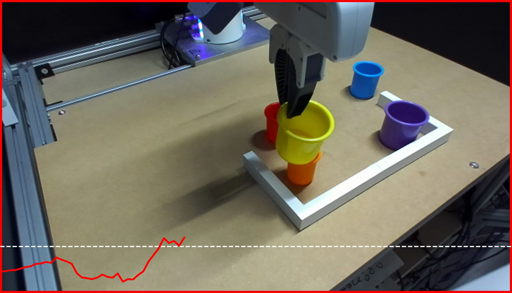}} &
            \fcolorbox{red}{white}{\includegraphics[width=0.165\linewidth]{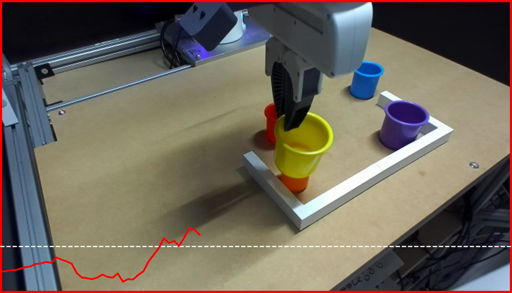}} &
            \fcolorbox{red}{white}{\includegraphics[width=0.165\linewidth]{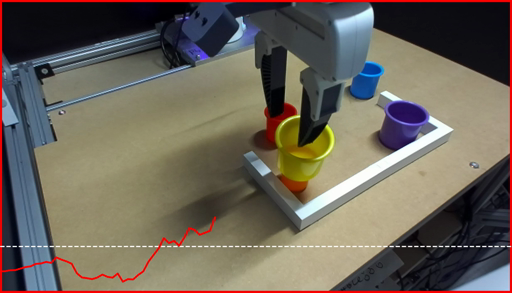}}
        \end{tabular}
    \end{subfigure}

    \caption{Qualitative failure detection over time. Each task is shown with two detector rows: ACC on top and LLMD below. Columns correspond to consecutive timesteps. Red boxes mark timesteps at which the corresponding detector triggers.}
    \label{fig:qualitative_successful_detection}
\end{figure}

\Cref{fig:qualitative_successful_detection} shows representative deployment rollouts in which \textsc{VLA-FAIL} successfully detects task failures.
The examples illustrate accurate triggers from both ACC and LLMD, with each detector capturing different failure modes.

\FloatBarrier
\subsection{Real-World VLA-FAIL Failure Cases}
\label{subsec:qualitative_failed_detection}

\begin{figure}[!htbp]
    \centering
    \setlength{\tabcolsep}{2pt}
    \renewcommand{\arraystretch}{0.9}

    \begin{subfigure}[t]{\linewidth}
        \centering
        \caption{$\pi_{0.5}$ Blocks}
        \label{subfig:qualitative_failed_detection_acc_delay}
        \begin{tabular}{c ccccc}
            & {\small\boldmath$\mathbf{t_1}$} & {\small\boldmath$\mathbf{t_2}$} & {\small\boldmath$\mathbf{t_3}$} & {\small\boldmath$\mathbf{t_4}$} & {\small\boldmath$\mathbf{t_5}$} \\

            \detrowlabelACC{ACC}
            &
            \includegraphics[width=0.18\linewidth]{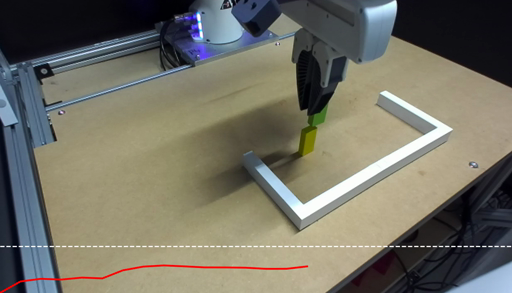} &
            \includegraphics[width=0.18\linewidth]{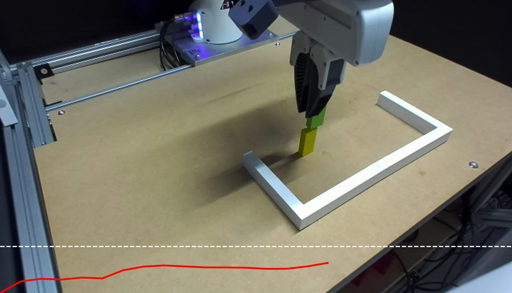} &
            \includegraphics[width=0.18\linewidth]{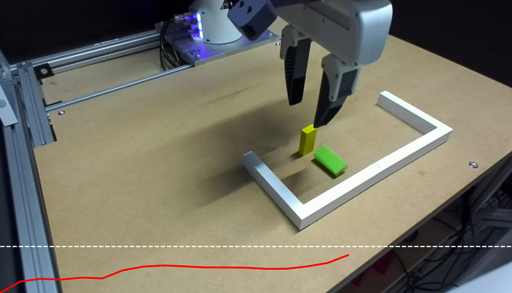} &
            \includegraphics[width=0.18\linewidth]{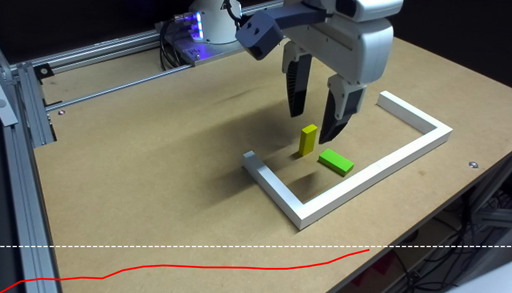} &
            \fcolorbox{red}{white}{\includegraphics[width=0.165\linewidth]{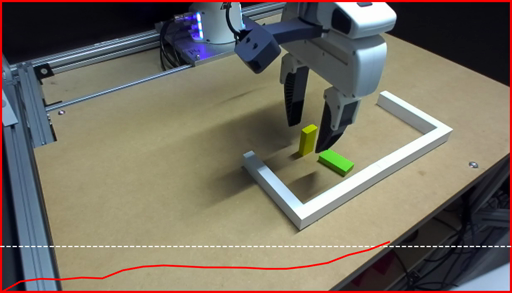}} \\[6pt]

            \detrowlabelLLMD{LLMD}
            &
            \fcolorbox{red}{white}{\includegraphics[width=0.165\linewidth]{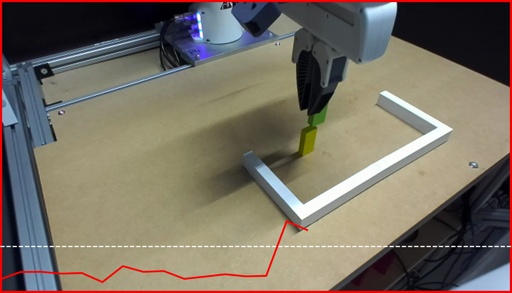}} &
            \fcolorbox{red}{white}{\includegraphics[width=0.165\linewidth]{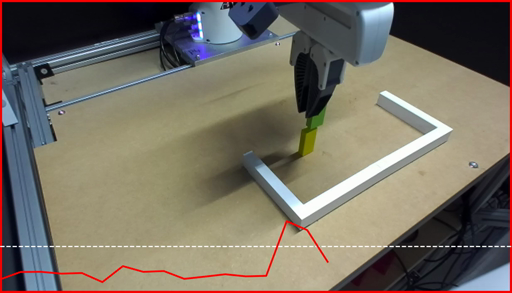}} &
            \fcolorbox{red}{white}{\includegraphics[width=0.165\linewidth]{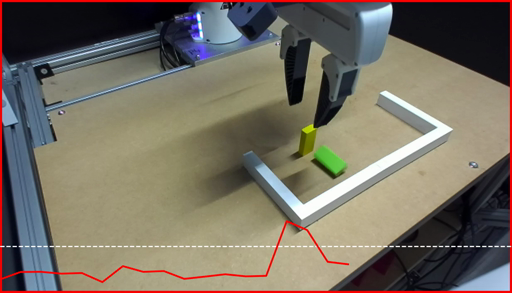}} &
            \fcolorbox{red}{white}{\includegraphics[width=0.165\linewidth]{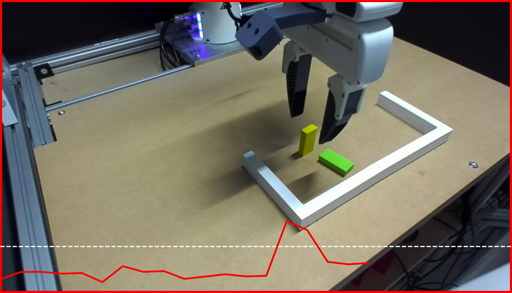}} &
            \fcolorbox{red}{white}{\includegraphics[width=0.165\linewidth]{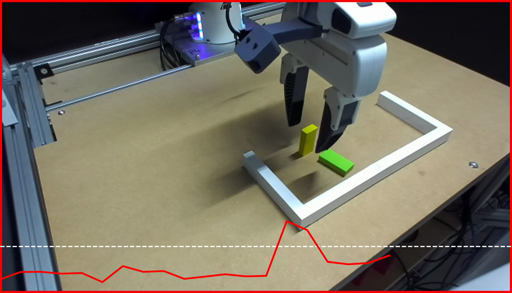}}
        \end{tabular}
    \end{subfigure}

    \begin{subfigure}[t]{\linewidth}
        \centering
        \caption{$\pi_{0.5}$ Drawer}
        \label{subfig:qualitative_failed_detection_llmd_no_detection}
        \begin{tabular}{c ccccc}
            & {\small\boldmath$\mathbf{t_1}$} & {\small\boldmath$\mathbf{t_2}$} & {\small\boldmath$\mathbf{t_3}$} & {\small\boldmath$\mathbf{t_4}$} & {\small\boldmath$\mathbf{t_5}$} \\

            \detrowlabelACC{ACC}
            &
            \includegraphics[width=0.18\linewidth]{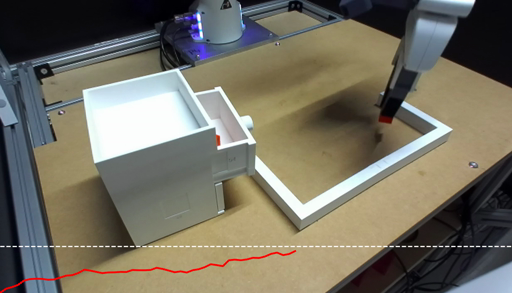} &
            \includegraphics[width=0.18\linewidth]{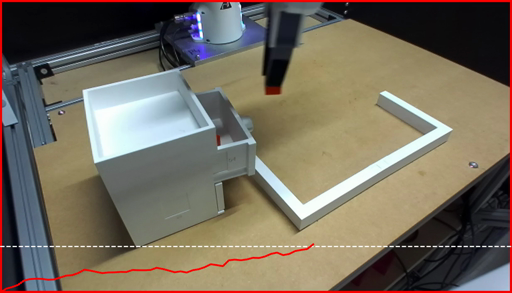} &
            \includegraphics[width=0.18\linewidth]{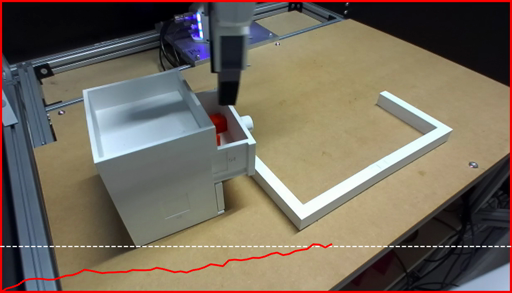} &
            \fcolorbox{red}{white}{\includegraphics[width=0.165\linewidth]{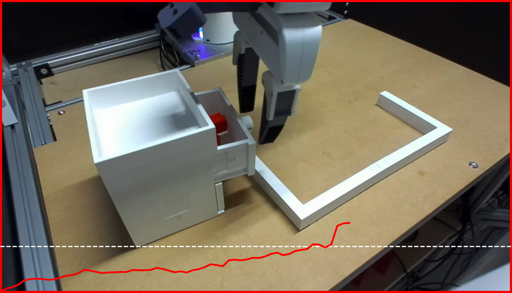}} &
            \fcolorbox{red}{white}{\includegraphics[width=0.165\linewidth]{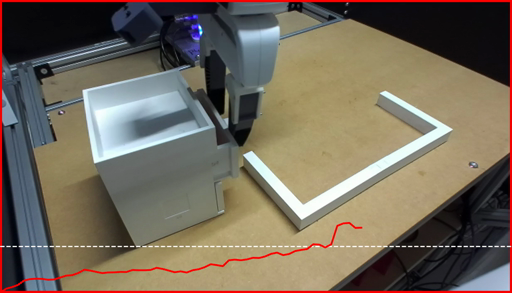}} \\[6pt]

            \detrowlabelLLMD{LLMD}
            &
            \includegraphics[width=0.18\linewidth]{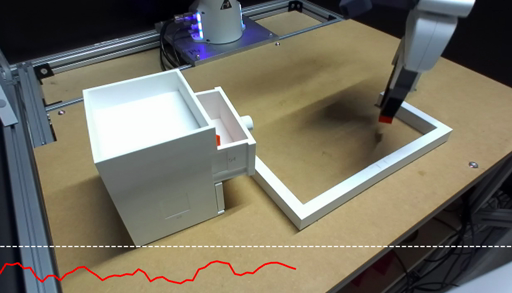} &
            \includegraphics[width=0.18\linewidth]{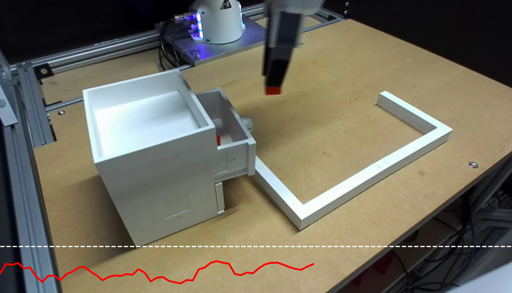} &
            \includegraphics[width=0.18\linewidth]{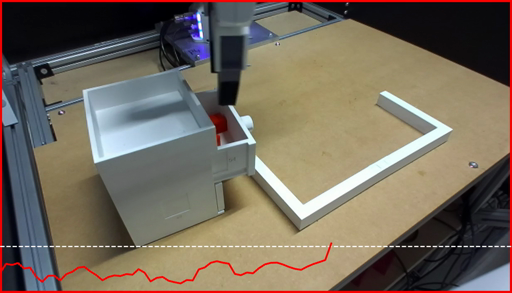} &
            \fcolorbox{red}{white}{\includegraphics[width=0.165\linewidth]{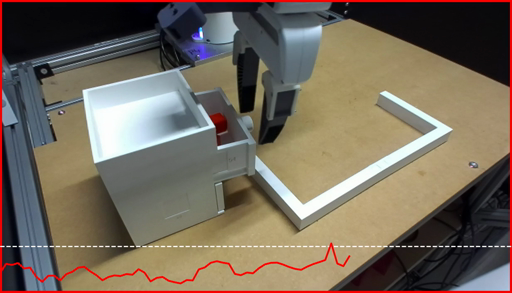}} &
            \includegraphics[width=0.18\linewidth]{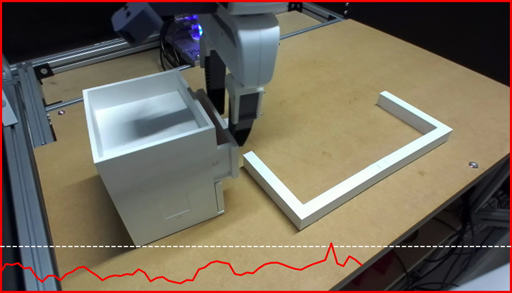}
        \end{tabular}
    \end{subfigure}

    \begin{subfigure}[t]{\linewidth}
        \centering
        \caption{X-VLA Stack T}
        \label{subfig:qualitative_failed_detection_both_no_detection}
        \begin{tabular}{c ccccc}
            & {\small\boldmath$\mathbf{t_1}$} & {\small\boldmath$\mathbf{t_2}$} & {\small\boldmath$\mathbf{t_3}$} & {\small\boldmath$\mathbf{t_4}$} & {\small\boldmath$\mathbf{t_5}$} \\

            \makebox[0.01\linewidth][c]{%
                \raisebox{1.5em}{\rotatebox{90}{\textbf{ACC}}}%
            }%
            &
            \includegraphics[width=0.18\linewidth]{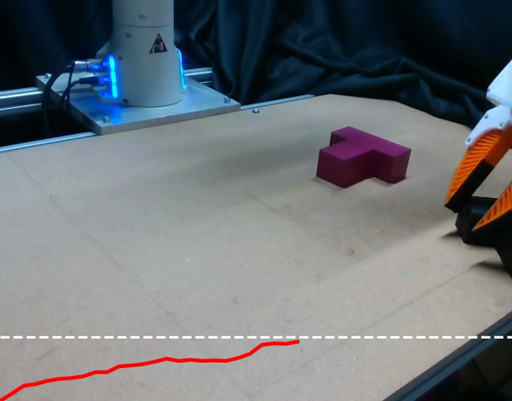} &
            \includegraphics[width=0.18\linewidth]{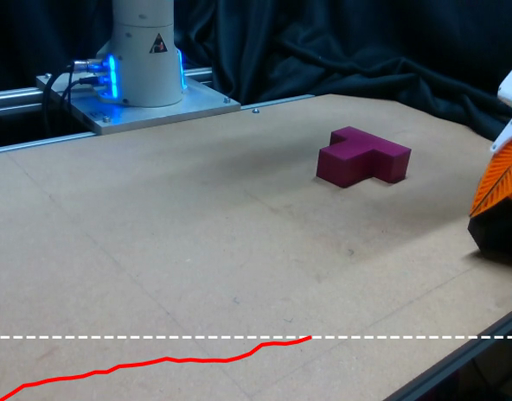} &
            \includegraphics[width=0.18\linewidth]{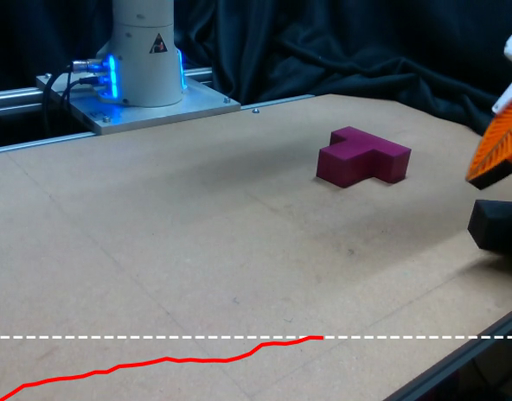} &
            \includegraphics[width=0.18\linewidth]{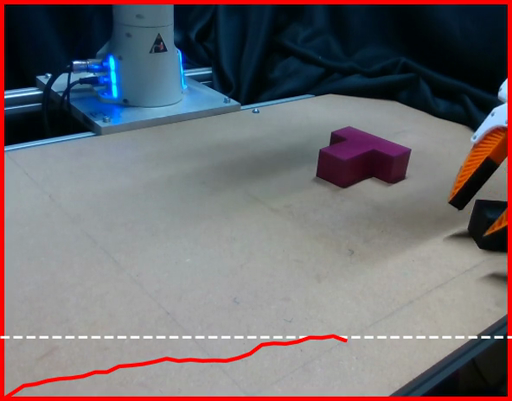} &
            \includegraphics[width=0.18\linewidth]{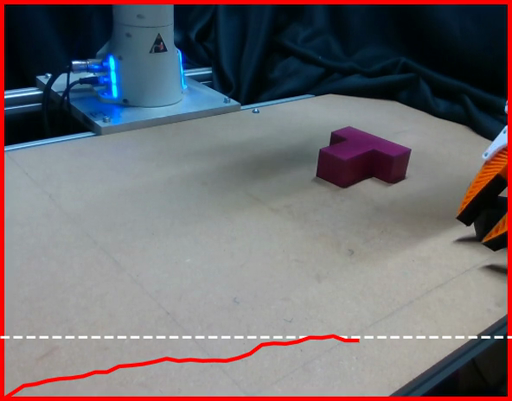} \\[6pt]

            \makebox[0.01\linewidth][c]{%
                \raisebox{1.0em}{\rotatebox{90}{\textbf{LLMD}}}%
            }%
            &
            \includegraphics[width=0.18\linewidth]{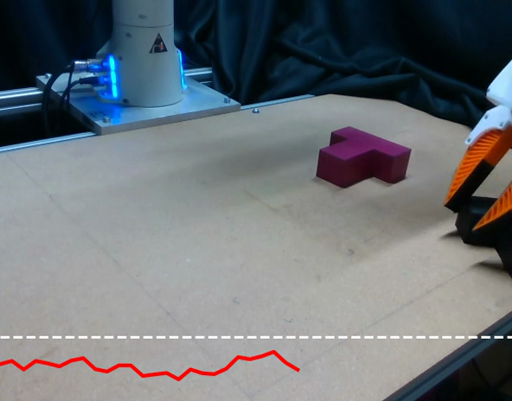} &
            \includegraphics[width=0.18\linewidth]{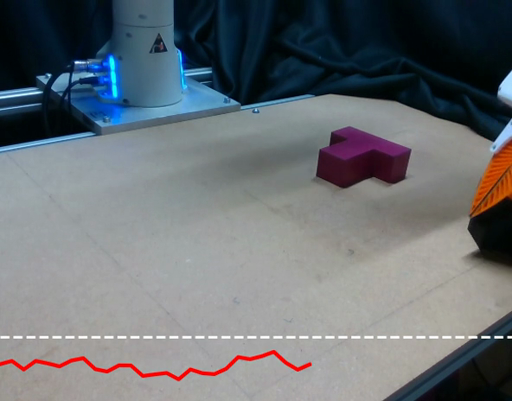} &
            \includegraphics[width=0.18\linewidth]{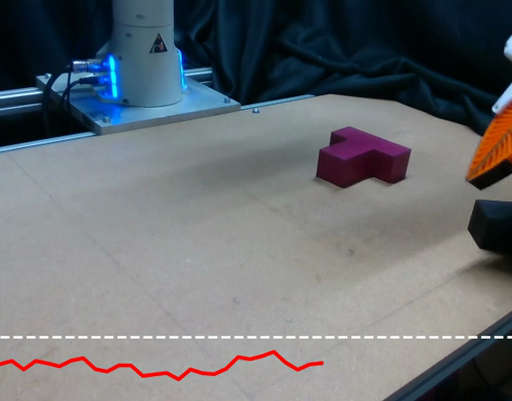} &
            \includegraphics[width=0.18\linewidth]{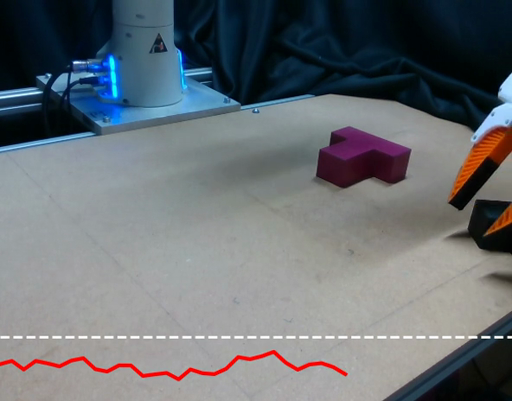} &
            \includegraphics[width=0.18\linewidth]{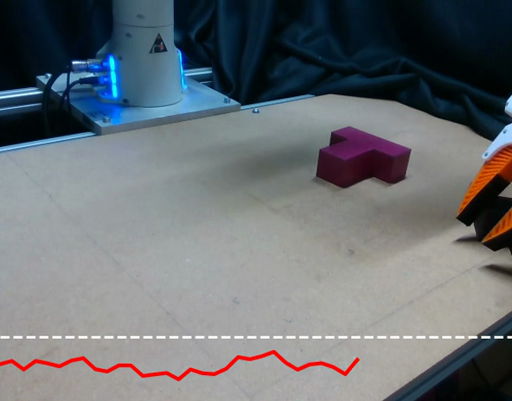}
        \end{tabular}
    \end{subfigure}

    \caption{Qualitative failure detection over time. Each task is shown with two detector rows: ACC on top and LLMD below. Columns correspond to consecutive timesteps. Red boxes mark timesteps at which the corresponding detector triggers.}
    \label{fig:qualitative_failed_detection}
\end{figure}

\Cref{fig:qualitative_failed_detection} shows representative failure cases of \textsc{VLA-FAIL}.
First, ACC can detect failures with a delay because its smoothed score requires persistent inconsistency between overlapping action chunks before crossing the threshold; in such cases, LLMD may provide an earlier alarm, as shown in \Cref{subfig:qualitative_failed_detection_acc_delay}.
Second, LLMD can miss failures when the policy's internal features remain close to the finetuning distribution, while ACC still detects the resulting inconsistent replanning, as shown in \Cref{subfig:qualitative_failed_detection_llmd_no_detection}.
Third, some failures remain both feature-consistent and action-consistent for too long, causing both detectors to miss or delay the alarm, as shown in \Cref{subfig:qualitative_failed_detection_both_no_detection}.

\FloatBarrier
\section{Hyperparameters}
We specify the chosen hyperparameters for VLA finetuning in \Cref{tab:finetuning_hyperparameters}, for failure detection on real-world tasks in \Cref{tab:model_hyperparameters_real}, and for failure detection on Libero-Plus in \Cref{tab:model_hyperparameters_libero}.
Failure detection hyperparameters were tuned per VLA and separately for simulated and real-world tasks, but not per task.
We optimized ACC's minimum velocity and smoothing factor through a small grid search ($v_\text{min} \in [10^{-4}, 0.1], \alpha \in [0.1, 1.0]$) on the \texttt{Cups} task and \texttt{Object} suite.

Below, we summarize additional observations and recommended practices for hyperparameter tuning.

\paragraph{VLA Finetuning.}
We leave most hyperparameters at their default values as recommended by \citet{intelligence2025pi05} and \citet{zheng2026xvla}.
For $\pi_{0.5}$, there is a strong dependence between final model performance and batch size, where a smaller batch size improves success rates.
For X-VLA, we find that training for more steps than recommended by \citet{zheng2026xvla} improves success rates.
We train all policies on four NVIDIA A100 GPUs.

\paragraph{ACC.}
We tune the minimal velocity $v_\text{min}$ for ACC separately for the real-world experiments and \texttt{Libero-Plus}, but not per task or per VLA.
In general, lower minimum velocities work well in simulation, where sensor noise is low, and policies tend to act more smoothly.
For the real-world results, we use a higher minimum velocity and find that lower values do not work well.
We generally recommend to tune this value by visualizing the ACC score over time on calibration rollouts and increasing it until the score does not inflate excessively for a robot that is nearly stationary after completing a task.

\paragraph{LLMD.}
The only hyperparameter of LLMD is the regularization penalty $\lambda$, which does not require tuning.
A small regularization penalty $\lambda = 10^{-6}$ is recommended, as it is only needed to ensure invertibility of the covariance matrix, not to regularize the Mahalanobis distance.

\paragraph{Baselines.}
We find that ACE's~\citep{romer2026fiper} cell size factor $\alpha$ needs to be tuned separately for simulation and real-world experiments, and for each VLA.
X-VLA requires significantly smaller values of $\alpha$ than $\pi_{0.5}$.
We do not tune $\alpha$ per task to ensure a fair comparison with other methods.
For each task, we use $20$ successful rollouts as a calibration set to determine the cell sizes based on $\alpha$, as described by \citet{romer2026fiper}.
For STAC~\citep{AgiaSinhaEtAl2024}, we use the maximum mean discrepancy metric with an RBF kernel and the median heuristic for the kernel's bandwidth, as recommended in the appendix of \citet{AgiaSinhaEtAl2024}.
We evaluate ACE and STAC with 32 action samples, which roughly corresponds to the smallest number of samples that \citet{AgiaSinhaEtAl2024} and \citet{romer2026fiper} use in their experiments.

We adapt the diffusion loss metric proposed by~\citet{lee2025diff} for flow matching policies by replacing the diffusion loss with the flow matching loss, yielding our Diff baseline.
\citet{lee2025diff} always use a large batch size of 512 samples for loss calculation, which we find to be intractably large for VLAs.
We therefore use only 32 samples, which aligns Diff with ACE and STAC.

\begin{table}[!htbp]
    \centering
    \caption{Model-specific finetuning parameters shared across tasks.}
    \label{tab:finetuning_hyperparameters}
    \tablesize
    \setlength{\tabcolsep}{4pt}
    \renewcommand{\arraystretch}{1.05}
    \begin{tabular*}{\textwidth}{@{\extracolsep{\fill}}lcc@{}}
        \toprule
        \textbf{Hyperparameter} & \boldmath$\pi_{0.5}$ & \textbf{X-VLA} \\
        \midrule
        Batch size & 16 & 32 \\
        \# Gradient Steps & 30k & 100k \\
        Action representation & Delta Pos + Quat & Delta EE6D \\
        Action chunk length $H$ & 20 & 30 \\
        \bottomrule
    \end{tabular*}
\end{table}

\begin{table}[!htbp]
    \centering
    \caption{Model-specific hyperparameters used for VLA-FAIL and baselines in the real world, shared across tasks.}
    \label{tab:model_hyperparameters_real}
    \tablesize
    \setlength{\tabcolsep}{4pt}
    \renewcommand{\arraystretch}{1.05}
    \begin{tabular*}{\textwidth}{@{\extracolsep{\fill}}lcc@{}}
        \toprule
        \textbf{Hyperparameter} & \boldmath$\pi_{0.5}$ & \textbf{X-VLA} \\
        \midrule
        LLMD feature dimension $\dim(f)$ & 1024 & 1024 \\
        LLMD flow timestep & $t=0$ & $t=0$ \\
        LLMD prior noise & fixed & fixed \\
        LLMD token aggregation & max & max \\
        ACC smoothing factor $\alpha$ & 0.9 & 0.9 \\
        ACC minimum velocity $v_{\min}$ & 0.01 & 0.01 \\
        ACE $\alpha$ & 0.025 & $5\cdot 10^{-4}$ \\
        Threshold type & constant & constant \\
        Calibration method & conformal & conformal \\
        \bottomrule
    \end{tabular*}
\end{table}

\begin{table}[!htbp]
    \centering
    \caption{Model-specific hyperparameters used for VLA-FAIL on Libero-Plus.}
    \label{tab:model_hyperparameters_libero}
    \tablesize
    \setlength{\tabcolsep}{4pt}
    \renewcommand{\arraystretch}{1.05}
    \begin{tabular*}{\textwidth}{@{\extracolsep{\fill}}lcc@{}}
        \toprule
        \textbf{Hyperparameter} & \boldmath$\pi_{0.5}$ & \textbf{X-VLA} \\
        \midrule
        LLMD feature dimension $\dim(f)$ & 1024 & 1024 \\
        LLMD flow timestep & $t=0$ & $t=0$ \\
        LLMD prior noise & fixed & fixed \\
        LLMD token aggregation & max & max \\
        ACC smoothing factor $\alpha$ & 0.9 & 0.9 \\
        ACC minimum velocity $v_{\min}$ & $10^{-3}$ & $10^{-3}$ \\
        ACE $\alpha$ & 0.1 & $10^{-4}$ \\
        Threshold type & constant & constant \\
        Calibration method & conformal & conformal \\
        \bottomrule
    \end{tabular*}
\end{table}

\FloatBarrier
\section{Additional Experiment Details}

Here, we provide additional details on the experiments.
Note that calibration rollouts are only used for the threshold-dependent results, taken and removed from the set of successful test rollouts.
There are no calibration sets for the threshold-independent results, maximizing the amount of available data.
ACC's action grid bounds require per-task tuning on a separate split, for which we reuse $20$ of the successful test rollouts.
This evaluation protocol is therefore slightly advantageous for ACC, as the grid bounds are optimized on rollouts that the method is tested on.

\subsection{Failure Prediction Details}
\Cref{tab:real_world_task_details_pi} and \Cref{tab:real_world_task_details_xvla} summarize experiment details for $\pi_{0.5}$ and X-VLA for the real-world tasks.
We use all available rollouts for the experiments, leading to a varying number of in-distribution (ID) and out-of-distribution (OOD) rollouts per task and model.
Note that the low success rates are largely due to OOD setups, where most policies fail.
On ID data, policies achieve significantly higher success rates.
For the threshold-independent experiments, we do not use a separate calibration set; instead, we evaluate on a single, large rollout dataset spanning calibration, ID, and OOD rollouts.

\begin{table}[!htbp]
    \centering
    \caption{Real-world failure prediction details for $\pi_{0.5}$. The VLA receives only the gripper and right overhead camera views as observations, since the public version of $\pi_{0.5}$ supports only two camera images.}
    \label{tab:real_world_task_details_pi}
    \tablesize
    \setlength{\tabcolsep}{3pt}
    \renewcommand{\arraystretch}{1.05}
    \begin{tabular*}{\textwidth}{@{\extracolsep{\fill}}lcccccc@{}}
        \toprule
        \textbf{Task}
        & \textbf{Blocks}
        & \textbf{Drawer}
        & \textbf{Cups}
        & \textbf{Kitchen}
        & \textbf{Stack T}
        & \textbf{Mixer} \\
        \midrule
        Observation dim & $[2,480,640]$ & $[2,480,640]$ & $[2,480,640]$ & $[2,480,640]$ & $[2,480,640]$ & $[2,480,640]$ \\
        Proprio dim & 8 & 8 & 8 & 8 & 8 & 8 \\
        Action dim & 8 & 8 & 8 & 8 & 8 & 8 \\
        Executed Actions $R$ & 10 & 10 & 5 & 10 & 10 & 10 \\
        ACC Overlap & 10 & 10 & 15 & 10 & 10 & 10 \\
        \# Finetuning demos & 100 & 100 & 100 & 100 & 100 & 100 \\
        \# Calibration rollouts & 20 & 20 & 20 & 20 & 20 & 20 \\
        \# ID test rollouts & 40 & 80 & 181 & 41 & 50 & 50 \\
        \# OOD test rollouts & 35 & 41 & 54 & 31 & 51 & 30 \\
        Success rate & 0.56 & 0.72 & 0.69 & 0.47 & 0.32 & 0.64 \\
        \bottomrule
    \end{tabular*}
\end{table}

\begin{table}[!htbp]
    \centering
    \caption{Real-world failure prediction details for X-VLA.}
    \label{tab:real_world_task_details_xvla}
    \tablesize
    \setlength{\tabcolsep}{3pt}
    \renewcommand{\arraystretch}{1.05}
    \begin{tabular*}{\textwidth}{@{\extracolsep{\fill}}lcccccc@{}}
        \toprule
        \textbf{Task}
        & \textbf{Blocks}
        & \textbf{Drawer}
        & \textbf{Cups}
        & \textbf{Kitchen}
        & \textbf{Stack T}
        & \textbf{Mixer} \\
        \midrule
        Observation dim & $[3,480,640]$ & $[3,480,640]$ & $[3,480,640]$ & $[3,480,640]$ & $[3,480,640]$ & $[3,480,640]$ \\
        Proprio dim & 8 & 8 & 8 & 8 & 8 & 8 \\
        Action dim & 8 & 8 & 8 & 8 & 8 & 8 \\
        Executed Actions $R$ & 10 & 10 & 10 & 10 & 10 & 10 \\
        ACC Overlap & 10 & 10 & 20 & 10 & 10 & 10 \\
        \# Finetuning demos & 100 & 100 & 100 & 100 & 100 & 100 \\
        \# Calibration rollouts & 20 & 20 & 20 & 20 & 20 & 20 \\
        \# ID test rollouts & 50 & 49 & 65 & 41 & 53 & 51 \\
        \# OOD test rollouts & 34 & 38 & 40 & 21 & 51 & 30 \\
        Success rate & 0.46 & 0.71 & 0.53 & 0.55 & 0.21 & 0.34 \\
        \bottomrule
    \end{tabular*}
\end{table}

\begin{table}[!htbp]
    \centering
    \caption{Libero-Plus failure prediction details. Chunk length and success rate are in format $\pi_{0.5}$/X-VLA.}
    \label{tab:simulation_task_details}
    \tablesize
    \setlength{\tabcolsep}{3pt}
    \renewcommand{\arraystretch}{1.05}
    \begin{tabular*}{\textwidth}{@{\extracolsep{\fill}}lcccc@{}}
        \toprule
        \textbf{Suite}
        & \textbf{Object}
        & \textbf{Spatial}
        & \textbf{Goal}
        & \textbf{10} \\
        \midrule
        Observation dim & [1, 256, 256] & [1, 256, 256] & [1, 256, 256] & [1, 256, 256] \\
        Proprio dim & 8 & 8 & 8 & 8 \\
        Action dim & 7 & 7 & 7 & 7 \\
        Chunk Length & 5 / 20 & 5 / 20 & 5 / 20 & 5 / 20 \\
        \# Calibration rollouts & 100 & 100 & 100 & 100 \\
        \# Rollouts & 2413 & 2301 & 2463 & 2431 \\
        Success rate & 0.88 / 0.63 & 0.90 / 0.67 & 0.79 / 0.66 & 0.77 / 0.59 \\
        \bottomrule
    \end{tabular*}
\end{table}

\newpage
\subsection{Real-World Environments}

Our real-world benchmark contains six diverse manipulation tasks requiring high precision (\texttt{Blocks}, \texttt{Stack T}), multi-modal behavior (\texttt{Cups}), long-horizon execution (\texttt{Kitchen}, \texttt{Drawer}), and language conditioning (\texttt{Mixer}).
We evaluate roughly $80$ rollouts per task and report confidence intervals over three seeds, capturing randomness from baseline action sampling and the fixed noise sample used by LLMD.
We introduce out-of-distribution (OOD) setups by varying the initial task setup.
Crucially, the task is always possible to complete in all OOD setups, meaning that not moving is never a valid action for the policy.

\paragraph{Blocks}
The robot needs to stack a small Jenga block on top of another Jenga block.
We vary the location and rotation of the blocks.
This task requires high precision, as the blocks can easily fall.
OOD situations are introduced by varying the block's positions and colors, and by placing three blocks instead of two in the robot's workspace.

\paragraph{Drawer}
The robot needs to open the top shelf, place all red cubes in it, and close it again.
For the training data, we vary the number of cubes from 1 to 3 and place more and differently colored cubes in OOD setups.

\paragraph{Cups}
The robot is tasked with placing any cup into any other smaller cups, out of 5 differently sized cups.
We vary the location of each cup and record all possible stacking combinations for each initial layout.
This introduces significant multi-modality in the training data.
OOD setups consist of cups placed outside the typical boundaries, fewer than 5 cups, and cups of different sizes.

\paragraph{Kitchen}
This is a long-horizon task in a toy kitchen: the robot first needs to remove the lid from the pot, pick up a carrot from the pan, place it in the pot, put the lid back on, and place the pan in the sink.
We introduce OOD situations by placing distractor objects, varying the toy kitchen's position, and using objects other than the carrot.

\paragraph{Stack T}
The robot is presented with two T-shaped objects lying flat on the table.
It needs to pick one up, place it vertically on the table, and then place the other object vertically on top.
This task requires high precision and significant end-effector rotation.
OOD situations are introduced by varying the initial positions of the T-shaped objects by a larger amount than in the training data.

\paragraph{Mixer}
In this language-conditioned task, the robot first needs to open a toy mixer, then place one of two differently colored cups in it, as specified by a language command, and close the mixer again.
The task requires significant rotation of the end effector during interaction with the mixer.
OOD situations are introduced by varying the mixer's and cups' positions more than in the training data.

\begin{figure}[!htbp]
    \centering
    \begin{minipage}[t]{0.42\linewidth}
        \centering
        \vspace{0pt}
        \includegraphics[width=\linewidth]{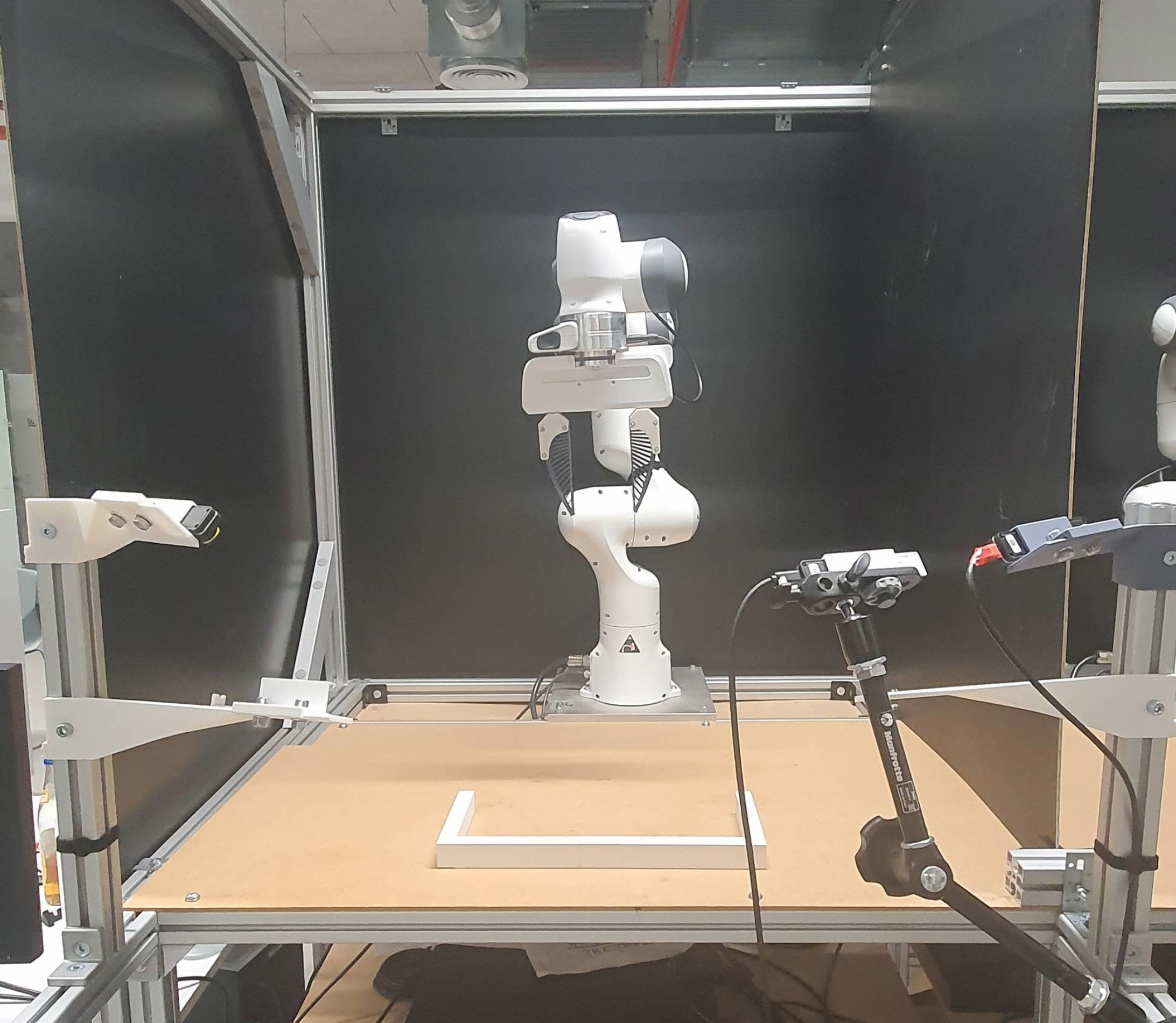}
    \end{minipage}%
    \hspace{3.0pt}
    \begin{minipage}[t]{0.50\linewidth}
        \centering
        \vspace{0pt}
        \begin{minipage}[b]{0.48\linewidth}
            \centering
            \includegraphics[width=\linewidth]{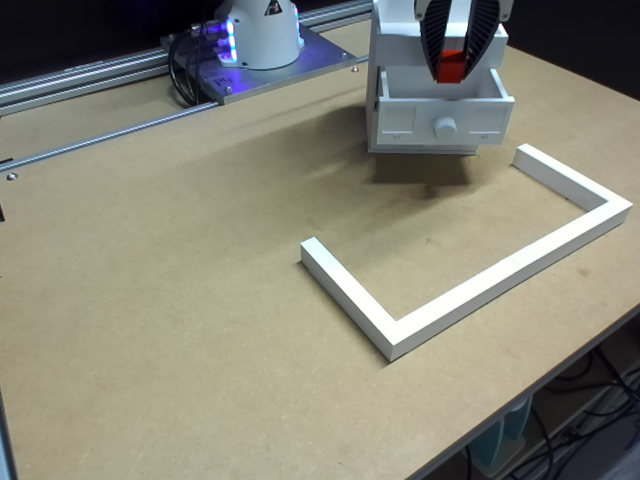}\\[2pt]
            \includegraphics[width=\linewidth]{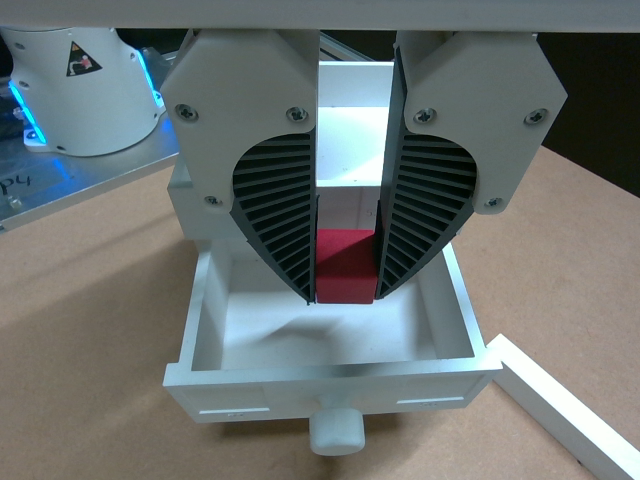}\\[2pt]
        \end{minipage}\hfill%
        \begin{minipage}[b]{0.48\linewidth}
            \centering
            \includegraphics[width=\linewidth]{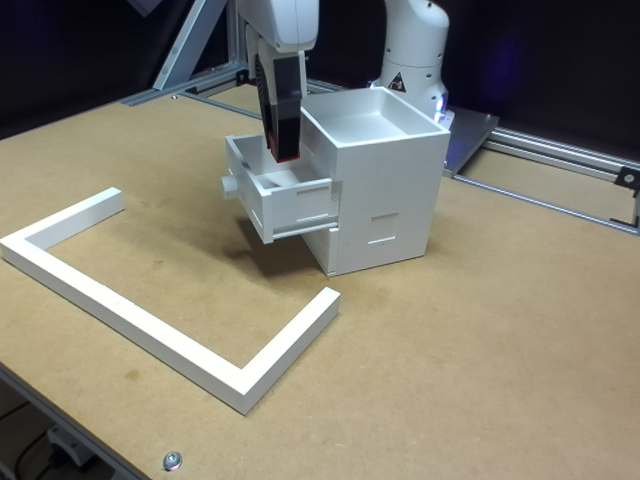}\\[2pt]
            \phantom{\includegraphics[width=\linewidth]{media/real_robot_setup/left_rgb.png}}\\[2pt]
        \end{minipage}
    \end{minipage}
    \caption{
        \textbf{Left}: Real-world setup.
        \textbf{Right}: RGB views from left, right, and gripper cameras.
    }
    \label{fig:results_real_setup}
\end{figure}

\begin{figure*}[!htbp]
    \centering

    \begin{subfigure}[t]{0.48\linewidth}
        \centering
        \includegraphics[width=0.49\linewidth]{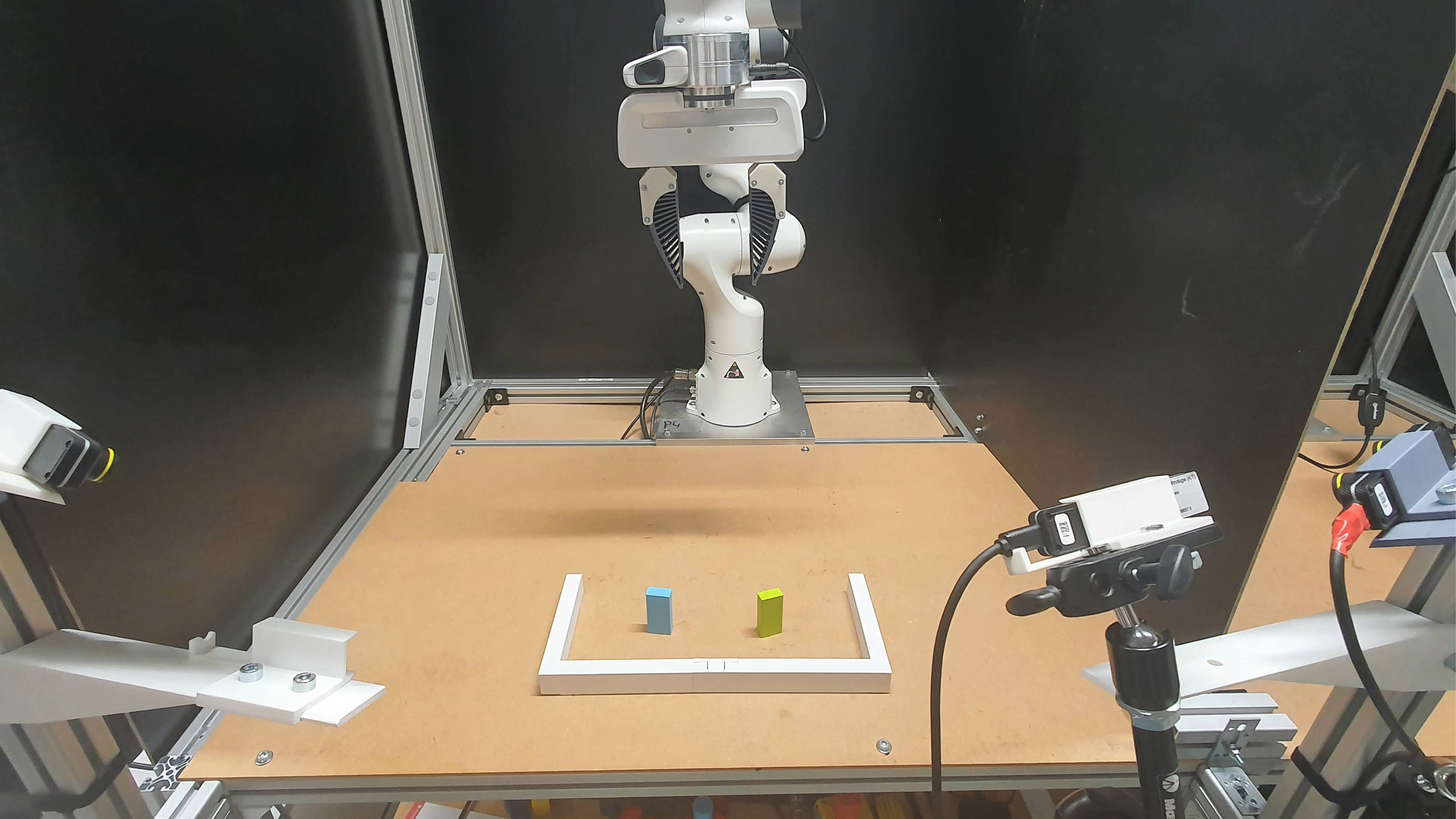}%
        \hfill
        \includegraphics[width=0.49\linewidth]{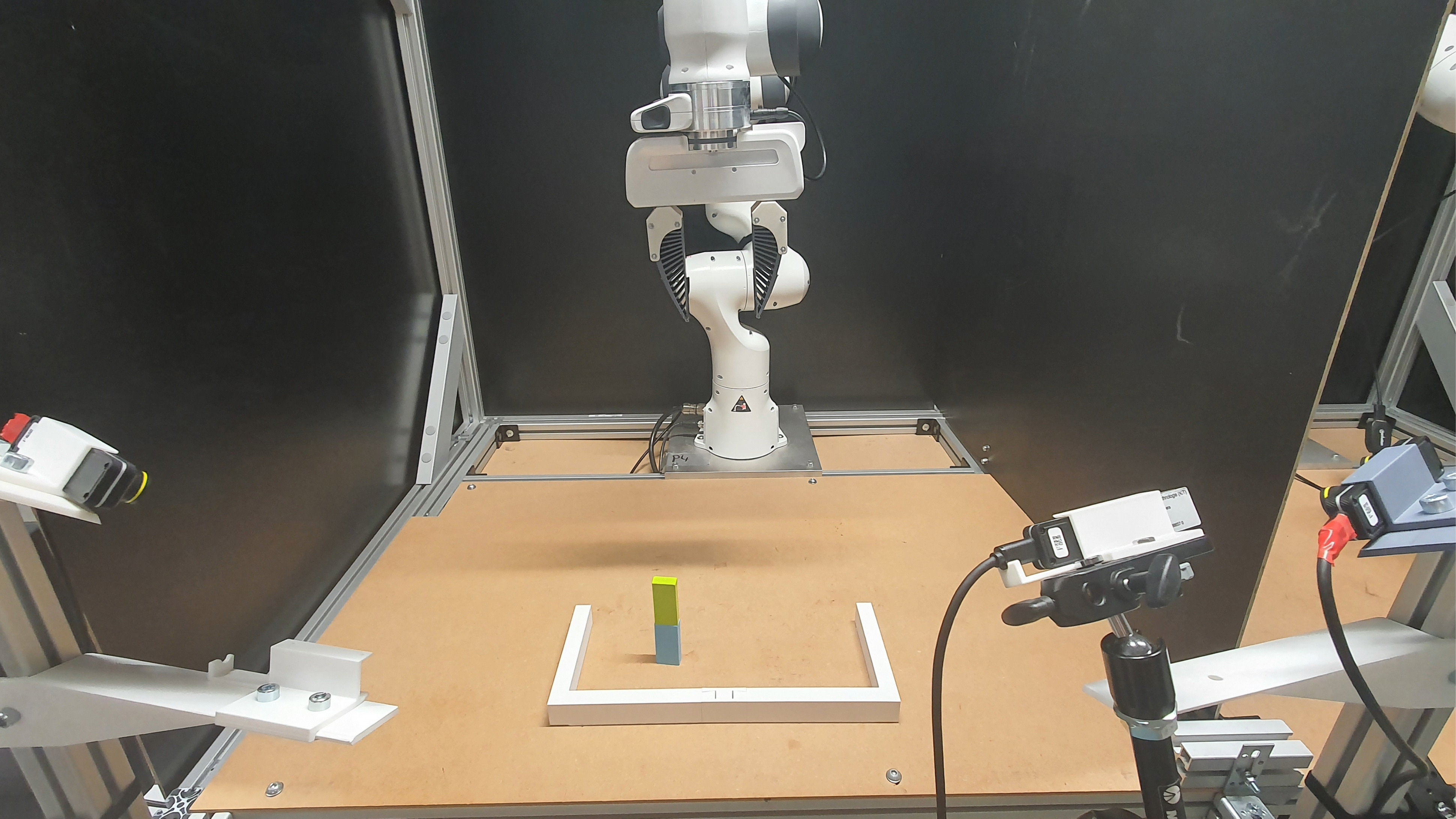}
        \caption{\texttt{Blocks}}
    \end{subfigure}\hfill
    \begin{subfigure}[t]{0.48\linewidth}
        \centering
        \includegraphics[width=0.49\linewidth]{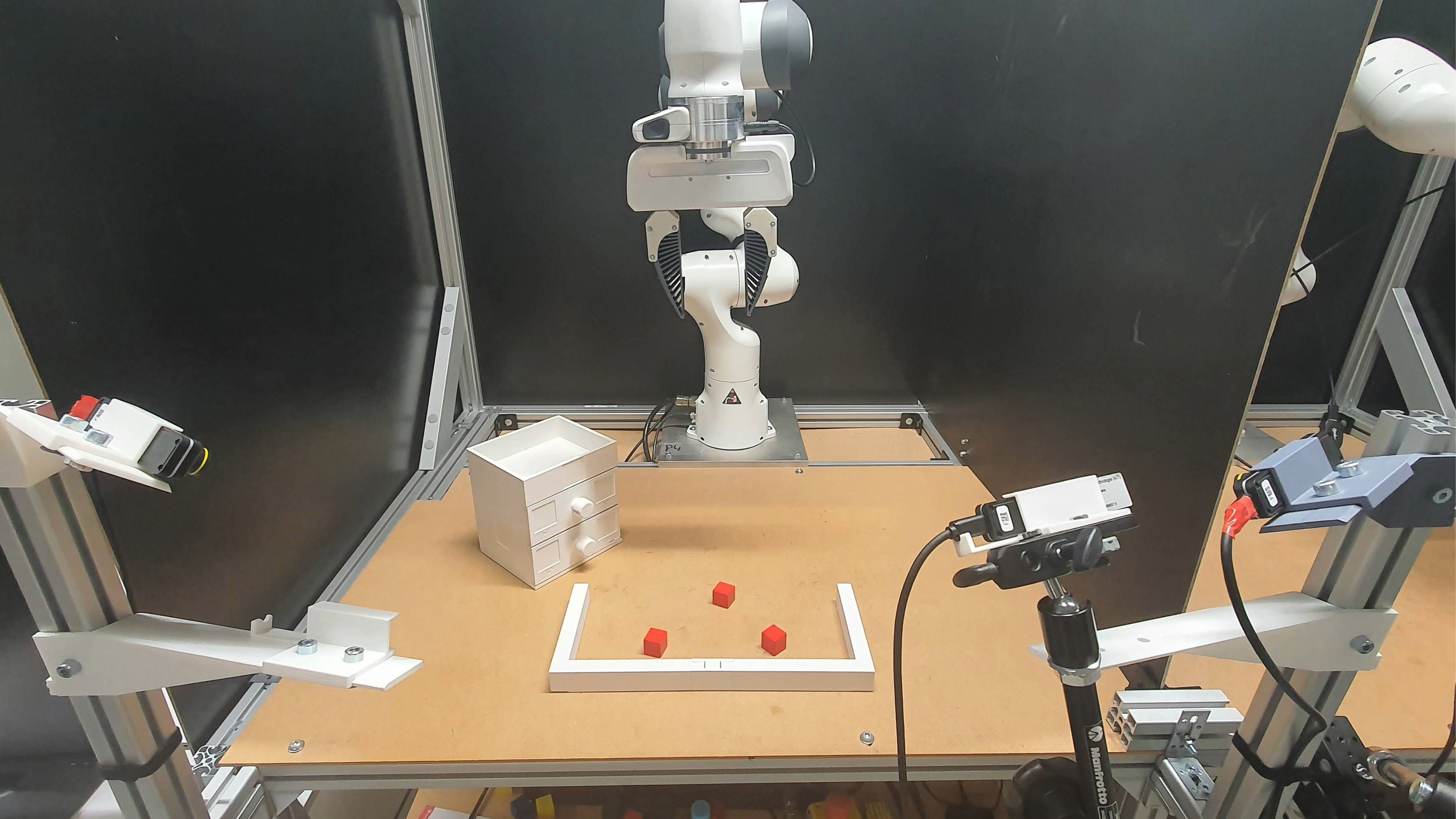}%
        \hfill
        \includegraphics[width=0.49\linewidth]{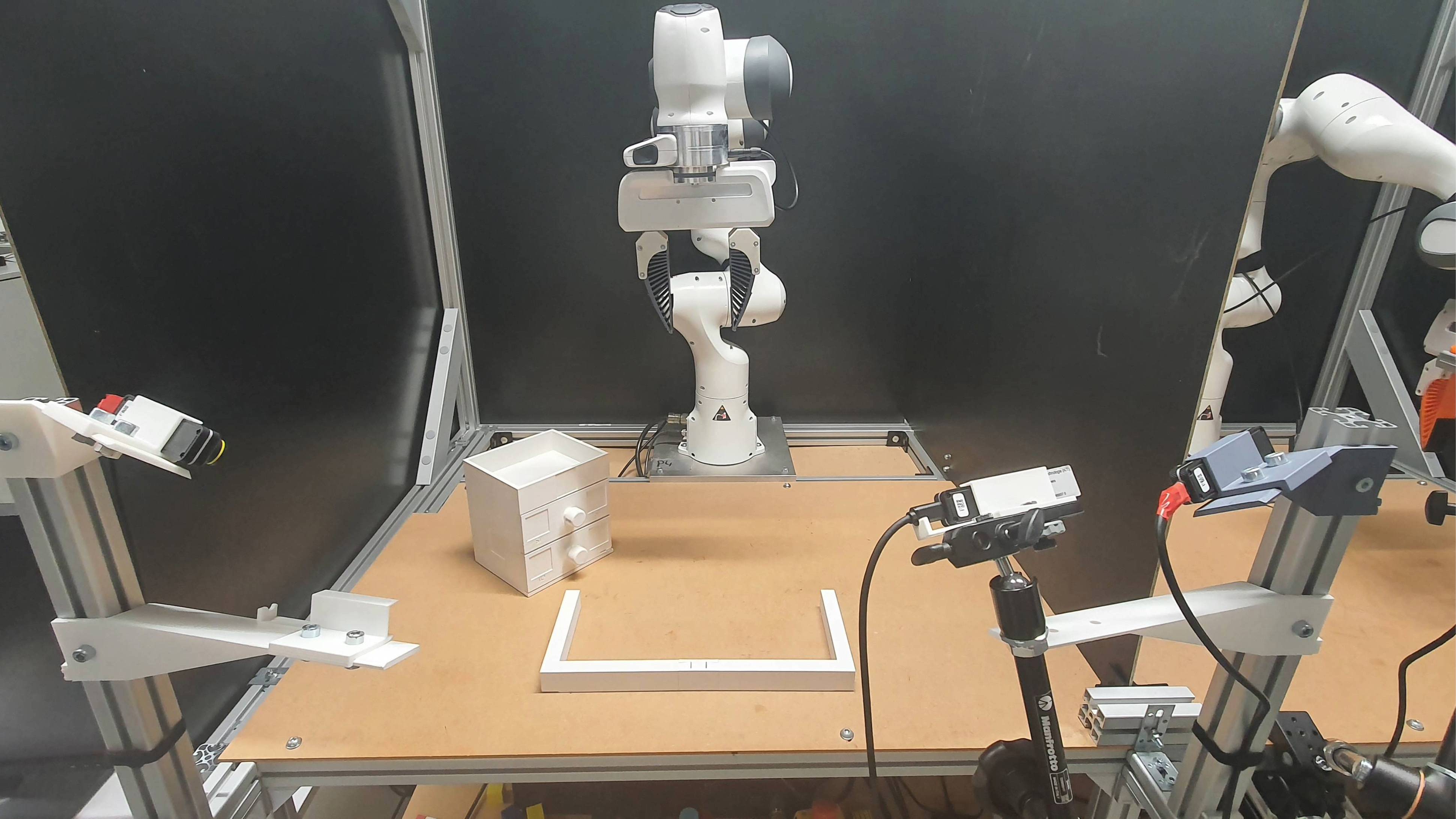}
        \caption{\texttt{Drawer}}
    \end{subfigure}

    \vspace{0.4em}

    \begin{subfigure}[t]{0.48\linewidth}
        \centering
        \includegraphics[width=0.49\linewidth]{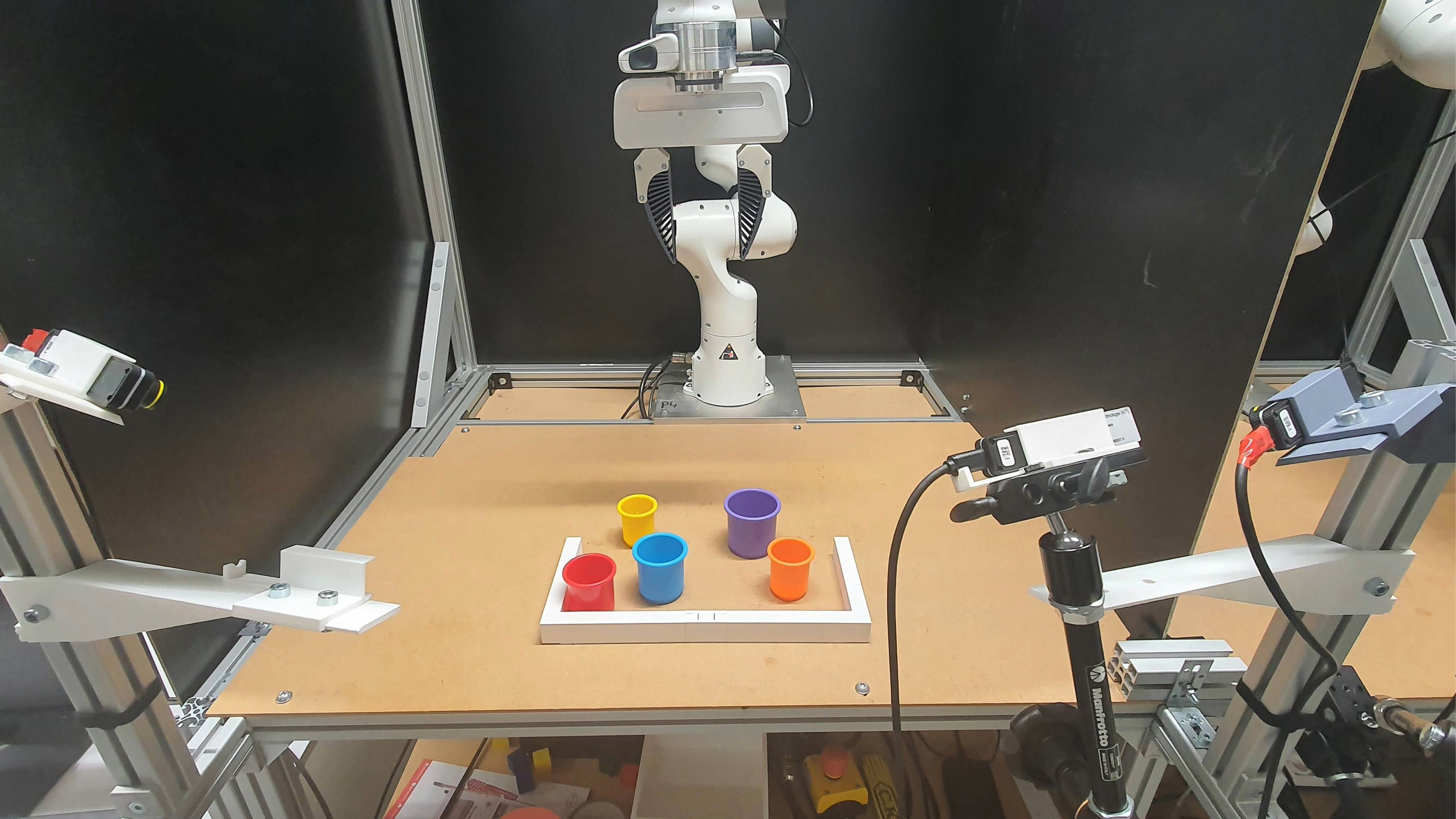}%
        \hfill
        \includegraphics[width=0.49\linewidth]{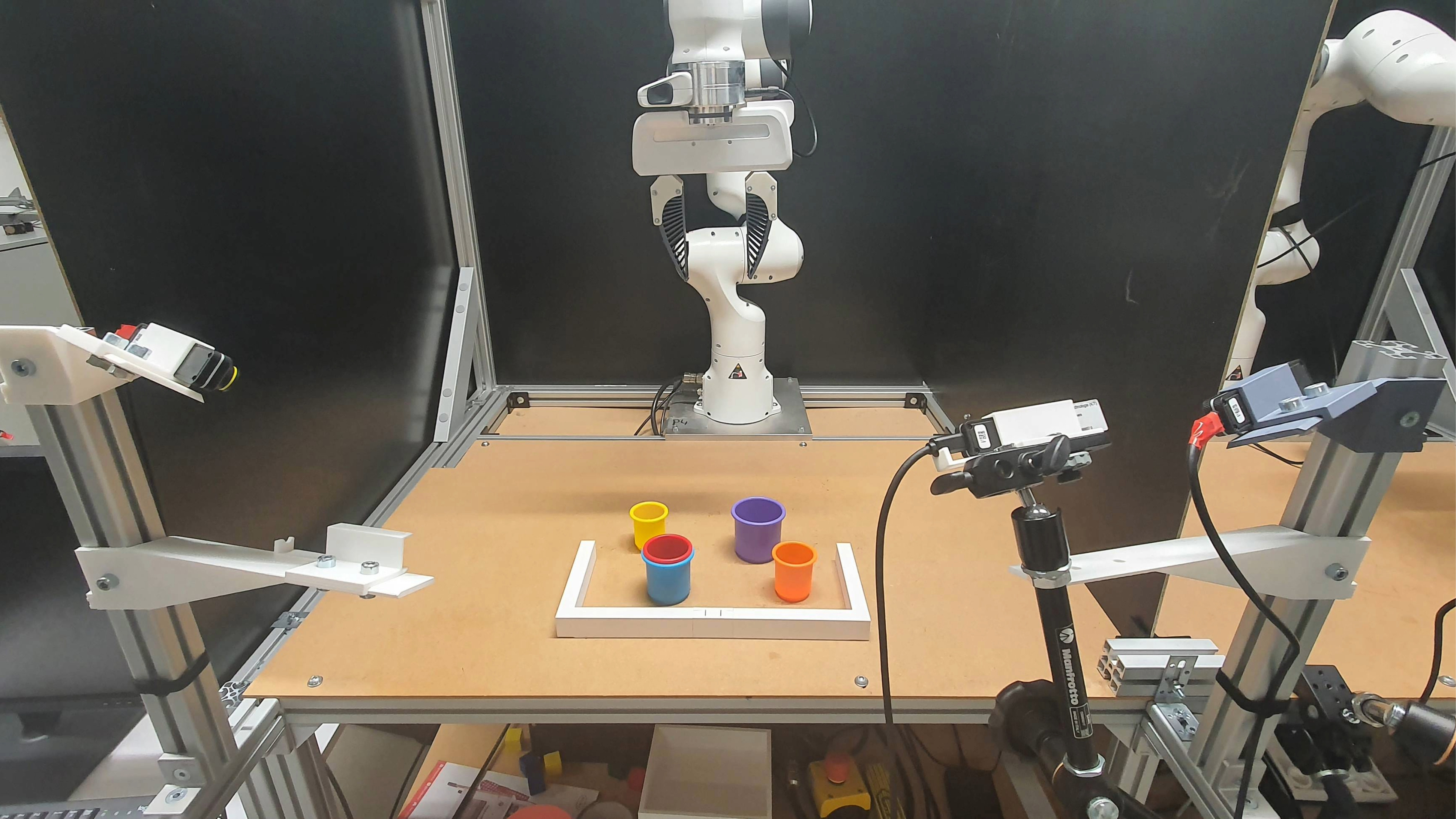}
        \caption{\texttt{Cups}}
    \end{subfigure}\hfill
    \begin{subfigure}[t]{0.48\linewidth}
        \centering
        \includegraphics[width=0.49\linewidth]{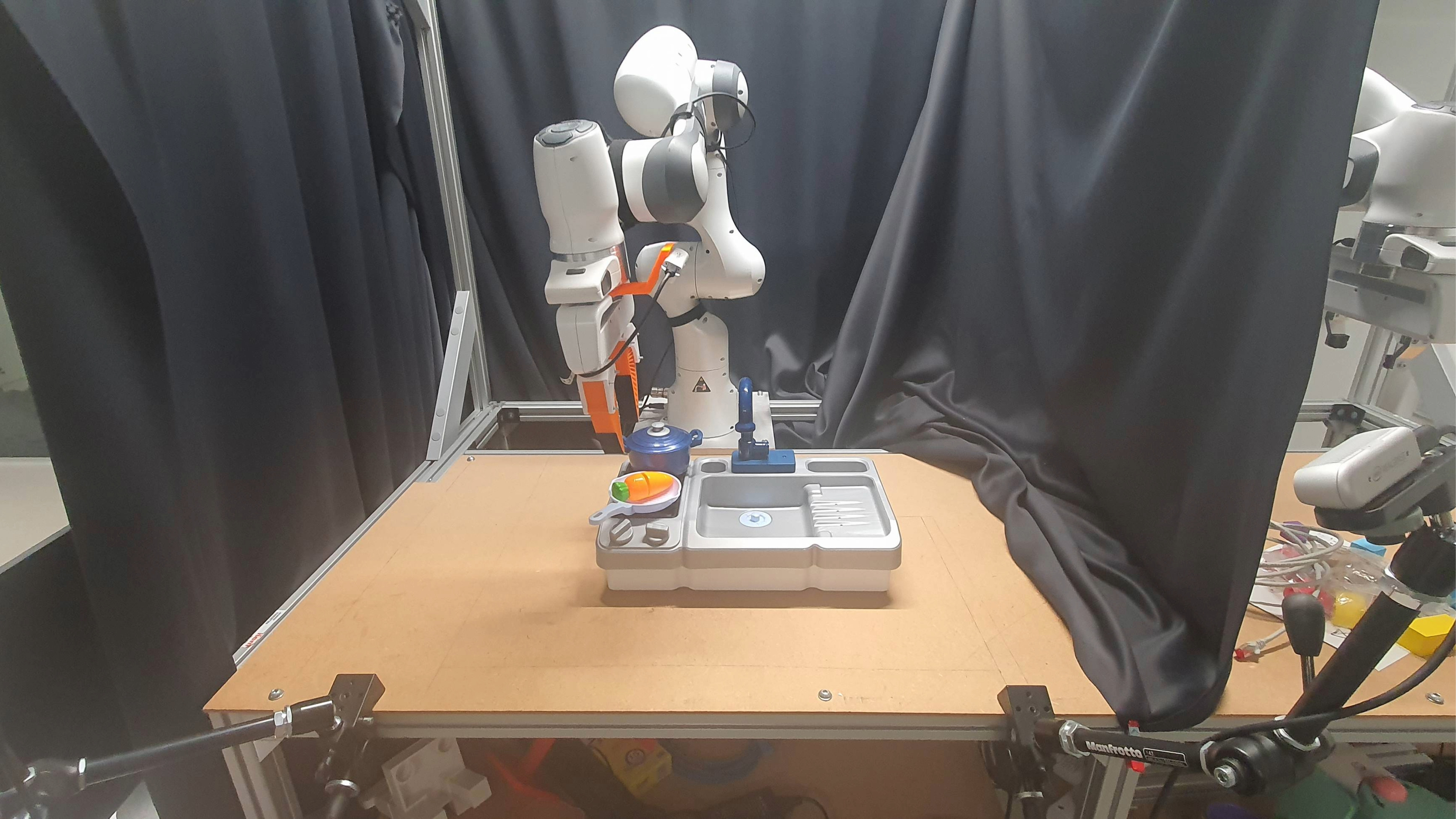}%
        \hfill
        \includegraphics[width=0.49\linewidth]{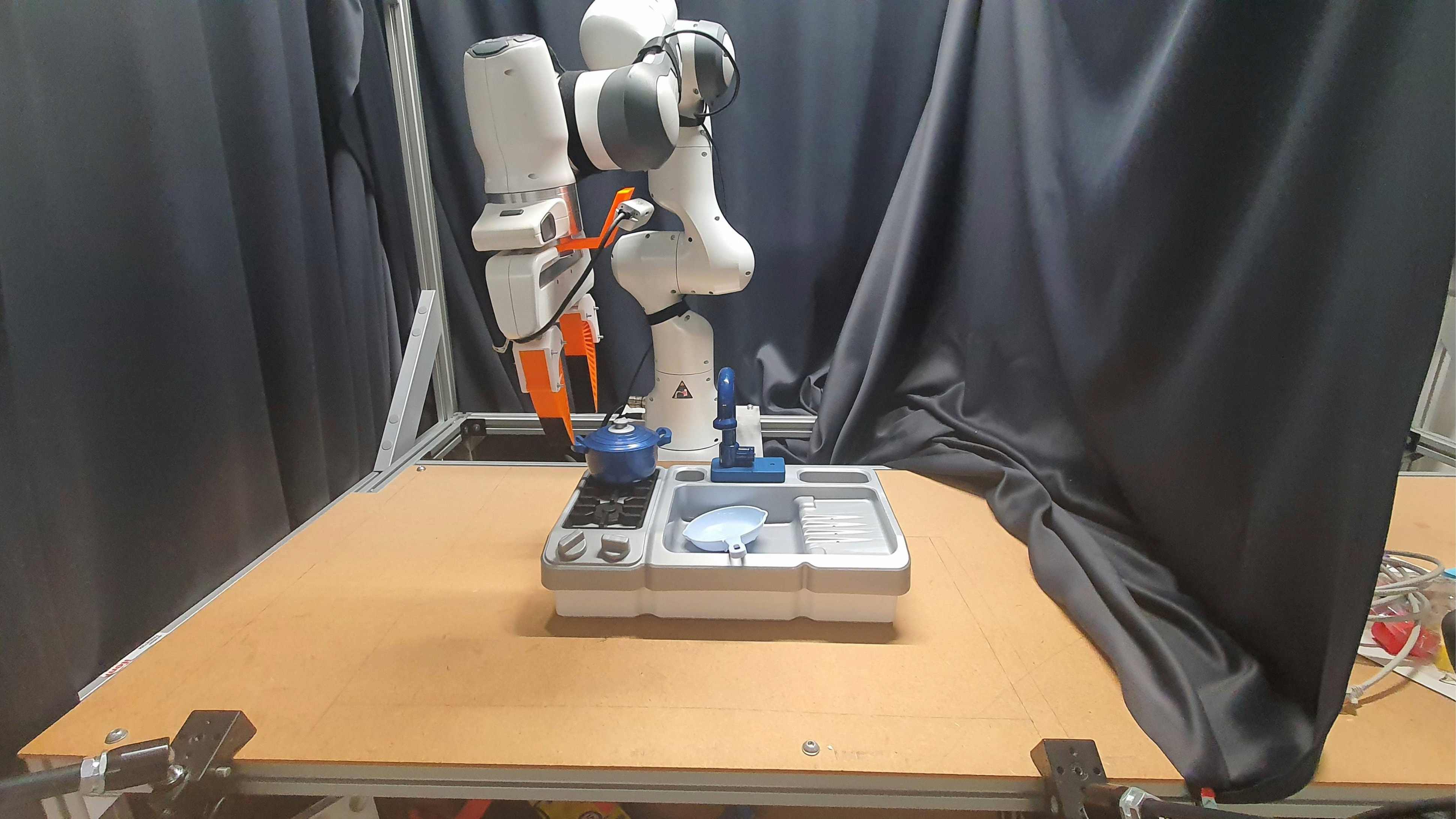}
        \caption{\texttt{Kitchen}}
    \end{subfigure}

    \vspace{0.4em}

    \begin{subfigure}[t]{0.48\linewidth}
        \centering
        \includegraphics[width=0.49\linewidth]{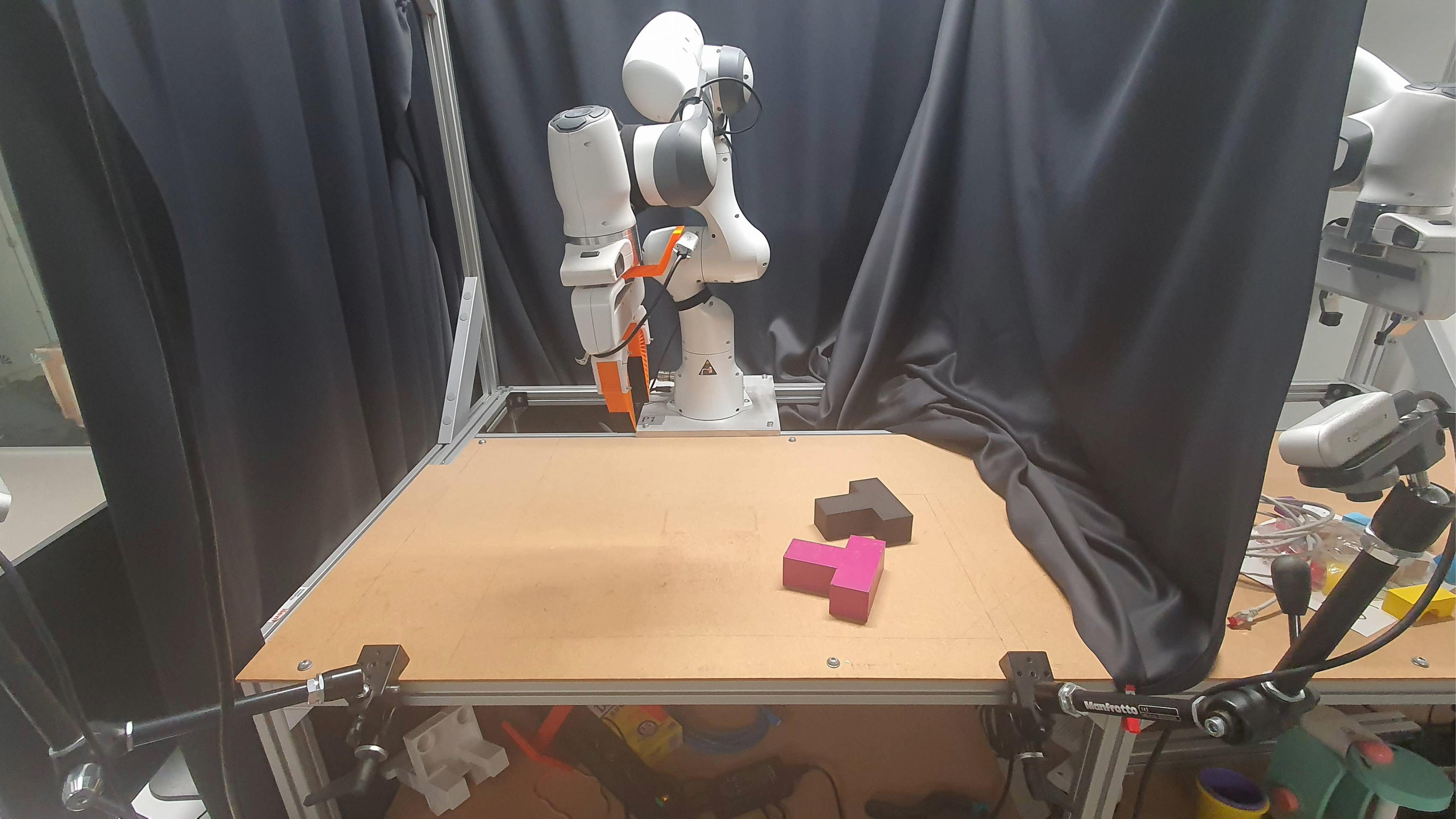}%
        \hfill
        \includegraphics[width=0.49\linewidth]{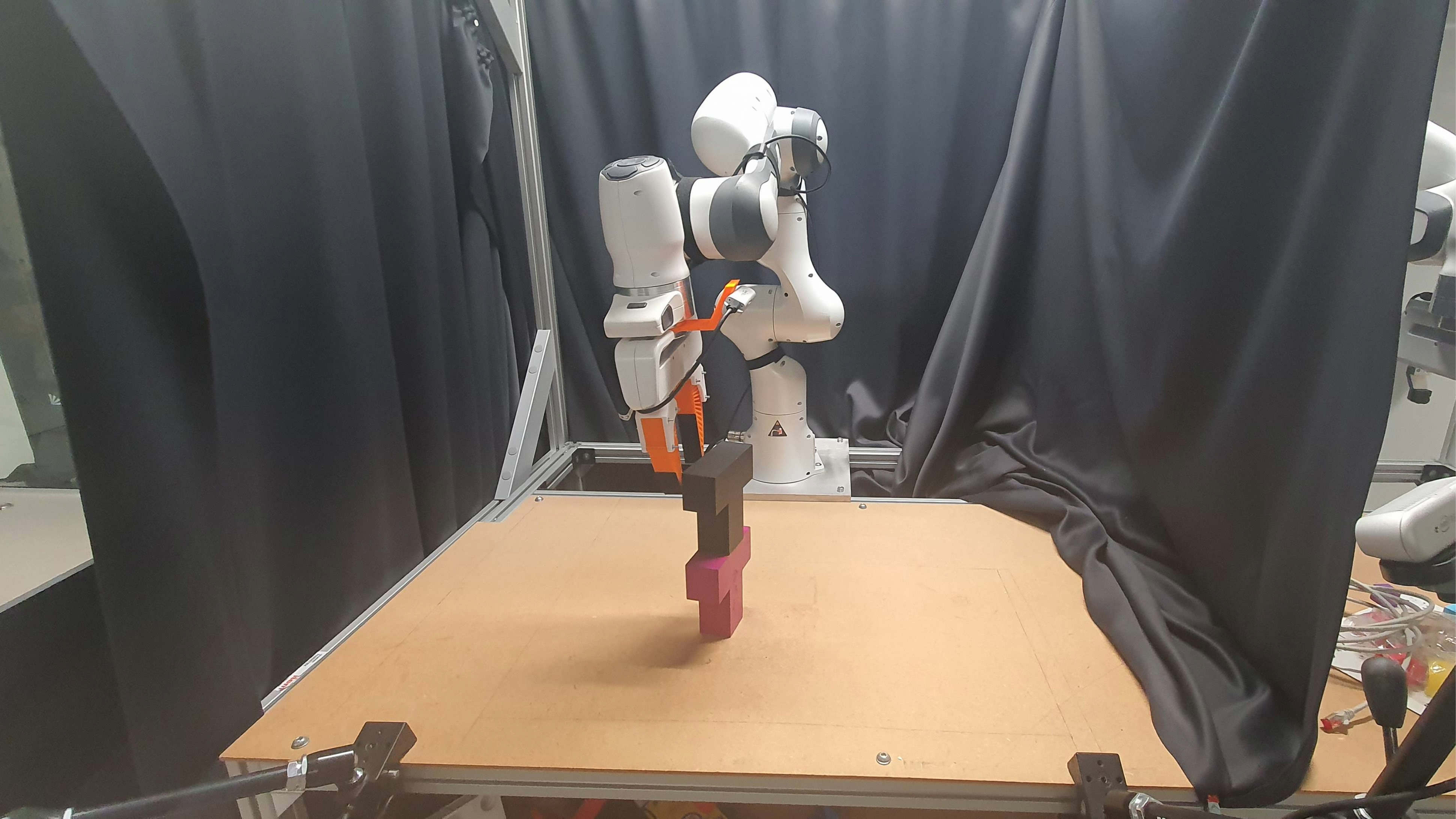}
        \caption{\texttt{Stack T}}
    \end{subfigure}\hfill
    \begin{subfigure}[t]{0.48\linewidth}
        \centering
        \includegraphics[width=0.49\linewidth]{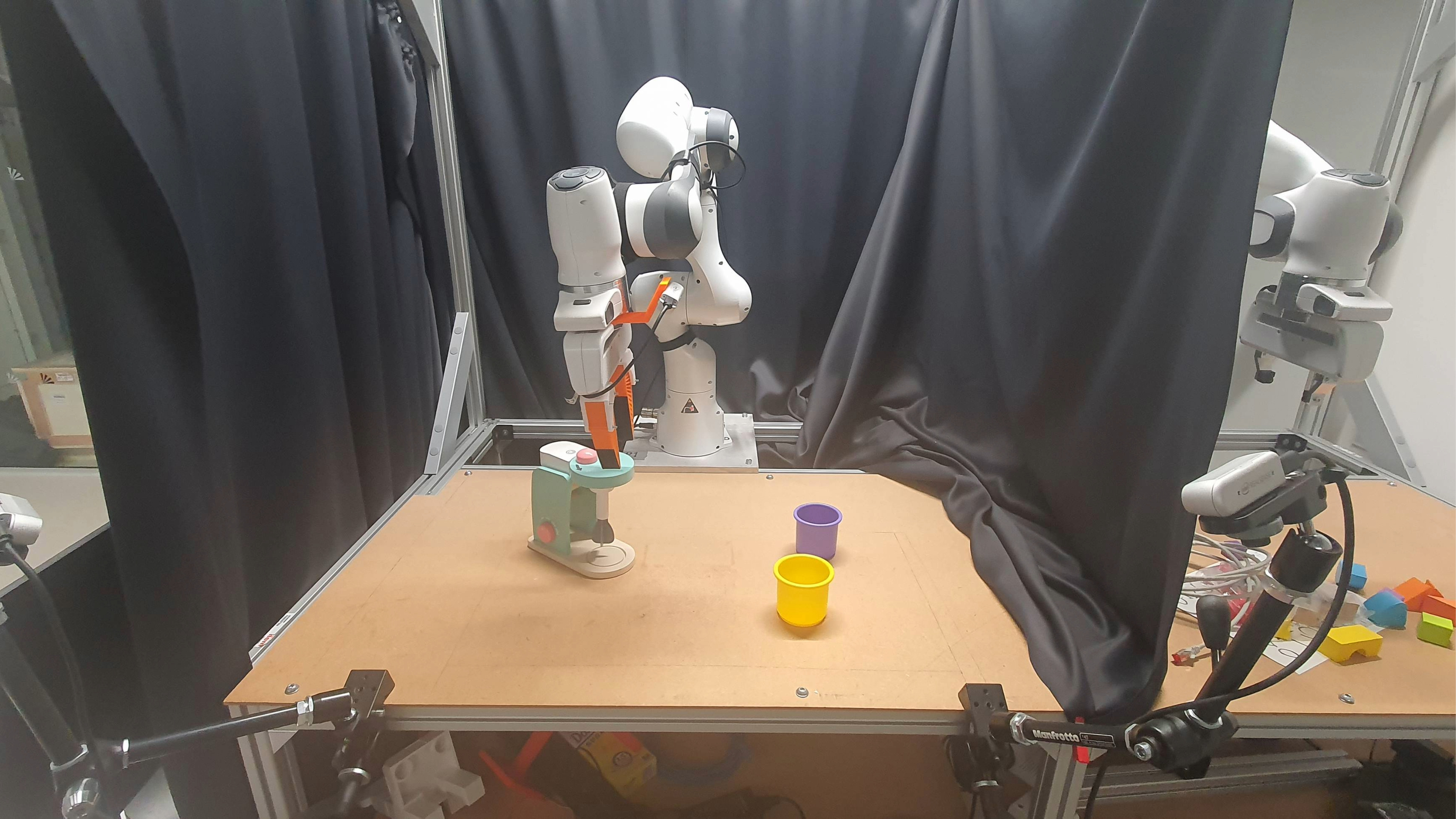}%
        \hfill
        \includegraphics[width=0.49\linewidth]{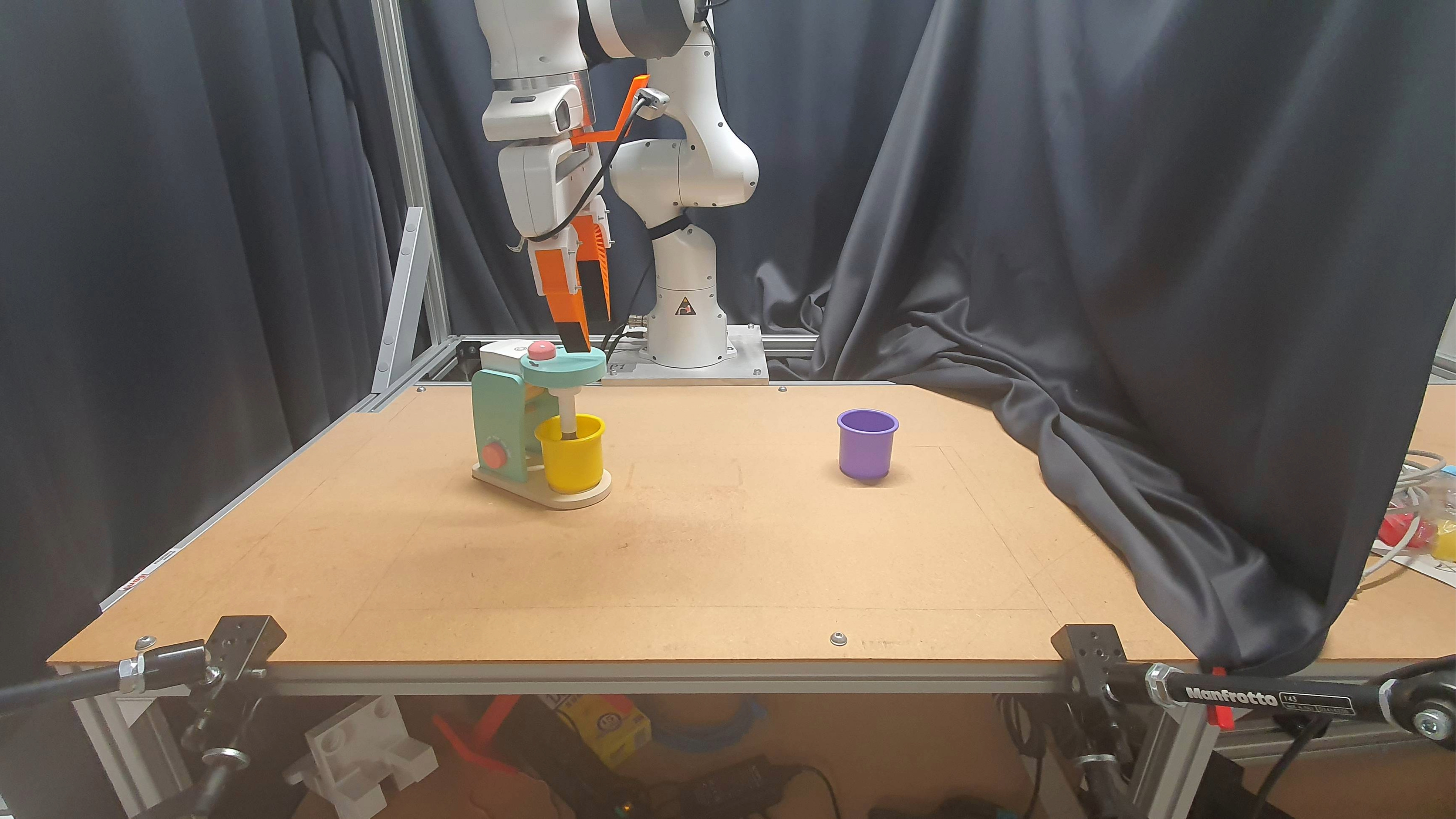}
        \caption{\texttt{Mixer}}
    \end{subfigure}

    \caption{Initial and final states for the real-world tasks. For each task, the left image shows the initial state and the right image shows the final target state.}
    \label{fig:real_world_task_overview}
\end{figure*}

\begin{table*}[!htbp]
    \centering
    \caption{Task descriptions for the real-world manipulation tasks.}
    \label{tab:task_descriptions}
    \tablesize
    \setlength{\tabcolsep}{4pt}
    \renewcommand{\arraystretch}{1.05}
    \begin{tabular*}{\textwidth}{@{\extracolsep{\fill}}lp{0.72\textwidth}@{}}
        \toprule
        \textbf{Task} & \textbf{Description} \\
        \midrule
        \texttt{Blocks}
            & Stack the green block on the blue block. \\
        \midrule
        \texttt{Drawer}
            & Put all red blocks in the top drawer. \\
        \midrule
        \texttt{Cups}
            & Pick up one cup and stack it into another. \\
        \midrule
        \texttt{Kitchen}
            & Remove the pot lid, move the carrot from the pan to the pot, replace the lid, and place the pan in the sink. \\
        \midrule
        \texttt{Stack T}
            & Stack both T-shaped objects on top of each other. \\
        \midrule
        \texttt{Mixer}
            & Put yellow cup in the mixer \\
            & Put lila cup in the mixer \\
        \bottomrule
    \end{tabular*}
\end{table*}

\newpage
\paragraph{Evaluation Protocol}
\label{appendix:eval_protocol}
We train $\pi_{0.5}$ and X-VLA on all tasks and evaluate them on a mixture of in-distribution and out-of-distribution initial object layouts.
All the policies are trained to return to the home position after completing a task and then remain stationary.
During evaluation, we terminate all episodes only after the episode horizon is reached, even when the task was successfully completed before.
This is crucial, as otherwise task failures could be detected with high probability based on the episode length.
Since we expect policies to operate in a setting where a ground-truth success signal is not available, this baseline is not valid.
We evaluate all policies on all tasks across approximately 50 ID and 30 OOD setups per task
As the high computational overhead of ACE, DIFF, and STAC does not allow us to run them in real-time, we store all observations from each rollout and evaluate these methods on episode replays.
To capture the variance induced by the action samples for ACE, Diff, and STAC, and by the random noise vector in LLMD, we evaluate all methods across three seeds, where we use the same rollouts across seeds.
This keeps the observations fixed and only varies the action samples and random noise vector.
ACC is deterministic and therefore has confidence intervals of zero width.

\subsection{Simulation Environments}
\label{app:simulation_environments}

\begin{figure}[!htbp]
    \centering
    \includegraphics[width=\linewidth]{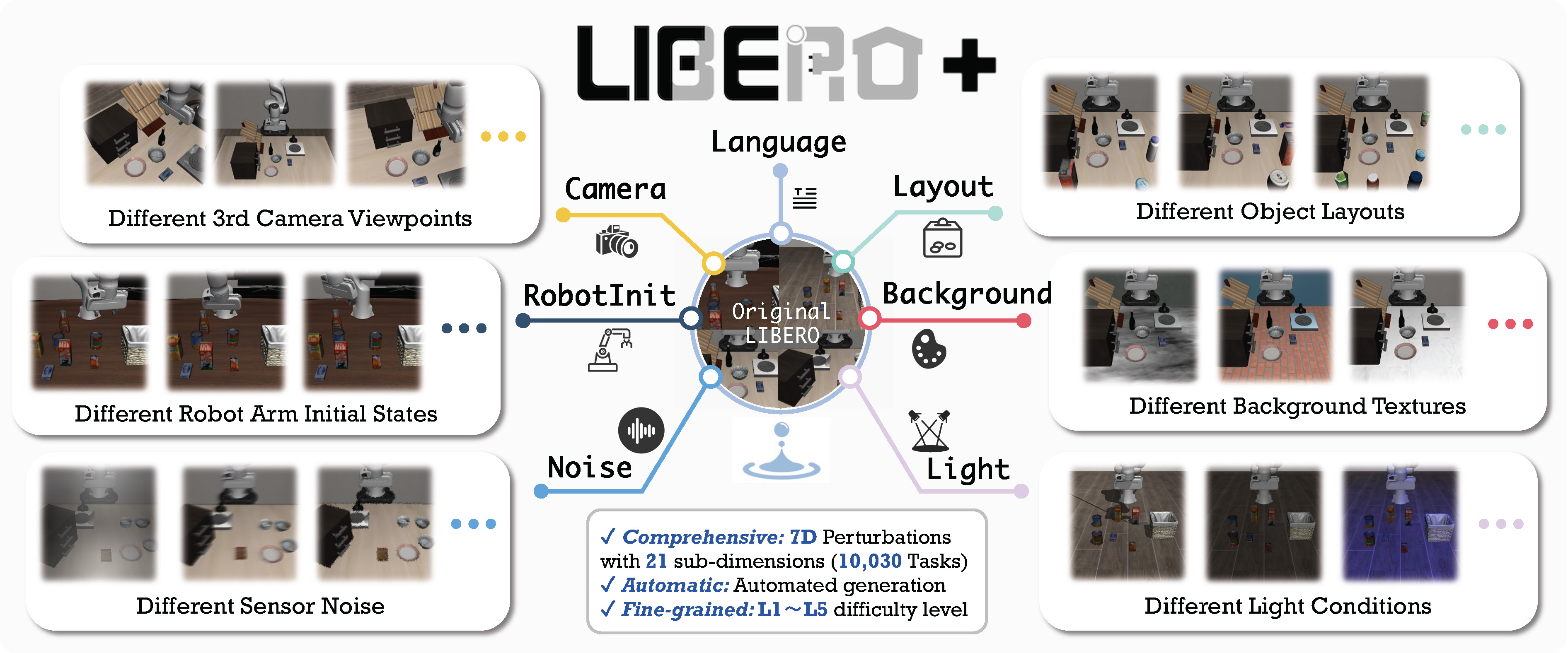}%
    \caption{LIBERO-Plus: In-depth Robustness Analysis of Vision-Language-Action Models. Reproduced from~\citet{fei25libero-plus}.}
\end{figure}

We evaluate simulation failure detection on \texttt{Libero-Plus}~\citep{fei25libero-plus}, which extends the original LIBERO benchmark~\citep{liu2023libero} with additional perturbations to evaluate generalization under distribution shift.
We use four task suites: \texttt{Object}, \texttt{Spatial}, \texttt{Goal}, and \texttt{10}.
The suites test complementary forms of generalization, including changes in object identities, spatial layouts, goal specifications, and diverse long-horizon manipulation behaviors.

For each suite, we finetune the VLA on the corresponding \texttt{LIBERO-Plus} training data and evaluate failure detection on perturbed rollout settings.

\begin{table}[!htbp]
    \centering
    \caption{Summary of the \texttt{Libero-Plus} simulation task suites used in our experiments.}
    \label{tab:libero_plus_suite_summary}
    \tablesize
    \setlength{\tabcolsep}{4pt}
    \renewcommand{\arraystretch}{1.05}
    \begin{tabular*}{\textwidth}{@{\extracolsep{\fill}}lp{0.72\textwidth}@{}}
        \toprule
        \textbf{Suite} & \textbf{Summary} \\
        \midrule
        \texttt{Object}
            & Object-centric manipulation tasks that evaluate generalization across manipulated object identities. \\
        \texttt{Spatial}
            & Spatial-reasoning tasks that evaluate robustness to changes in object positions and scene layouts. \\
        \texttt{Goal}
            & Goal-conditioned manipulation tasks that evaluate generalization across different target goals. \\
        \texttt{10}
            & A diverse long-horizon task suite that evaluates generalization across multiple manipulation skills and task structures. \\
        \bottomrule
    \end{tabular*}
\end{table}

\end{document}